%% file: main.tex
\documentclass{article} 
\usepackage{iclr_style/iclr2025_conference}
\input{iclr_style/math_commands.tex} 
\iclrfinalcopy 

\usepackage{hyperref}
\usepackage{url}
\usepackage{fixltx2e}
 \usepackage{array,multirow}
\usepackage{colortbl}
\usepackage{graphicx}
\usepackage{wrapfig}
\usepackage{lipsum}
\usepackage{textpos}
\usepackage{ragged2e}
\usepackage[none]{hyphenat}
\usepackage{booktabs} 
\usepackage{float}
\usepackage{placeins}
\usepackage[most]{tcolorbox}
\usepackage{enumitem}

\title{Regularization Through Reasoning: Systematic Improvements in Language Model Classification via Explanation-Enhanced Fine-Tuning}

\author{%
Vivswan Shah\textsuperscript{*}, Randy Cogill, Hanwei Yue, Gopinath Chennupati, Rinat Khaziev  \\
Amazon Central Analytics and Research Science\\
\texttt{\textsuperscript{*}vivswanshah@pitt.edu, \{\textsuperscript{*}vivswan, cograndy, hanweiyu, chennug, rinatk\}@amazon.com}
}

\begin{document}

\maketitle

\FloatBarrier
\input{sections/00-abstract}
\input{sections/10-introductions}
\input{sections/20-background}
\FloatBarrier
\input{sections/30-experimental-setup}

\FloatBarrier
\input{sections/40-results}
\FloatBarrier
\input{sections/50-analysis}
\FloatBarrier
\input{sections/60-discussion}
\FloatBarrier
\input{sections/70-conclusion}
\FloatBarrier

\FloatBarrier
\newpage
\bibliography{ref/references.bib}
\bibliographystyle{iclr_style/iclr2025_conference}
\FloatBarrier
\newpage
\appendix
\input{sections/99-appendix}

\end{document}

%% file: iclr_style/math_commands.tex
\usepackage{amsmath,amsfonts,bm}

\def\eqref#1{equation~\ref{#1}}

\def\1{\bm{1}}

\DeclareMathAlphabet{\mathsfit}{\encodingdefault}{\sfdefault}{m}{sl}
\SetMathAlphabet{\mathsfit}{bold}{\encodingdefault}{\sfdefault}{bx}{n}



%% file: sections/00-abstract.tex
\begin{abstract}

Traditional approaches to fine-tuning large language models for classification applications rely on direct label prediction without additional context. This work explores the impact of augmenting classification labels with explanatory text during the fine-tuning process to enhance model performance. Our investigation centers on evaluating conversational response quality across three dimensions: naturalness of language use, response comprehensiveness, and adherence to conversational topics, each assessed using 5-point rating scales. We employ ensemble-generated training data from multiple large language modelTraditional approaches to fine-tuning large language models for classification applications rely on direct label prediction without additional context. This work explores the impact of augmenting classification labels with explanatory text during the fine-tuning process to enhance model performance. Our investigation centers on evaluating conversational response quality across three dimensions: naturalness of language use, response comprehensiveness, and adherence to conversational topics, each assessed using 5-point rating scales. We employ ensemble-generated training data from multiple large language models to fine-tune a 7B parameter architecture, subsequently testing performance across six diverse conversational datasets. The experimental results demonstrate that models trained with combined label-explanation pairs achieve superior performance compared to label-only training approaches across all 18 dataset-task configurations tested. Remarkably, our experiments reveal that models benefit from explanatory content even when it consists of certain types of randomly generated word sequences that maintain vocabulary overlap with original explanations, indicating that structured noise can function as a regularization mechanism during the training process. Internal model analysis demonstrates that explanation-augmented training produces models with elevated entropy levels in intermediate computational layers coupled with more concentrated prediction confidence in output layers, suggesting enhanced deliberation processes prior to final decision-making. These results establish that explanation-augmented fine-tuning represents a viable approach for enhancing both accuracy and reliability in language model classification applications, while simultaneously offering valuable understanding of how explanatory components shape model computational patterns during inference.

\end{abstract}

%% file: sections/10-introductions.tex
\section{Introduction}
\label{section:introduction}

Large language models (LLMs) have demonstrated remarkable capabilities across diverse natural language processing tasks \citep{zhangInstructionTuningLarge2023}, excelling particularly in text generation and complex reasoning scenarios \citep{mulalOrca2TeachingSmall2023}. However, adapting these models for domain-specific classification tasks through fine-tuning presents unique challenges \citep{baiTrainingHelpfulHarmless2022}. A critical concern is ensuring that fine-tuned models develop robust decision-making processes rather than exploiting superficial correlations between inputs and labels \citep{aghajanyanExplanationBasedFinetuningMakes2023}. This challenge becomes particularly pronounced in conversational quality evaluation, where models must demonstrate nuanced understanding of multiple communication aspects simultaneously.

Building on recent advances in explanation-based learning \citep{lampinen2022can, zhaoShowMeHow2024}, this paper investigates how incorporating explanations during fine-tuning affects model performance on conversational quality assessment tasks. We focus on three fundamental dimensions of conversation quality: naturalness of language, comprehensiveness of responses, and adherence to topic. These dimensions represent critical components of effective human-computer interaction and are applicable across diverse conversational contexts \citep{zhengJudgingLLMJudgeMTBench2023}. For each dimension, models learn to predict ratings on a 5-point scale, with higher scores indicating superior quality.

Our experimental approach centers on fine-tuning a 7B parameter student model using training data generated from an ensemble of larger teacher models. This methodology leverages the knowledge of more capable teacher models while maintaining the computational efficiency of smaller student models for practical deployment \citep{ho2022teaching}. We systematically compare two fine-tuning strategies: training exclusively on class labels versus training on both class labels and natural language explanations that justify those labels. To ensure comprehensive evaluation, we test these approaches across 6 diverse conversational datasets, evaluating performance on 3 quality dimensions each, yielding 18 distinct dataset-task combinations that span various conversational contexts and domains.

Our investigation reveals a consistent pattern: models fine-tuned with explanations outperform those trained on labels alone across all 18 dataset-task combinations. The improvement is robust and generalizable, occurring regardless of the specific conversational domain or quality dimension being evaluated. Intriguingly, even when explanations consist of random word sequences, models still achieve better performance than baseline approaches, suggesting that explanation tokens may provide regularization benefits during training.

To understand the mechanisms underlying these performance gains, we conducted comprehensive analyses of model internals and behavior patterns. Our entropy analysis across transformer layers reveals that explanation-trained models maintain higher uncertainty in intermediate processing stages while producing more confident final predictions. This suggests a fundamentally different decision-making process: explanation-trained models appear to explore multiple possibilities more thoroughly before converging on their final outputs, whereas models trained without explanations commit to decisions earlier in their computational pipeline.

We also investigate how explanation characteristics affect learning outcomes. Our experiments with varying explanation lengths demonstrate that substantial benefits can be achieved with relatively short explanations (25-50 words), with diminishing returns for longer texts. Analysis of explanation content reveals that the most informative portions typically appear at the beginning, providing early contextual cues that guide the model's reasoning process.

Our contributions are threefold. First, we provide systematic empirical evidence that explanation-based fine-tuning consistently improves classification performance across diverse conversational domains. Second, we offer mechanistic insights into how explanations influence model behavior, revealing distinct computational patterns in explanation-trained versus standard fine-tuned models. Third, we demonstrate practical applications of these findings, showing how explanation-based approaches can enhance model robustness and generalization capabilities.

These results have important implications for language model development and deployment. They suggest that incorporating explanations during fine-tuning represents a practical strategy for improving model reliability and performance, particularly in applications requiring nuanced judgment and robust generalization. Our work also opens new research directions in understanding how auxiliary training objectives can shape model learning dynamics and decision-making processes.


\begin{figure}[b]
   \centering
   \includegraphics[width=0.95\textwidth]{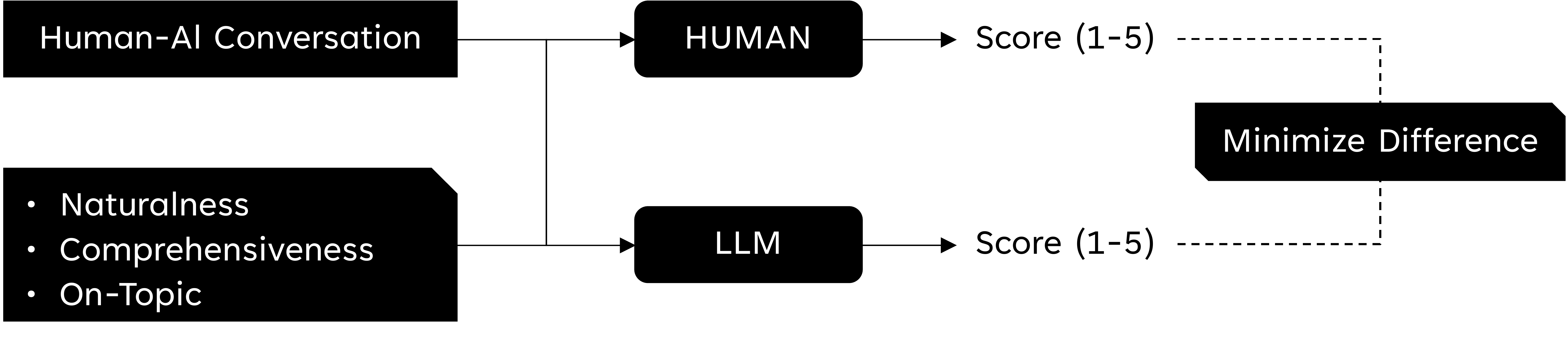}
   \caption{The objective for this study is to minimize the difference between human and LLM rating.}
   \label{fig:background-minimise-human-llm}
\end{figure}

%% file: sections/20-background.tex
\section{Background and Related Work}
\label{section:background}

\subsection{Related Work}

The integration of explanations in machine learning has garnered significant attention as a means to improve model performance and interpretability. \citet{camburu2018snli} pioneered the use of natural language explanations in NLP with e-SNLI, demonstrating that models can be trained to generate explanations alongside predictions for natural language inference tasks. This work established the foundation for explanation-augmented training in language models.

Recent advances in explanation-based fine-tuning have shown promising results across various domains. \citet{aghajanyanExplanationBasedFinetuningMakes2023} demonstrated that explanation-based fine-tuning makes models more robust to spurious correlations, leading to better generalization on challenging inference tasks. Similarly, \citet{zhaoShowMeHow2024} investigated the role of explanations in fine-tuning and found that explanation-augmented training consistently improves model performance compared to standard approaches.

The development of reasoning capabilities in smaller language models has been a key focus of recent research. \citet{mulalOrca2TeachingSmall2023} showed that smaller models can achieve remarkable performance when trained with step-by-step explanations from larger teacher models, while \citet{ho2022teaching} demonstrated effective methods for teaching reasoning to small language models through explanation-based fine-tuning. These works highlight the potential for explanation-based approaches to bridge the performance gap between large and small models.

\citet{lampinen2022can} provided a formal framework for understanding when and how models can learn from explanations, identifying key factors that determine explanation effectiveness. Their theoretical analysis complements empirical findings and provides guidelines for designing explanation-based learning systems. The broader field of instruction tuning \citep{zhangInstructionTuningLarge2023} has also emphasized the importance of providing models with rich training signals beyond simple input-output pairs.

\subsection{Task Definition and Evaluation Framework}

Building on this foundation, our work focuses on conversational quality assessment, a task that requires nuanced understanding of multiple communication aspects. We evaluate each conversation by scoring it along three fundamental dimensions that capture essential aspects of effective communication. These metrics are obtained through a systematic evaluation process using both human expert judgment and large language model assessments, as detailed in our experimental setup.

We define the three evaluation dimensions as follows:
\begin{itemize}[leftmargin=*]
   \item \textbf{Naturalness} assesses how natural a response sounds. A response which sounds artificial, robotic or unlike how a human would respond would score low on naturalness. Other factors which may lower the naturalness score could be repetitive phrases, redundant follow-up questions, a mismatch in responding to the user's conversational style, tone or emotions. In general, if a conversation has a scripted quality to it, it would score poorly on this dimension. 
    \item \textbf{Comprehensiveness} measures if a response addresses all the salient details in a user's question or request. A highly comprehensive response offers a thorough and holistic overview of all the requested information, while capturing diverse perspectives if applicable. If a response were incomplete, missing critical information, or failing to cover the requested topic adequately, it would score poorly on comprehensiveness.
    \item \textbf{On-Topic} refers to how directly relevant and aligned a response is to the user's conversational intent, without deviating into unrelated information. A completely on-topic response stays focused on the user's query. Lower ratings indicate the response becomes increasingly indirectly related or incorporates off-topic elements. A totally off-topic response lacks meaningful connection to the user's original context.
\end{itemize}

\subsection{Golden Dataset and Evaluation Baseline}

To provide a reliable reference standard for our experiments, we developed an internal golden dataset that serves multiple critical functions in our evaluation framework. This curated dataset consists of 500 carefully selected conversational exchanges that represent high-quality examples across diverse interaction contexts. The dataset serves as both a validation benchmark for our scoring methodology and a consistent evaluation standard against which all model performance is measured throughout our experiments.

The golden dataset contains real conversational interactions spanning various domains and communication patterns, providing a representative sample of the conversational quality challenges that our models must address. Each conversation in the dataset includes complete multi-turn exchanges between users and AI systems, capturing the natural flow and complexity of authentic dialogues. This dataset contains proprietary conversational data and therefore specific examples cannot be disclosed in this publication.

The curation process involved a rigorous multi-stage evaluation protocol conducted by a team of domain experts, including scientists and engineers with extensive experience in conversational AI systems. Each conversational exchange was independently evaluated by two expert annotators who assessed all three quality dimensions (naturalness, comprehensiveness, and on-topic relevance) using a standardized 1-4 point scale. Only conversations where both evaluators achieved perfect agreement on all three dimensions were included in the final golden dataset, ensuring exceptionally high annotation quality and reliability.

This golden dataset provides several key advantages for our experimental framework. First, it establishes consistent evaluation standards that remain constant across all experiments, enabling fair comparison between different fine-tuning approaches. Second, it serves as a validation mechanism for our automated scoring systems, allowing us to calibrate and verify the accuracy of our model-generated evaluations. Finally, it provides ground truth labels that enable us to measure the effectiveness of different explanation-based fine-tuning strategies in achieving human-level assessment quality.

%% file: sections/30-experimental-setup.tex
\section{Experimental Setup}
\label{section:experimental-setup}
\FloatBarrier

To study the impact of explanation-based fine-tuning on model performance, we first needed datasets with explanations. However, no such dataset existed publicly, so we had to construct our own. Since we are focusing on evaluating conversations, as mentioned in the Introduction (Section \ref{section:introduction}), we selected the following public conversational datasets to generate conversation evaluation scores and explanations:

\begin{itemize}[leftmargin=*]
    \item \textbf{ChatbotAC: }Chatbot Arena Conversations \citep{zhengJudgingLLMJudgeMTBench2023}
    \item HH-RLHF: Human preference data about helpfulness and harmlessness \citep{baiTrainingHelpfulHarmless2022}
    \begin{itemize}
        \item \textbf{HH-C: } Chosen Conversations from HH-RLHF.
        \item \textbf{HH-R: } Rejected Conversations from HH-RLHF.
    \end{itemize}
    \item \textbf{AmazonQA} \citep{guptaAmazonQAReviewBasedQuestion2019}
    \item \textbf{CoQA: }Conversational Question Answering \citep{reddyCoQAConversationalQuestion2019}
    \item \textbf{Movies: }Cornell Movie Dialogs Corpus \citep{Danescu-Niculescu-Mizil+Lee:11a}
\end{itemize}

These datasets represent a diverse set of conversations, including both human-LLM and human-human interactions. The Chatbot Arena Conversations and HH-RLHF datasets contain conversations between language models and humans, while AmazonQA, CoQA, and the Cornell Movie Dialogs Corpus contain conversations between humans. AmazonQA and CoQA are conversational question-answering datasets, while the Cornell Movie Dialogs Corpus contains more natural, open-ended conversations. By including this variety of datasets, we aim to study the impact of explanation-based fine-tuning in a robust manner.

\subsection{Conversation Evaluation Score and Explanation Generation}
Since the golden dataset was scored with a scale of 1-4, but in real-world scenarios, most surveys and forms utilize a scale of 1-5, we will test both scales with and without examples. To generate the conversation evaluation scores and explanations, we will employ four distinct prompts:
\begin{itemize}
    \item Prompt with scale 1-4
    \item Prompt with scale 1-4 with Examples
    \item Prompt with scale 1-5
    \item Prompt with scale 1-5 with Examples
\end{itemize}

These four prompts were sent to three foundation models:
\begin{itemize}
    \item Titan Text G1 - Premier [amazon.titan-text-premier-v1:0]
    \item Mixtral 8x7B Instruct [Mixtral-8x7B-Instruct-v0.1] \citep{jiangMixtralExperts2024}
    \item Mistral 7B Instruct [Mistral-7B-Instruct-v0.2] \citep{jiangMistral7B2023}
\end{itemize}

This process created 12 scores and 12 explanations for each data point in the dataset for each evaluation dimension (naturalness, comprehensiveness, on-topic). The results from these prompts are discussed in Section \ref{section:results-datasets}.

\subsection{Conversation Evaluation Scores and Explanations Merger}
\begin{figure}
    \centering
    \includegraphics[width=0.95\textwidth]{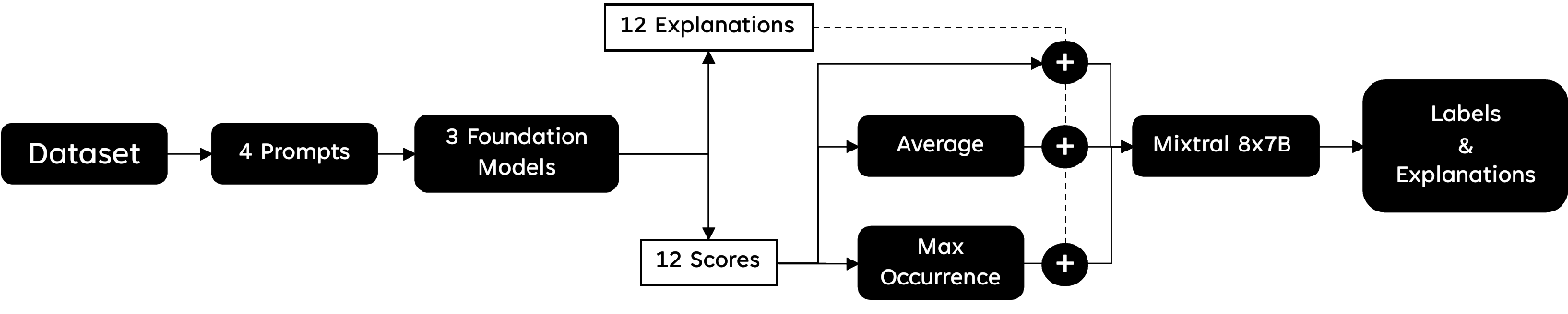}
    \caption{Process to merge multiple scores and explanations into one.}
    \label{fig:explanations-merge-system}
\end{figure}

As shown in Figure \ref{fig:explanations-merge-system}, three different approaches were tried to combine the 12 scores generated for each data point:

\begin{enumerate}[leftmargin=*]
    \item \textbf{Average: }The average of all 12 scores was taken as the final score.
    \item \textbf{Most Occurrence: }The score that occurred most frequently among the 12 was chosen as the final score.
    \item \textbf{LLM: }All 12 scores along with their explanations were sent to the Mixtral 8x7B Instruct model, which selected the final score.
\end{enumerate}

To merge the 12 explanations into one coherent explanation, the Mixtral 8x7B Instruct model was used to combine and summarize the explanations. This process created a unified conversation evaluation scores and explanations for each data point in the datasets, which can be used as ground truth for model fine-tuning. 

The results of this merge process are discussed in Section \ref{section:results-datasets}.

\subsection{Fine-tuning}
We fine-tuned the Mistral-7B-v0.1 model \citep{jiangMistral7B2023} with 4-bit quantization and LoRA adapters \citep{huLoRALowRankAdaptation2021} on each of the datasets, with and without explanations. The loss during training was calculated based on both conversation evaluation scores and explanations when explanations were present. However, the loss during evaluation was calculated solely on the conversation evaluation scores. This approach provides a fair way to compare the loss between models fine-tuned with and without explanations. The fine-tuning parameters used are listed in Table \ref{table:fine-tuning-parameters}.

\begin{table}[h]
    \caption{Fine-tuning Parameters}
    \label{table:fine-tuning-parameters}
    \begin{center}
    \begin{small}
    \begin{sc}
        \begin{tabular}{ll}
            \toprule
            \textbf{Parameter} & \textbf{Value} \\
            \midrule
            \textbf{Model} & Mistral-7B-v0.1 \\
            \textbf{Quantization} & 4-bit \\
            \textbf{Quantization Type} & Normalized Float 4\\
            \textbf{Double Quantization} & True\\
            \textbf{Quantization Compute dtype} & bfloat16\\
            \textbf{LoRA Rank} & 16 \\
            \textbf{LoRA Alpha} & 32 \\
            \textbf{LoRA Dropout} & 0.05 \\
            \textbf{LoRA Bias} & None \\
            \textbf{LoRA Targets} & Query, Key, Value, Output, Gate, Up \& Down projection \\
            \textbf{Optimizer} & AdamW($lr = 5 * 10 ^{-5}$) \\
            \textbf{Train Size} & 10000 \\
            \textbf{Test Size} & 5000 \\
            \textbf{Train Loss on} & Conversation Evaluation Scores and Explanations \\
            \textbf{Test Loss on} & Conversation Evaluation Scores \\
            \bottomrule
        \end{tabular}
    \end{sc}
    \end{small}
    \end{center}
    \vskip -0.1in
\end{table}

%% file: sections/40-results.tex
\section{Results}
\label{section:results}
\FloatBarrier

\subsection{Datasets}
\label{section:results-datasets}
As seen in Table \ref{table:prompt-performance}, when evaluated on the golden dataset, all four prompts performed similarly. However, the average merge method had the lowest mean absolute error (MAE) and mean squared error (MSE) across all conversational evaluation dimensions, closely followed by the most occurrence method. Hence, we used the average merge method to create the scores and explanations for all the conversations in the datasets. 

\begin{table}[h]
    \caption{Performance of Prompts on Golden Datasets using Mean Absolute Error and Mean Squared Error (lower is better).}
    \label{table:prompt-performance}
    \begin{center}
    \begin{small}
    \begin{sc}
        \begin{tabular}{lcccr}
            \toprule
            \textbf{Prompt (MAE, MSE)} & \textbf{Naturalness} & \textbf{Comprehensiveness} & \textbf{On Topic} \\
            \midrule
            \textbf{Scale 1-4} & 0.1467, 0.0458 & 0.1120, 0.0393 & 0.1018, 0.0375 \\
            \textbf{Scale 1-4 with Examples} & 0.1426, 0.0452 & 0.1067, 0.0405 & 0.1023, 0.0410 \\
            \textbf{Scale 1-5} & 0.1608, 0.0506 & 0.1204, 0.0398 & 0.1235, 0.0487 \\
            \textbf{Scale 1-5 with Examples} & 0.1492, 0.0464 & 0.1161, 0.0376 & 0.1045, 0.0401 \\
            \rowcolor{green!50}\textbf{Average} & 0.1265, 0.0310 & 0.0929, 0.0212 & 0.0934, 0.0248 \\
            \rowcolor{green!15}\textbf{Most Occurrence} & 0.1342, 0.0412 & 0.0951, 0.0333 & 0.0903, 0.0362 \\
            \textbf{LLM} & 0.1438, 0.0435 & 0.1065, 0.0378 & 0.1065, 0.0438 \\
            \bottomrule
        \end{tabular}
    \end{sc}
    \end{small}
    \end{center}
    \vskip -0.1in
\end{table}

\begin{figure}
    \centering
    \includegraphics[width=0.95\textwidth]{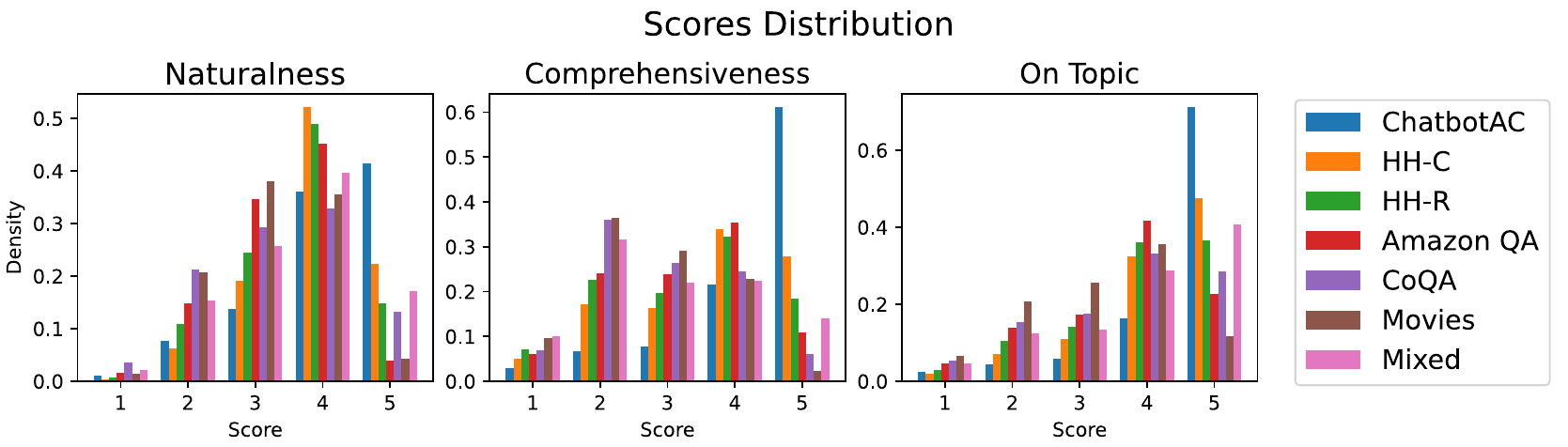}
    \caption{Conversation Evaluation Score Distribution in each datasets.}
    \label{fig:score-distribution}
\end{figure}

Each dataset consists of 15,000 conversations, with 10,000 used for training and 5,000 for evaluation or testing. As seen in Figure \ref{fig:score-distribution}, the six datasets exhibit different conversation evaluation score distributions with their own biases and variances, providing a robust setting to test the impact of explanations.

\subsection{Impact of Explanations}
\subsubsection{Assessment vs Confidence Explanations}
We investigated the relative impact of including two distinct types of explanations during fine-tuning: assessment explanations, which provide a rationale for the assigned conversational evaluation score, and confidence explanations, which explain the model's level of confidence in its score prediction. 

For example, an assessment explanation for a low naturalness score might state: "The response is unnatural due to repetitive phrasing and a mismatch in tone compared to the user's request." A confidence explanation could be: "I am highly confident this response is unnatural, as the issues are obvious and severe." 

\begin{figure}
    \centering
    \includegraphics[width=\textwidth]{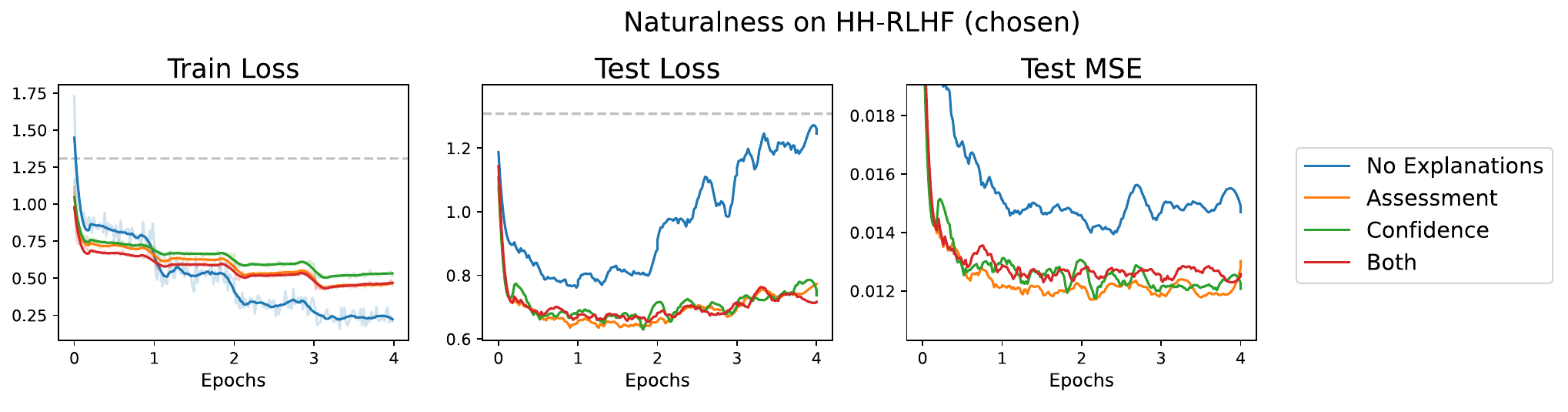}
    \caption{Performance of fine-tuning with different types of explanations - no explanations, assessment explanations, confidence explanations, and both assessment and confidence explanations.}
    \label{fig:type-of-explanations}
\end{figure}

As shown in Figure \ref{fig:type-of-explanations}, models fine-tuned with either type of explanation consistently outperformed those trained without any explanations across all conversational dimensions. Interestingly, the performance of models fine-tuned on assessment explanations, confidence explanations, and both types of explanations together was largely similar. 

A potential explanation for why assessment and confidence explanations yield comparable results is that both types of explanations are fundamentally aimed at rationalizing the same underlying conversational scores. As such, there is likely significant overlap in the language and reasoning used in the assessment explanations, which justify the assigned scores, and the confidence explanations, which explain the model's certainty in those scores for a given dialogue context. This shared lexical and semantic information may allow the model to effectively learn the task regardless of which specific type of explanation is provided during fine-tuning.

\subsubsection{With vs Without Explanations}
\begin{figure}
    \centering
    \includegraphics[width=\textwidth]{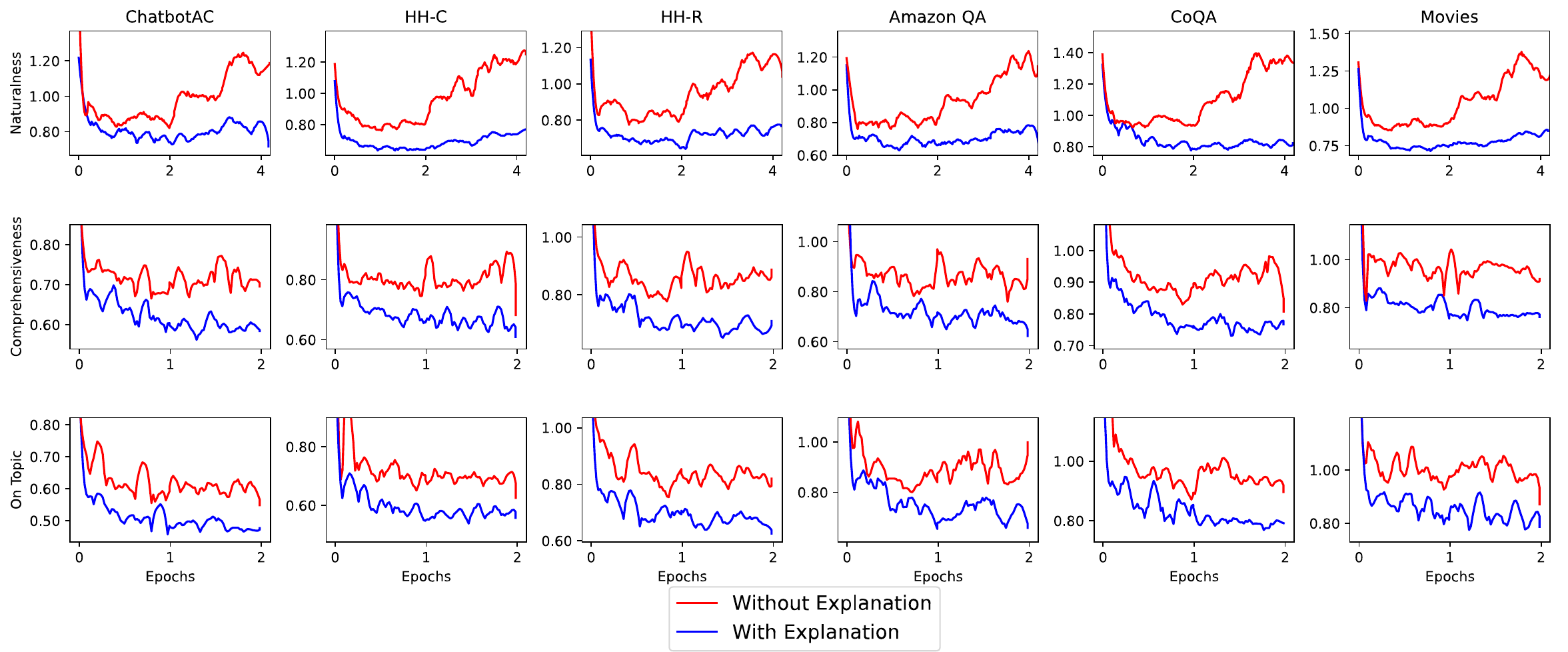}
    \caption{Loss on Test set during the fine-tuning. Full plots for fine-tuning are present in Appendix \ref{section:appendix-with-without-explanations}}
    \label{fig:finetuning-loss}
\end{figure}

Our core finding is that fine-tuning language models with explanations leads to improved performance compared to models trained solely on conversational scores. This improvement is evident from the lower test loss achieved by models fine-tuned with explanations, as shown in Figure \ref{fig:finetuning-loss}. This phenomenon is consistently observed across all six datasets and the three conversational dimensions of Naturalness, Comprehensiveness, and On-Topic.

Furthermore, models trained with explanations demonstrate greater robustness and resistance to overfitting. In contrast, models trained without explanations often start to overfit after the first epoch, leading to an increase in test loss, as seen in Figure \ref{fig:finetuning-loss}. However, the test loss remains stable for models trained with explanations across multiple epochs, indicating their ability to generalize better and resist overfitting.

These results validate our hypothesis that incorporating explanations during fine-tuning can lead to more accurate and robust language models for evaluating conversational quality. In the following sections, we analyze potential mechanisms that could underlie this phenomenon.

%% file: sections/50-analysis.tex
\section{Analysis}
\label{section:analysis}
\FloatBarrier

We perform a comprehensive analysis to understand the mechanisms underlying the performance gains observed when fine-tuning with explanations. Our analysis spans multiple aspects, including cross-dataset generalization, the effect of explanation length, the impact of random token substitutions in explanations, differences in model weights induced by fine-tuning with explanations, and entropy comparisons between models at each transformer layer.

\subsection{Cross-Dataset Performance}
\begin{table}[h]
    \caption{Naturalness Cross Dataset Performance of Fine-Tuned Models after single epoch using Mean Squared Error (lower is better). Note: Baseline is Mistral with no fine-tuning. Cross-Dataset Performance for comprehensiveness and on-topic are shown in Appendix \ref{section:appendix-cross-dataset-performance}}
    \label{table:naturalness-cross-dataset-performance}
    \begin{center}
    \begin{small}
    \begin{sc}
        \begin{tabular}{llllllll}
            \toprule
            \textbf{MSE} & \textbf{Golden} & \textbf{ChatbotAC} & \textbf{HH-C} & \textbf{HH-R} & \textbf{AmazonQA} & \textbf{CoQA} & \textbf{Movies} \\ 
            \midrule
            \textbf{Baseline} & \textcolor{red}{0.3321} & \textcolor{red}{0.0569} & \textcolor{red}{0.0395} & \textcolor{red}{0.0526} & \textcolor{red}{0.0588} & \textcolor{red}{0.1337} & \textcolor{red}{0.1186} \\ 
            \textbf{ChatbotAC} & 0.0160 & 0.0192 & 0.0168 & 0.0168 & 0.0163 & 0.0305 & 0.0168 \\ 
            \textbf{ChatbotAC (E)} & 0.0143 & 0.0164 & 0.0149 & 0.0155 & 0.0171 & 0.0221 & 0.0178 \\ 
            \textbf{HH-C} & 0.0151 & 0.0244 & 0.0146 & 0.0160 & 0.0176 & 0.0451 & 0.0176 \\ 
            \textbf{HH-C (E)} & 0.0148 & 0.0179 & 0.0123 & 0.0127 & 0.0160 & 0.0381 & \textcolor{red}{0.0226} \\ 
            \textbf{HH-R} & 0.0147 & 0.0227 & 0.0154 & 0.0179 & 0.0180 & 0.0294 & 0.0167 \\ 
            \textbf{HH-R (E)} & 0.0121 & 0.0185 & 0.0125 & 0.0136 & 0.0153 & 0.0250 & 0.0155 \\ 
            \textbf{AmazonQA} & 0.0254 & 0.0295 & 0.0212 & 0.0217 & 0.0129 & 0.0235 & 0.0354 \\ 
            \textbf{AmazonQA (E)} & 0.0157 & 0.0227 & 0.0157 & 0.0170 & 0.0119 & 0.0216 & 0.0182 \\ 
            \textbf{CoQA} & 0.0196 & 0.0283 & 0.0227 & 0.0220 & 0.0176 & 0.0211 & 0.0341 \\ 
            \textbf{CoQA (E)} & 0.0169 & 0.0240 & 0.0179 & 0.0165 & 0.0171 & 0.0211 & 0.0166 \\ 
            \textbf{Movies} & 0.0200 & 0.0314 & 0.0213 & 0.0263 & 0.0385 & 0.0362 & 0.0182 \\ 
            \textbf{Movies (E)} & 0.0133 & 0.0236 & 0.0151 & 0.0177 & 0.0162 & 0.0191 & 0.0129 \\ 
            \textbf{Mixed} & 0.0175 & 0.0184 & 0.0160 & 0.0179 & 0.0169 & 0.0194 & 0.0192 \\ 
            \textbf{Mixed (E)} & 0.0117 & 0.0126 & 0.0113 & 0.0118 & 0.0117 & 0.0139 & 0.0138 \\ 
            \bottomrule
        \end{tabular}
    \end{sc}
    \end{small}
    \end{center}
    \vskip -0.1in
\end{table}

Table \ref{table:naturalness-cross-dataset-performance} shows the cross-dataset performance of fine-tuned models on the naturalness dimension after a single epoch. Models fine-tuned with explanations (denoted by (E)) consistently outperform those trained without explanations across all test datasets. This trend of improved performance with explanations holds true even when a model is evaluated on a dataset that differs from its fine-tuning dataset, except for one model, the model fine-tuned on HH-C dataset with explanations performs slightly worse on Movies dialogue dataset, than the model fine-tuned on HH-C dataset without explanations. This demonstrates strong generalization capabilities gained through explanation-based training. Additionally, all fine-tuned models outperform the baseline Mistral model with no fine-tuning.

The cross-dataset performance for the comprehensiveness and on-topic dimensions is provided in Appendix \ref{section:appendix-cross-dataset-performance}. Similar trends are observed, with models fine-tuned with explanations consistently outperforming those trained without explanations across most test datasets.

\subsection{Explanation Length}
\textbf{Mixed Dataset:} The combined dataset, referred to as the "mixed dataset," was created from the previous six individual datasets (ChatbotAC, HH-C, HH-R, AmazonQA, CoQA, and Movies), where explanations longer than 150 words. An equal number of data points were included from each of the six datasets to ensure a diverse and representative sample. The dataset consists of 15,000 conversations, with 10,000 used for training and 5,000 for evaluation or testing.

\begin{figure}
    \centering
    \includegraphics[width=\textwidth]{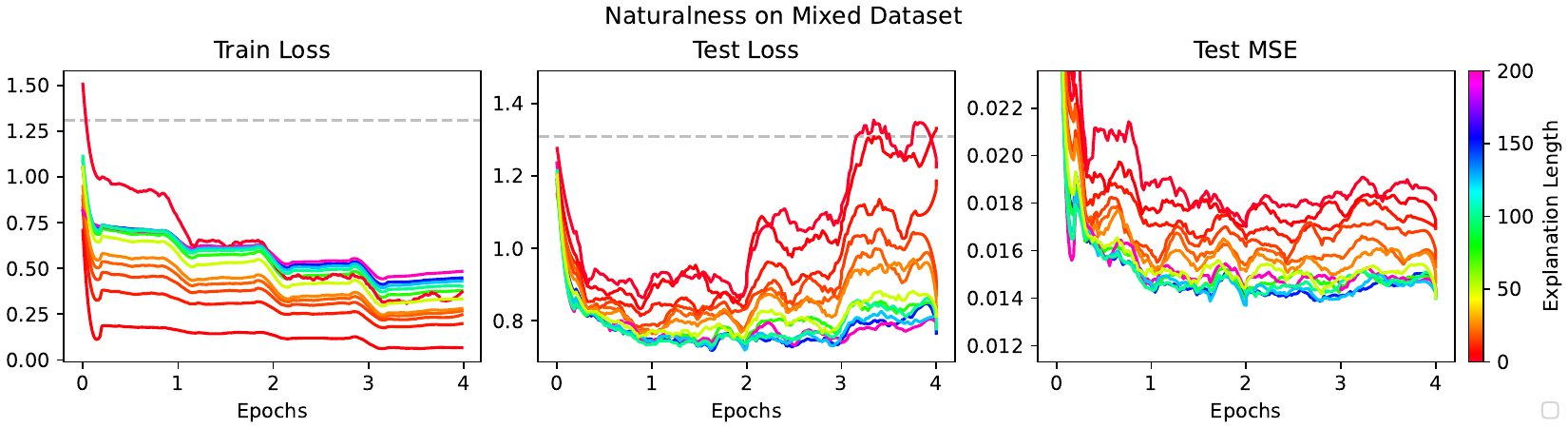}
    \caption{Effect of length of explanations on the model.}
    \label{fig:el}
\end{figure}

We investigate the influence of explanation length by fine-tuning models on the mixed dataset with explanations truncated to different word counts, ranging from 0 to 150 words (Figure \ref{fig:el}). Our results indicate that models can effectively learn from relatively short explanations of 25 to 50 words while still achieving performance comparable to models trained on full-length explanations. As the number of words in the explanations increases further, we observe diminishing returns in the improvement of performance.

\begin{figure}
    \centering
    \includegraphics[width=\textwidth]{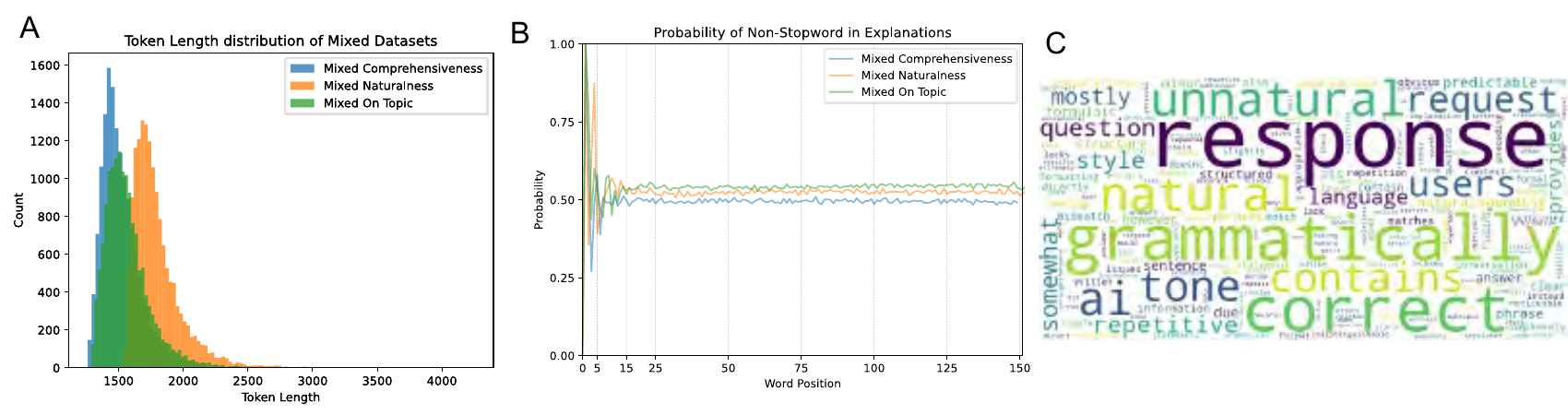}
    \caption{\textbf{A.} The histogram of the total token length of the data point combined with prompt and explanations in mixed datasets. \textbf{B.} The probability of a non-stopword, indicating the likelihood of a meaningful word at a given position in the explanations.  \textbf{C.} A word cloud displaying the most frequent initial 25 words in the mixed naturalness dataset explanations. }
    \label{fig:explanation-length-words}
\end{figure}

From Figure \ref{fig:explanation-length-words}A, we can observe that all token lengths are well within the context length limit of 8192 tokens for the Mistral 7B model, ruling out the possibility of context length limitations. Figure \ref{fig:explanation-length-words}B shows that after the first few words, the probability of encountering a meaningful (non-stopword) token stabilizes, suggesting a consistent use of meaningful content throughout the explanations, with no particular importance given to the first 25 words. Interestingly, Figure \ref{fig:explanation-length-words}C reveals that the explanations tend to begin with conclusions or assessments of naturalness, often mentioning the AI response and its characteristics. This overall summary at the beginning of the explanations gives the model all the important context needed for effective fine-tuning within the first 25 words.


\subsection{Random Token}

Table \ref{table:naturalness-random-dataset-performance} shows the performance of models fine-tuned on datasets where the explanations in the mixed dataset have been replaced by various types of random token substitutions:

\begin{itemize}[leftmargin=*]
    \item \textbf{Fully Random:} Random dictionary words.
    \item \textbf{Shuffle Random:} Words from all unique non-stopwords from all explanations in the mixed dataset.
    \item \textbf{Associated Random:} Words from all unique non-stopwords in the explanations where the score matches the given score.
    \item \textbf{Weighted Shuffle Random:} Weighted version of Shuffle Random, where the frequency of the word determines its probability of being used in the explanation.
    \item \textbf{Weighted Associated Random:} Weighted version of Associated Random, where the frequency of the word determines its probability of being used in the explanation.
\end{itemize}

\begin{table}[h]
    \caption{Naturalness performance on random datasets after a single epoch of fine-tuning, evaluated using Mean Squared Error (lower is better). Baseline refers to the Mistral model without fine-tuning. \textbf{W} prefix denotes the weighted version of the corresponding dataset. Fine-tuning plots for these random datasets are provided in Appendix \ref{section:appendix-random-dataset}.}
    \label{table:naturalness-random-dataset-performance}.
    \begin{center}
    \begin{small}
    \begin{sc}
        \begin{tabular}{llllllll}
            \toprule
            \textbf{MSE $\downarrow$} & \textbf{Golden} & \textbf{ChatbotAC} & \textbf{HH-C} & \textbf{HH-R} & \textbf{AmazonQA} & \textbf{CoQA} & \textbf{Movies} \\ 
            \midrule
            \rowcolor{red!30}\textbf{Baseline} & 0.3321 & 0.0569 & 0.0395 & 0.0526 & 0.0588 & 0.1337 & 0.1186 \\ 
            \textbf{Mixed} & \cellcolor{red!15}0.0175 & \cellcolor{red!15}0.0184 & \cellcolor{red!15}0.0160 & \cellcolor{red!15}0.0179 & \cellcolor{red!15}0.0169 & 0.0194 & 0.0192 \\
            \rowcolor{green!50}\textbf{Mixed (E)} & 0.0117 & 0.0126 & 0.0113 & 0.0118 & 0.0117 & 0.0139 & 0.0138 \\ 
            \textbf{Fully} & 0.0128 & \cellcolor{red!15}0.0184 & 0.0135 & 0.0169 & 0.0162 & \cellcolor{red!15}0.0202 & \cellcolor{red!15}0.0194 \\
            \textbf{Shuffle} & \cellcolor{green!25}0.0118 & 0.0170 & 0.0130 & 0.0150 & 0.0148 & 0.0184 & 0.0181 \\ 
            \textbf{Associated} & 0.0125 & \cellcolor{green!25}0.0160 & \cellcolor{green!25}0.0129 & \cellcolor{green!25}0.0147 & \cellcolor{green!25}0.0142 & \cellcolor{green!25}0.0174 & \cellcolor{green!25}0.0172 \\ 
            \textbf{W-Shuffle} & 0.0141 & 0.0171 & 0.0142 & 0.0155 & 0.0148 & 0.0183 & 0.0179 \\ 
            \textbf{W-Associated} & 0.0134 & 0.0167 & 0.0141 & 0.0157 & 0.0147 & 0.0184 & 0.0180 \\ 
            \bottomrule
        \end{tabular}
    \end{sc}
    \end{small}
    \end{center}
    \vskip -0.1in
\end{table}

As seen in Table \ref{table:naturalness-random-dataset-performance}, models fine-tuned with the original explanations consistently outperformed those trained with random tokens across all random substitution types. This suggests that the benefits of explanation-based fine-tuning stem not only from the presence of additional tokens but also from the specific content and structure of the explanations themselves.

The baseline model without any fine-tuning always performed the worst. But an intriguing observation emerges when comparing the performance of models fine-tuned with random token substitutions for explanations against the model fine-tuned without explanations. All random substitution types, except for Fully Random, consistently outperformed the model fine-tuned without explanations, while the Fully Random approach showed similar or slightly worse performance on several datasets (ChatbotAC, CoQA, and Movies). This suggests that completely random tokens provide limited benefit. A potential explanation for this counter-intuitive finding is that the presence of random tokens, when combined with the model's inherent complexities such as dropout layers, may act as a form of noise regularization. This phenomenon is analogous to the observed improvements in the performance of computer vision models when trained on noise-augmented images. By introducing a degree of structured random noise during the fine-tuning process, the model may become more robust to overfitting and better equipped to generalize to unseen data.

While the inclusion of random tokens provided performance gains over the model fine-tuned without explanations, models fine-tuned with the original, coherent explanations consistently outperformed their randomly fine-tuned counterparts. This finding underscores the importance of the specific content and structure of the explanations in enabling effective knowledge transfer during the fine-tuning process.

The models fine-tuned on the Associated Random and Shuffle Random datasets exhibited better performance compared to those trained on the Fully Random dataset. This observation suggests that the presence of words from the original explanations, even in a random order, provides a measurable benefit to the fine-tuning process. Furthermore, the Associated Random dataset outperformed the Shuffle Random dataset, also indicating the association between conversation evaluation scores and their corresponding explanations. This result aligns with the definitions and guidelines provided in the prompt for each score, which establish a clear connection between the assigned scores and the rationale behind them.

However, the models trained on the Weighted Associated Random and Weighted Shuffle Random datasets did not demonstrate significant improvements over their unweighted counterparts. This may be due to the fact that some high-frequency words are repeated often in the weighted explanations, hence not really helping with the training, and even hinder the process due to their own repetitiveness and creating a bias in the gradients.

This analysis opens up the possibility of more dataset augmentation techniques for language models, similar to those employed in computer vision when dataset size is small or limited. By introducing carefully crafted noise or random tokens during fine-tuning, models may become more robust and better able to generalize to unseen data.

Additional details, including the mean absolute error (MAE) and fine-tuning plots for the random datasets, are provided in Appendix \ref{section:appendix-random-dataset}.

\subsection{LoRA Weight Differences}
\begin{table}[h]
    \caption{Normalized LoRA Weight Differences between fine-tuning with and without explanations using the Frobenius norm. Additional raw data is available in the Appendix \ref{section:appendix-lora-weight-diffrences}.}
    \label{table:lora-weight-diffrences}
    \begin{center}
    \begin{small}
    \begin{sc}
        \begin{tabular}{lllllllll}
            \toprule
            ~ & ~ & \textbf{L-Down} & \textbf{L-Gate} & \textbf{L-Up} & \textbf{Attn-K} & \textbf{Attn-O} & \textbf{Attn-Q} & \textbf{Attn-V} \\ 
            \midrule
            \parbox[t]{2mm}{\multirow{6}{*}{\rotatebox[origin=c]{90}{Naturalness}}} & \textbf{ChatbotAC} & 0.6313 & \cellcolor{green!30}0.8356 & \cellcolor{green!15}0.8250 & 0.7707 & 0.6575 & \cellcolor{green!50}1.0000 & 0.6342 \\ 
            ~ & \textbf{HH-C} & 0.7215 & \cellcolor{green!50}1.0000 & \cellcolor{green!30}0.9825 & \cellcolor{green!15}0.9567 & 0.8152 & 0.8944 & 0.7235 \\ 
            ~ & \textbf{HH-R} & 0.6051 & \cellcolor{green!30}0.8145 & \cellcolor{green!15}0.7958 & 0.7536 & 0.6548 & \cellcolor{green!50}1.0000 & 0.6073 \\ 
            ~ & \textbf{Amazon QA} & 0.5837 & \cellcolor{green!30}0.8566 & \cellcolor{green!15}0.8391 & 0.6859 & 0.6693 & \cellcolor{green!50}1.0000 & 0.6026 \\ 
            ~ & \textbf{CoQA} & 0.4960 & \cellcolor{green!30}0.6821 & \cellcolor{green!15}0.6748 & 0.5927 & 0.5744 & \cellcolor{green!50}1.0000 & 0.5397 \\ 
            ~ & \textbf{Movies} & 0.5705 & \cellcolor{green!30}0.8186 & \cellcolor{green!15}0.7891 & 0.7517 & 0.6392 & \cellcolor{green!50}1.0000 & 0.5955 \\ 

            \parbox[t]{2mm}{\multirow{7}{*}{\rotatebox[origin=c]{90}{Comprehensive\;\;\:}}}  \\
            ~ & \textbf{ChatbotAC} & 0.6826 & \cellcolor{green!50}1.0000 & \cellcolor{green!15}0.9649 & 0.7701 & 0.8347 & \cellcolor{green!30}0.9855 & 0.7831 \\ 
            ~ & \textbf{HH-C} & 0.6743 & \cellcolor{green!30}0.9968 & \cellcolor{green!50}1.0000 & 0.7844 & 0.8287 & \cellcolor{green!15}0.9212 & 0.7190 \\ 
            ~ & \textbf{HH-R} & 0.6936 & \cellcolor{green!50}1.0000 & \cellcolor{green!30}0.9806 & 0.8273 & 0.8348 & \cellcolor{green!15}0.9520 & 0.6866 \\ 
            ~ & \textbf{Amazon QA} & 0.6566 & \cellcolor{green!30}0.9671 & \cellcolor{green!15}0.9637 & 0.7247 & 0.8309 & \cellcolor{green!50}1.0000 & 0.7252 \\ 
            ~ & \textbf{CoQA} & 0.5605 & \cellcolor{green!30}0.8043 & \cellcolor{green!15}0.7958 & 0.7034 & 0.6981 & 1\cellcolor{green!50}.0000 & 0.6246 \\ 
            ~ & \textbf{Movies} & 0.6093 & \cellcolor{green!30}0.9165 & \cellcolor{green!15}0.8808 & 0.7297 & 0.7549 & \cellcolor{green!50}1.0000 & 0.6846 \\ 
            
            ~ \\
            \parbox[t]{2mm}{\multirow{6}{*}{\rotatebox[origin=c]{90}{On-Topic}}} & \textbf{ChatbotAC} & 0.6843 & \cellcolor{green!30}0.9903 & \cellcolor{green!50}1.0000 & 0.8746 & \cellcolor{green!15}0.9211 & 0.8875 & 0.8376 \\ 
            ~ & \textbf{HH-C} & 0.6721 & \cellcolor{green!30}0.9902 & \cellcolor{green!50}1.0000 & 0.8355 & 0.8662 & \cellcolor{green!15}0.9569 & 0.7422 \\ 
            ~ & \textbf{HH-R} & 0.7194 & \cellcolor{green!30}0.9854 & \cellcolor{green!15}0.9836 & 0.8191 & 0.8655 & \cellcolor{green!50}1.0000 & 0.7310 \\ 
            ~ & \textbf{T Amazon QA} & 0.5211 & \cellcolor{green!30}0.7629 & \cellcolor{green!15}0.7504 & 0.6541 & 0.6490 & \cellcolor{green!50}1.0000 & 0.5712 \\ 
            ~ & \textbf{CoQA} & 0.5152 & \cellcolor{green!30}0.7382 & \cellcolor{green!15}0.7271 & 0.6917 & 0.6578 & \cellcolor{green!50}1.0000 & 0.6287 \\ 
            ~ & \textbf{Movies} & 0.6777 & \cellcolor{green!30}0.9496 & \cellcolor{green!15}0.9298 & 0.7712 & 0.8252 & \cellcolor{green!50}1.0000 & 0.7143 \\
            
            \bottomrule
        \end{tabular}
    \end{sc}
    \end{small}
    \end{center}
    \textsuperscript{*}\textbf{L:} MLP Layer, \textbf{Attn:} Attention Layer, \textbf{K:} Key, \textbf{O:} Ouput, \textbf{Q:} Query, \textbf{V:} Value
    \vskip -0.1in
\end{table}

Table \ref{table:lora-weight-diffrences} presents the normalized differences in LoRA weights between models fine-tuned with and without explanations, using the Frobenius norm. Our analysis of attention weight patterns and hidden representations \citep{danilevsky2023opening} reveals that the largest differences occur in the Query projection of attention layers, followed by the MLP Gate projection and Up projections. This suggests that explanations during fine-tuning primarily impact the model's ability to attend to relevant input tokens and modulate the flow of information through the network's gating mechanisms. The MLP layers, responsible for non-linear transformations, also exhibit substantial differences, particularly in the Gate projections that control the information flow. The Key and Value projections, responsible for encoding the input representations, exhibit smaller but non-negligible differences, indicating that explanations also influence how the model perceives and encodes input data. The Output projection, which combines attention outputs, shows relatively minor differences, implying that final output generation is less affected by explanation presence during fine-tuning.

The consistency of which layers are changing can be seen even after fine-tuning with only 200 data points, and the difference grows as training progresses. The raw data for the LoRA weight differences is available in Appendix \ref{section:appendix-lora-weight-diffrences}.

\subsection{Entropy}
In the earlier transformer blocks (1-20 and 21-25), the entropy differences between the models fine-tuned with and without explanations are relatively small, indicating that both models exhibit similar levels of uncertainty in the initial stages of processing. However, as the transformer blocks progress, the entropy differences become more pronounced. The model fine-tuned with explanations generally exhibits higher entropy in the later blocks (26-31), suggesting that it is considering multiple possibilities and has not yet reached a certain conclusion.

In the final transformer block (32), the entropy differences are consistently negative across all datasets, with a few exceptions. This implies that the model fine-tuned with explanations exhibits lower entropy at this final stage, indicating more confident predictions compared to the model trained without explanations.

The higher entropy observed in the later transformer blocks for the model fine-tuned with explanations suggests that it is exploring multiple options and considering diverse perspectives before converging to a more confident prediction in the final block. In contrast, the model trained without explanations appears to reach its conclusions earlier, potentially missing out on considering alternative possibilities.

The raw entropy data and output token rankings for each transformer block are provided in Appendices \ref{section:appendix-entropy-ouput-token} and \ref{section:appendix-ouput-token-ranking}, respectively, allowing for a more detailed analysis of the entropy patterns and token distributions across the transformer layers.

\begin{table}[h]
    \caption{Normalized Entropy Differences of Output Logits after Each Transformer Block in Mistral 7B. Positive values indicate higher entropy for the model fine-tuned with explanations, while negative values denote higher entropy for the model trained without explanations. Additional data, including raw entropy data and output token rankings, are provided in Appendices \ref{section:appendix-entropy-ouput-token} and \ref{section:appendix-ouput-token-ranking}, respectively. Note: The entropy for the transformer block 1-20 and 21-25 are averaged.}
    \label{table:entropy}
    \begin{center}
    \begin{small}
    \begin{sc}
        \begin{tabular}{lllllllllll}
            \toprule
            ~ & \textbf{Dataset $\downarrow$ / Block $\rightarrow$} & \textbf{1-20} & \textbf{21-25} & \textbf{26} & \textbf{27} & \textbf{28} & \textbf{29} & \textbf{30} & \textbf{31} & \textbf{32} \\ 
            \midrule
            
            \parbox[t]{2mm}{\multirow{6}{*}{\rotatebox[origin=c]{90}{Naturalness}}} & \textbf{ChatbotAC} & 0.02 & 0.13 & 0.20 & 0.19 & 0.27 & 0.42 & 0.59 & 1.00 & \cellcolor{green!30}-0.07 \\ 
            ~ & \textbf{HH-C} & 0.01 & 0.14 & 0.25 & 0.30 & 0.37 & 0.56 & 0.71 & 1.00 & \cellcolor{green!30}-0.06 \\ 
            ~ & \textbf{HH-R} & 0.01 & 0.14 & 0.25 & 0.30 & 0.35 & 0.54 & 0.68 & 1.00 & \cellcolor{green!30}-0.01 \\ 
            ~ & \textbf{Amazon QA} & 0.01 & 0.13 & 0.22 & 0.27 & 0.29 & 0.44 & 0.63 & 1.00 & \cellcolor{green!30}-0.15 \\ 
            ~ & \textbf{CoQA} & 0.01 & 0.15 & 0.27 & 0.30 & 0.37 & 0.52 & 0.70 & 1.00 & \cellcolor{green!30}-0.03 \\ 
            ~ & \textbf{Movies} & 0.02 & 0.16 & 0.24 & 0.26 & 0.29 & 0.48 & 0.69 & 1.00 & \cellcolor{green!30}-0.14 \\ 
            
            \parbox[t]{2mm}{\multirow{7}{*}{\rotatebox[origin=c]{90}{Comprehensive\;\;\:}}}  \\
            ~ & \textbf{ChatbotAC} & 0.02 & 0.13 & 0.23 & 0.24 & 0.33 & 0.51 & 0.64 & 1.00 & \cellcolor{red!30}0.06 \\ 
            ~ & \textbf{Amazon QA} & 0.02 & 0.11 & 0.19 & 0.24 & 0.31 & 0.46 & 0.66 & 1.00 & \cellcolor{green!30}-0.10 \\ 
            ~ & \textbf{HH-C} & 0.02 & 0.15 & 0.23 & 0.26 & 0.33 & 0.49 & 0.67 & 1.00 & \cellcolor{green!30}-0.05 \\ 
            ~ & \textbf{HH-R} & 0.02 & 0.15 & 0.25 & 0.30 & 0.38 & 0.53 & 0.67 & 1.00 & \cellcolor{green!30}-0.07 \\ 
            ~ & \textbf{CoQA} & 0.02 & 0.16 & 0.23 & 0.27 & 0.35 & 0.46 & 0.64 & 1.00 & \cellcolor{green!30}-0.01 \\ 
            ~ & \textbf{Movies} & 0.02 & 0.15 & 0.27 & 0.30 & 0.37 & 0.51 & 0.73 & 1.00 & \cellcolor{green!30}-0.03 \\ 
            
            ~ \\
            \parbox[t]{2mm}{\multirow{6}{*}{\rotatebox[origin=c]{90}{On-Topic}}} & \textbf{ChatbotAC} & 0.01 & 0.08 & 0.16 & 0.17 & 0.26 & 0.41 & 0.59 & 1.00 & \cellcolor{green!30}-0.06 \\ 
            ~ & \textbf{HH-C} & 0.02 & 0.12 & 0.19 & 0.21 & 0.29 & 0.42 & 0.56 & 1.00 & \cellcolor{green!30}-0.03 \\ 
            ~ & \textbf{HH-R} & 0.01 & 0.11 & 0.19 & 0.22 & 0.27 & 0.44 & 0.58 & 1.00 & \cellcolor{green!30}-0.15 \\ 
            ~ & \textbf{Amazon QA} & 0.01 & 0.12 & 0.23 & 0.27 & 0.34 & 0.48 & 0.67 & 1.00 & \cellcolor{green!30}-0.01 \\ 
            ~ & \textbf{CoQA} & 0.01 & 0.12 & 0.24 & 0.24 & 0.32 & 0.48 & 0.64 & 1.00 & \cellcolor{red!30}0.01 \\ 
            ~ & \textbf{Movies} & 0.01 & 0.15 & 0.27 & 0.30 & 0.36 & 0.51 & 0.66 & 1.00 & \cellcolor{green!30}-0.09 \\ 
            
            \bottomrule
        \end{tabular}
    \end{sc}
    \end{small}
    \end{center}
    \vskip -0.1in
\end{table}

%% file: sections/60-discussion.tex
\section{Implications and Discussion}
\label{section:discussion}
\FloatBarrier

\subsection{Key Implications}
\begin{itemize}[leftmargin=*]
    \item \textbf{Improved Model Performance and Robustness:} Incorporating explanations during fine-tuning leads to improved performance across multiple datasets and conversational quality dimensions. Models trained with explanations exhibit strong generalization capabilities and increased resistance to overfitting.

    \item \textbf{Insights into Model Behavior and Learning Dynamics:} 
    \begin{itemize}
        \item LoRA weight analysis reveals significant changes in attention and gating mechanisms, impacting the model's ability to attend to relevant tokens and modulate information flow.
        \item Entropy analysis suggests models fine-tuned with explanations consider multiple possibilities before converging, while models without explanations reach conclusions earlier.
    \end{itemize}
    
    \item \textbf{Potential for Dataset Augmentation and Noise Regularization:} Performance gains observed with random token substitutions in explanations suggest the possibility of using carefully crafted noise or random tokens as a form of regularization, enhancing robustness and generalization.

    \item \textbf{Potential for Improved Interpretability and Explainability:} Incorporating human-provided explanations may align model representations and decision-making processes with human reasoning, increasing transparency and trust in model decisions.
\end{itemize}

\subsection{Discussion and Future Work}

Our findings demonstrate the significant benefits of incorporating explanations during the fine-tuning of large language models for conversational quality evaluation tasks. By providing the model with llm-generated rationales and reasoning, we enable more effective knowledge transfer and enhance the model's ability to understand and assess critical dimensions like naturalness, comprehensiveness, and topic relevance.

Moreover, the performance gains observed with random token substitutions in explanations open up intriguing possibilities for dataset augmentation and noise regularization techniques. By introducing carefully crafted noise or random tokens during fine-tuning, we may be able to enhance the model's robustness and generalization capabilities, especially in data-scarce scenarios.

Future work could explore more advanced approaches to dataset augmentation, such as combining random text with coherent explanations or introducing noise into positional encodings. 

Another promising direction is the application of explanation-based fine-tuning to other domains and tasks beyond conversational quality evaluation. For instance, incorporating explanations during fine-tuning for tasks like question answering, summarization, or natural language inference could potentially yield similar benefits.

Furthermore, a deeper analysis of the entropy patterns and token distributions across transformer layers could provide valuable insights into the model's decision-making processes. Understanding how explanations influence the model's attention and information flow at different stages of processing may uncover opportunities for architectural improvements or more targeted fine-tuning strategies.

Finally, as the field of large language models continues to advance, exploring the scalability and generalizability of explanation-based fine-tuning to even larger and more capable models will be of great interest. The potential impact of this approach on the interpretability, robustness, and trustworthiness of future AI systems should not be underestimated.

%% file: sections/70-conclusion.tex
\section{Conclusion}
\label{section:conclusion}

In this work, we have demonstrated that incorporating explanations during the fine-tuning process of large language models consistently improves their performance on classification tasks. Through extensive experimentation across three different conversational quality assessment tasks and six distinct datasets, we found that models fine-tuned with both class labels and explanations consistently outperform those trained on labels alone. This improvement was observed regardless of the specific task or dataset, suggesting that the benefits of explanation-based fine-tuning are robust and generalizable.

Our investigation into the mechanisms behind this improvement revealed several key insights. First, the finding that even random word sequences serving as explanations led to improved performance suggests that the benefits of explanation-based fine-tuning extend beyond the semantic content of the explanations themselves. This observation points to a potential regularization effect, where the presence of additional tokens in the fine-tuning objective helps prevent overfitting on label tokens and encourages more robust learning.

Analysis of model internals provided further evidence for how explanations influence model behavior. The higher entropy observed in intermediate layers of explanation-trained models, coupled with more focused predictions in final layers, suggests that these models engage in more thorough deliberation before making decisions. This pattern of activation may explain why explanation-trained models achieve better performance: they appear to consider a broader range of possibilities before converging on their final predictions.

These findings have important implications for the development and deployment of language models in classification tasks. They suggest that practitioners should consider incorporating explanations in their fine-tuning protocols, even in cases where explanations are not required in the final deployment. The consistent improvements we observed across different tasks and datasets indicate that this approach could be valuable in a wide range of applications.

However, our work also raises several interesting questions for future research. For instance, investigating how the length and structure of explanations affect model performance could lead to more optimized training protocols. Additionally, exploring whether these findings generalize to other types of tasks beyond conversational quality assessment could expand the applicability of our approach.

In conclusion, our research provides strong evidence for the benefits of explanation-based fine-tuning in improving the performance of large language models on classification tasks. By demonstrating both the practical advantages and providing insights into the underlying mechanisms, this work contributes to our understanding of how to develop more effective and reliable language model classifiers. As the field continues to evolve, we believe that incorporating explanations during fine-tuning will become an increasingly important tool in the development of robust and accurate language models.

%% file: sections/99-appendix.tex
\section{Appendix}
\label{section:appendix}

\FloatBarrier
\subsection{With vs Without Explanations}
\label{section:appendix-with-without-explanations}
\FloatBarrier
\begin{figure}[!h]
    \centering
    \includegraphics[width=\textwidth]{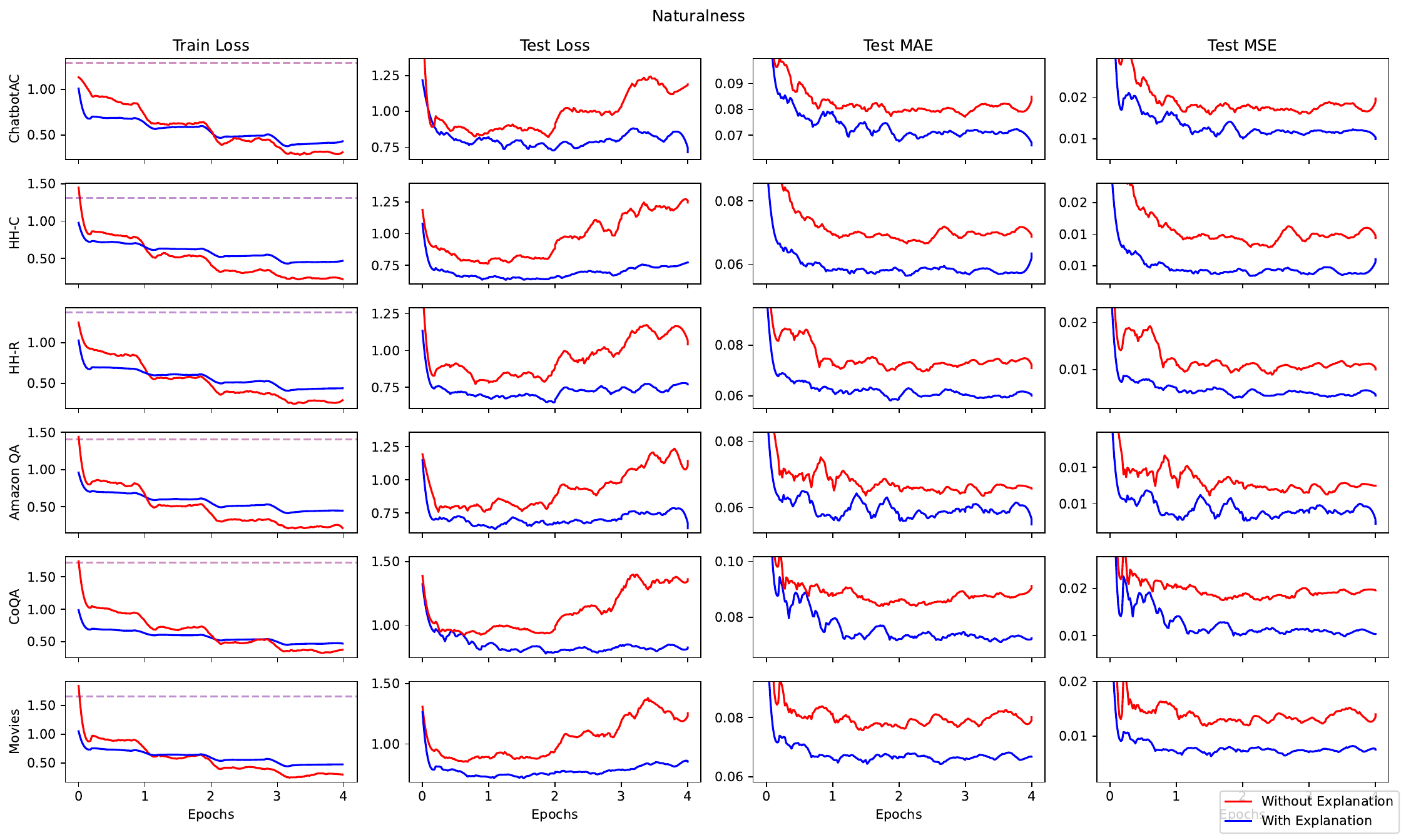}
    \caption{Naturalness fine-tuning with and without Explanations}
\end{figure}
\begin{figure}[!h]
    \centering
    \includegraphics[width=\textwidth]{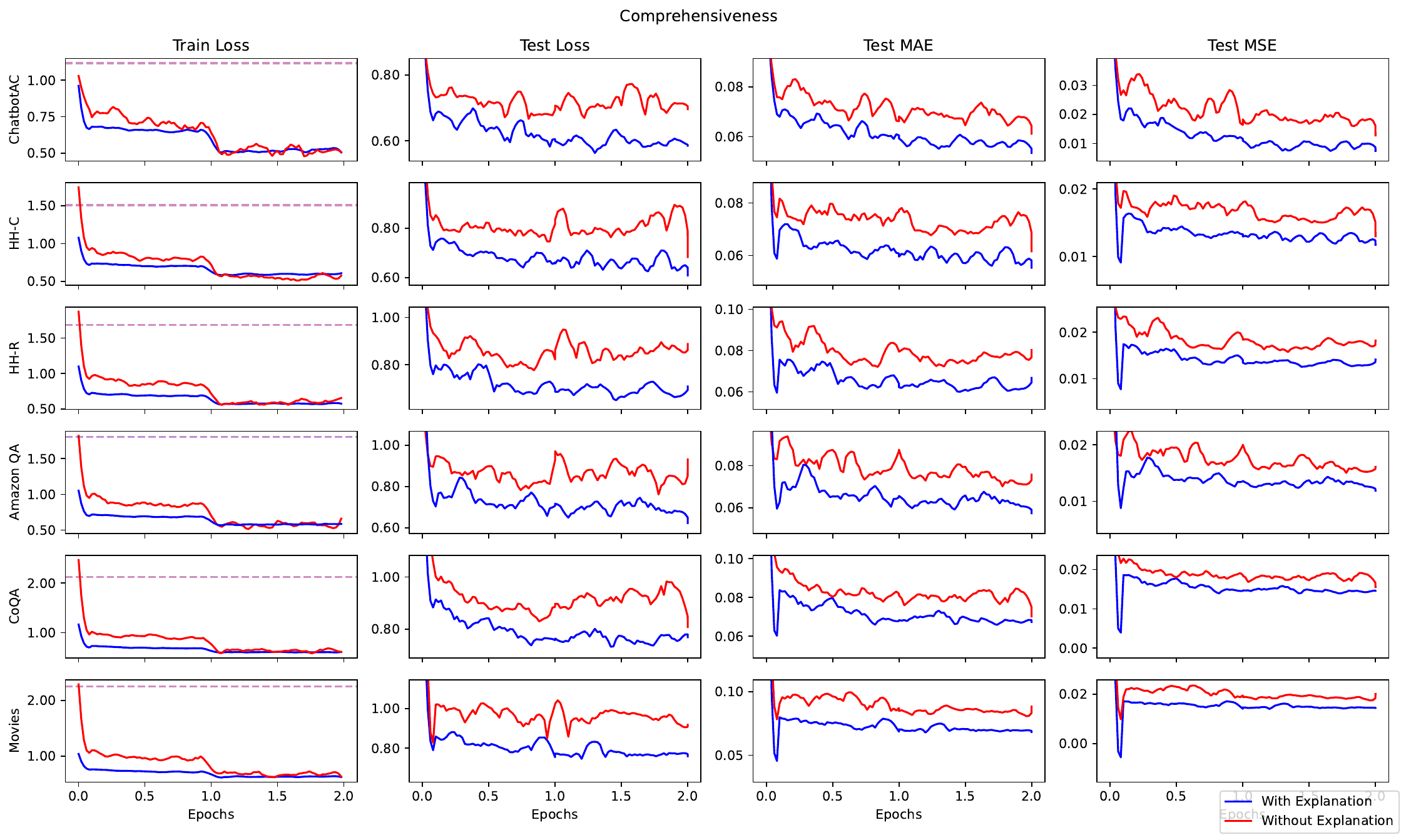}
    \caption{Comprehensiveness fine-tuning with and without Explanations}
\end{figure}
\begin{figure}[!h]
    \centering
    \includegraphics[width=\textwidth]{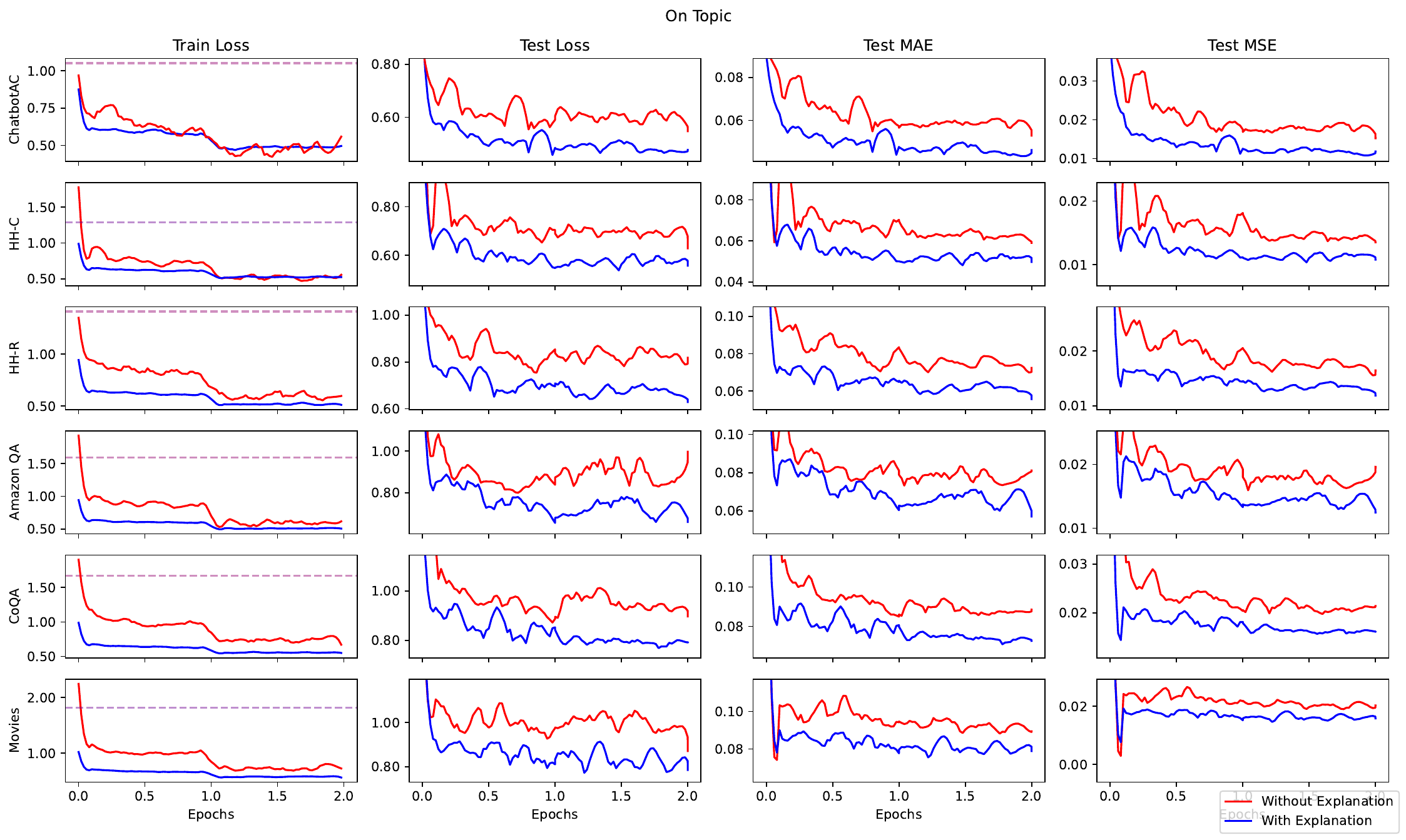}
    \caption{On-Topic fine-tuning with and without Explanations}
\end{figure}

\FloatBarrier
\subsection{Cross Dataset Performance}
\label{section:appendix-cross-dataset-performance}
\FloatBarrier

\begin{table}[!h]
    \caption{Naturalness Cross Dataset Performance of Fine-Tuned Models after single epoch using Mean Absolute Error (lower is better). Note: Baseline is Mistral with no fine-tuning.}
    \vskip 0.15in
    \begin{center}
    \begin{small}
    \begin{sc}
        \begin{tabular}{llllllll}
            \toprule
            \textbf{MAE $\downarrow$} & \textbf{Golden} & \textbf{ChatbotAC} & \textbf{HH-C} & \textbf{HH-R} & \textbf{AmazonQA} & \textbf{CoQA} & \textbf{Movies} \\ 
            \midrule
            \textbf{Baseline} & 0.2929 & 0.1574 & 0.1351 & 0.1582 & 0.1747 & 0.3054 & 0.2891 \\ 
            \textbf{ChatbotAC} & 0.0758 & 0.0824 & 0.0775 & 0.0764 & 0.0752 & 0.1185 & 0.0783 \\ 
            \textbf{ChatbotAC (E)} & 0.0709 & 0.0752 & 0.0707 & 0.0722 & 0.0799 & 0.0950 & 0.0824 \\ 
            \textbf{HH-C} & 0.0648 & 0.0959 & 0.0674 & 0.0729 & 0.0805 & 0.1570 & 0.0822 \\ 
            \textbf{HH-C (E)} & 0.0652 & 0.0778 & 0.0578 & 0.0601 & 0.0759 & 0.1406 & 0.0998 \\ 
            \textbf{HH-R} & 0.0668 & 0.0928 & 0.0718 & 0.0812 & 0.0830 & 0.1194 & 0.0782 \\ 
            \textbf{HH-R (E)} & 0.0587 & 0.0818 & 0.0592 & 0.0644 & 0.0717 & 0.1042 & 0.0730 \\ 
            \textbf{AmazonQA} & 0.1206 & 0.1274 & 0.0940 & 0.0943 & 0.0621 & 0.1062 & 0.1397 \\ 
            \textbf{AmazonQA (E)} & 0.0754 & 0.1025 & 0.0739 & 0.0775 & 0.0580 & 0.0990 & 0.0845 \\ 
            \textbf{CoQA} & 0.0876 & 0.1130 & 0.1008 & 0.0979 & 0.0836 & 0.0888 & 0.1371 \\ 
            \textbf{CoQA (E)} & 0.0811 & 0.1047 & 0.0838 & 0.0778 & 0.0830 & 0.0945 & 0.0787 \\ 
            \textbf{Movies} & 0.0929 & 0.1285 & 0.0906 & 0.1034 & 0.1354 & 0.1407 & 0.0829 \\ 
            \textbf{Movies (E)} & 0.0648 & 0.1036 & 0.0711 & 0.0789 & 0.0743 & 0.0867 & 0.0627 \\ 
            \textbf{Mixed} & 0.0754 & 0.0808 & 0.0715 & 0.0794 & 0.0765 & 0.0841 & 0.0834 \\ 
            \textbf{Mixed (E)} & 0.0570 & 0.0599 & 0.0556 & 0.0583 & 0.0571 & 0.0655 & 0.0654 \\ 
            \bottomrule
        \end{tabular}
    \end{sc}
    \end{small}
    \end{center}
    \vskip -0.1in
\end{table}

\FloatBarrier

\begin{table}[!h]
    \caption{Comprehensiveness Cross Dataset Performance of Fine-Tuned Models after single epoch using Mean Squared Error (lower is better). Note: Baseline is Mistral with no fine-tuning.}
    \vskip 0.15in
    \begin{center}
    \begin{small}
    \begin{sc}
        \begin{tabular}{llllllll}
            \toprule
            \textbf{MSE $\downarrow$} & \textbf{Golden} & \textbf{ChatbotAC} & \textbf{HH-C} & \textbf{HH-R} & \textbf{AmazonQA} & \textbf{CoQA} & \textbf{Movies} \\ 
            \midrule
            \textbf{Baseline} & 0.3275 & 0.0573 & 0.1128 & 0.1457 & 0.1514 & 0.2127 & 0.2440 \\ 
            \textbf{ChatbotAC} & 0.0194 & 0.0198 & 0.0230 & 0.0298 & 0.0278 & 0.0478 & 0.0365 \\ 
            \textbf{ChatbotAC (E)} & 0.0177 & 0.0170 & 0.0181 & 0.0202 & 0.0191 & 0.0282 & 0.0248 \\ 
            \textbf{HH-C} & 0.0159 & 0.0286 & 0.0163 & 0.0229 & 0.0229 & 0.0352 & 0.0251 \\ 
            \textbf{HH-C (E)} & 0.0121 & 0.0206 & 0.0126 & 0.0152 & 0.0142 & 0.0190 & 0.0169 \\ 
            \textbf{HH-R} & 0.0160 & 0.0258 & 0.0162 & 0.0198 & 0.0199 & 0.0254 & 0.0203 \\ 
            \textbf{HH-R (E)} & 0.0130 & 0.0178 & 0.0131 & 0.0139 & 0.0139 & 0.0204 & 0.0166 \\ 
            \textbf{Amazon QA} & 0.0178 & 0.0257 & 0.0221 & 0.0283 & 0.0206 & 0.0361 & 0.0364 \\ 
            \textbf{Amazon QA (E)} & 0.0133 & 0.0200 & 0.0206 & 0.0265 & 0.0143 & 0.0248 & 0.0232 \\ 
            \textbf{CoQA} & 0.0245 & 0.0345 & 0.0341 & 0.0341 & 0.0267 & 0.0197 & 0.0546 \\ 
            \textbf{CoQA (E)} & 0.0239 & 0.0290 & 0.0250 & 0.0228 & 0.0220 & 0.0154 & 0.0294 \\ 
            \textbf{Movies} & 0.0174 & 0.0258 & 0.0268 & 0.0341 & 0.0361 & 0.0285 & 0.0181 \\ 
            \textbf{Movies (E)} & 0.0138 & 0.0208 & 0.0202 & 0.0264 & 0.0240 & 0.0172 & 0.0147 \\ 
            \textbf{Mixed} & 0.0151 & 0.0192 & 0.0164 & 0.0168 & 0.0177 & 0.0186 & 0.0183 \\ 
            \textbf{Mixed (E)} & 0.0147 & 0.0151 & 0.0146 & 0.0145 & 0.0144 & 0.0151 & 0.0150 \\ 
            \bottomrule
        \end{tabular}
    \end{sc}
    \end{small}
    \end{center}
    \vskip -0.1in
\end{table}

\FloatBarrier

\begin{table}[!h]
    \caption{Comprehensiveness Cross Dataset Performance of Fine-Tuned Models after single epoch using Mean Absolute Error (lower is better). Note: Baseline is Mistral with no fine-tuning.}
    \vskip 0.15in
    \begin{center}
    \begin{small}
    \begin{sc}
        \begin{tabular}{llllllll}
            \toprule
            \textbf{MAE $\downarrow$} & \textbf{Golden} & \textbf{ChatbotAC} & \textbf{HH-C} & \textbf{HH-R} & \textbf{AmazonQA} & \textbf{CoQA} & \textbf{Movies} \\ 
            \midrule
            \textbf{Baseline} & 0.3079 & 0.1244 & 0.2480 & 0.3024 & 0.3226 & 0.4097 & 0.4519 \\ 
            \textbf{ChatbotAC} & 0.0774 & 0.0689 & 0.0897 & 0.1069 & 0.1064 & 0.1542 & 0.1300 \\ 
            \textbf{ChatbotAC (E)} & 0.0754 & 0.0641 & 0.0766 & 0.0828 & 0.0851 & 0.1070 & 0.1018 \\ 
            \textbf{HH-C} & 0.0631 & 0.0822 & 0.0691 & 0.0864 & 0.0931 & 0.1279 & 0.0997 \\ 
            \textbf{HH-C (E)} & 0.0530 & 0.0679 & 0.0570 & 0.0671 & 0.0666 & 0.0834 & 0.0770 \\ 
            \textbf{HH-R} & 0.0660 & 0.0789 & 0.0699 & 0.0816 & 0.0842 & 0.1025 & 0.0884 \\ 
            \textbf{HH-R (E)} & 0.0538 & 0.0623 & 0.0600 & 0.0628 & 0.0647 & 0.0863 & 0.0756 \\ 
            \textbf{AmazonQA} & 0.0725 & 0.0776 & 0.0860 & 0.1026 & 0.0882 & 0.1342 & 0.1308 \\ 
            \textbf{AmazonQA (E)} & 0.0574 & 0.0687 & 0.0824 & 0.0984 & 0.0671 & 0.1015 & 0.0959 \\ 
            \textbf{CoQA} & 0.1055 & 0.1360 & 0.1334 & 0.1311 & 0.1096 & 0.0846 & 0.1846 \\ 
            \textbf{CoQA (E)} & 0.1055 & 0.1133 & 0.1066 & 0.0984 & 0.0963 & 0.0701 & 0.1180 \\ 
            \textbf{Movies} & 0.0725 & 0.0895 & 0.1054 & 0.1242 & 0.1345 & 0.1153 & 0.0822 \\ 
            \textbf{Movies (E)} & 0.0619 & 0.0724 & 0.0867 & 0.1040 & 0.1040 & 0.0819 & 0.0707 \\ 
            \textbf{Mixed} & 0.0664 & 0.0789 & 0.0734 & 0.0748 & 0.0776 & 0.0779 & 0.0781 \\ 
            \textbf{Mixed (E)} & 0.0652 & 0.0678 & 0.0667 & 0.0668 & 0.0669 & 0.0681 & 0.0681 \\ 
            \bottomrule
        \end{tabular}
    \end{sc}
    \end{small}
    \end{center}
    \vskip -0.1in
\end{table}

\FloatBarrier

\begin{table}[!h]
    \caption{On-Topic Cross Dataset Performance of Fine-Tuned Models after single epoch using Mean Squared Error (lower is better). Note: Baseline is Mistral with no fine-tuning.}
    \vskip 0.15in
    \begin{center}
    \begin{small}
    \begin{sc}
        \begin{tabular}{llllllll}
            \toprule
            \textbf{MSE $\downarrow$} & \textbf{Golden} & \textbf{ChatbotAC} & \textbf{HH-C} & \textbf{HH-R} & \textbf{AmazonQA} & \textbf{CoQA} & \textbf{Movies} \\ 
            \midrule
            \textbf{Baseline} & 0.3640 & 0.0548 & 0.0654 & 0.0821 & 0.1145 & 0.1209 & 0.1680 \\ 
            \textbf{ChatbotAC} & 0.0165 & 0.0171 & 0.0167 & 0.0245 & 0.0279 & 0.0612 & 0.0284 \\ 
            \textbf{ChatbotAC (E)} & 0.0151 & 0.0130 & 0.0156 & 0.0203 & 0.0211 & 0.0323 & 0.0265 \\ 
            \textbf{HH-C} & 0.0194 & 0.0196 & 0.0207 & 0.0256 & 0.0291 & 0.0400 & 0.0386 \\ 
            \textbf{HH-C (E)} & 0.0115 & 0.0142 & 0.0108 & 0.0130 & 0.0193 & 0.0305 & 0.0211 \\ 
            \textbf{HH-R} & 0.0216 & 0.0214 & 0.0166 & 0.0218 & 0.0299 & 0.0452 & 0.0256 \\ 
            \textbf{HH-R (E)} & 0.0130 & 0.0145 & 0.0118 & 0.0151 & 0.0149 & 0.0240 & 0.0188 \\ 
            \textbf{Amazon QA} & 0.0249 & 0.0201 & 0.0243 & 0.0287 & 0.0176 & 0.0258 & 0.0501 \\ 
            \textbf{Amazon QA (E)} & 0.0160 & 0.0172 & 0.0179 & 0.0243 & 0.0133 & 0.0283 & 0.0264 \\ 
            \textbf{CoQA} & 0.0275 & 0.0267 & 0.0292 & 0.0347 & 0.0240 & 0.0207 & 0.0479 \\ 
            \textbf{CoQA (E)} & 0.0209 & 0.0218 & 0.0201 & 0.0225 & 0.0238 & 0.0190 & 0.0512 \\ 
            \textbf{Movies} & 0.0196 & 0.0234 & 0.0229 & 0.0295 & 0.0368 & 0.0381 & 0.0220 \\ 
            \textbf{Movies (E)} & 0.0152 & 0.0177 & 0.0162 & 0.0207 & 0.0199 & 0.0208 & 0.0152 \\ 
            \textbf{Mixed} & 0.0205 & 0.0204 & 0.0161 & 0.0167 & 0.0189 & 0.0212 & 0.0217 \\ 
            \textbf{Mixed (E)} & 0.0152 & 0.0140 & 0.0120 & 0.0129 & 0.0131 & 0.0145 & 0.0150 \\ 
            \bottomrule
        \end{tabular}
    \end{sc}
    \end{small}
    \end{center}
    \vskip -0.1in
\end{table}

\FloatBarrier

\begin{table}[!h]
    \caption{On-Topic Cross Dataset Performance of Fine-Tuned Models after single epoch using Mean Absolute Error (lower is better). Note: Baseline is Mistral with no fine-tuning.}
    \vskip 0.15in
    \begin{center}
    \begin{small}
    \begin{sc}
        \begin{tabular}{llllllll}
            \toprule
            \textbf{MAE $\downarrow$} & \textbf{Golden} & \textbf{ChatbotAC} & \textbf{HH-C} & \textbf{HH-R} & \textbf{AmazonQA} & \textbf{CoQA} & \textbf{Movies} \\ 
            \midrule
            \textbf{Baseline} & 0.2941 & 0.1018 & 0.1580 & 0.1961 & 0.2589 & 0.2590 & 0.3442 \\ 
            \textbf{ChatbotAC} & 0.0668 & 0.0560 & 0.0677 & 0.0900 & 0.1025 & 0.1696 & 0.1079 \\ 
            \textbf{ChatbotAC (E)} & 0.0615 & 0.0475 & 0.0648 & 0.0801 & 0.0843 & 0.1160 & 0.1057 \\ 
            \textbf{HH-C} & 0.0725 & 0.0591 & 0.0768 & 0.0921 & 0.1059 & 0.1312 & 0.1362 \\ 
            \textbf{HH-C (E)} & 0.0509 & 0.0496 & 0.0486 & 0.0589 & 0.0816 & 0.1126 & 0.0907 \\ 
            \textbf{HH-R} & 0.0762 & 0.0645 & 0.0686 & 0.0875 & 0.1125 & 0.1421 & 0.1049 \\ 
            \textbf{HH-R (E)} & 0.0578 & 0.0512 & 0.0537 & 0.0680 & 0.0673 & 0.0937 & 0.0834 \\ 
            \textbf{AmazonQA} & 0.0847 & 0.0624 & 0.0823 & 0.0962 & 0.0758 & 0.0989 & 0.1666 \\ 
            \textbf{AmazonQA (E)} & 0.0611 & 0.0551 & 0.0651 & 0.0849 & 0.0606 & 0.1029 & 0.1032 \\ 
            \textbf{CoQA} & 0.0929 & 0.0922 & 0.1095 & 0.1225 & 0.0965 & 0.0842 & 0.1612 \\ 
            \textbf{CoQA (E)} & 0.0770 & 0.0737 & 0.0810 & 0.0900 & 0.0962 & 0.0811 & 0.1708 \\ 
            \textbf{Movies} & 0.0835 & 0.0873 & 0.0904 & 0.1078 & 0.1274 & 0.1357 & 0.0905 \\ 
            \textbf{Movies (E)} & 0.0668 & 0.0718 & 0.0700 & 0.0854 & 0.0891 & 0.0910 & 0.0716 \\ 
            \textbf{Mixed} & 0.0758 & 0.0751 & 0.0645 & 0.0685 & 0.0776 & 0.0788 & 0.0826 \\ 
            \textbf{Mixed (E)} & 0.0619 & 0.0582 & 0.0532 & 0.0582 & 0.0593 & 0.0609 & 0.0638 \\ 
            \bottomrule
        \end{tabular}
    \end{sc}
    \end{small}
    \end{center}
    \vskip -0.1in
\end{table}
\FloatBarrier

\FloatBarrier
\subsection{Random Dataset}
\label{section:appendix-random-dataset}
\FloatBarrier
\begin{figure}[!h]
    \centering
    \includegraphics[width=\textwidth]{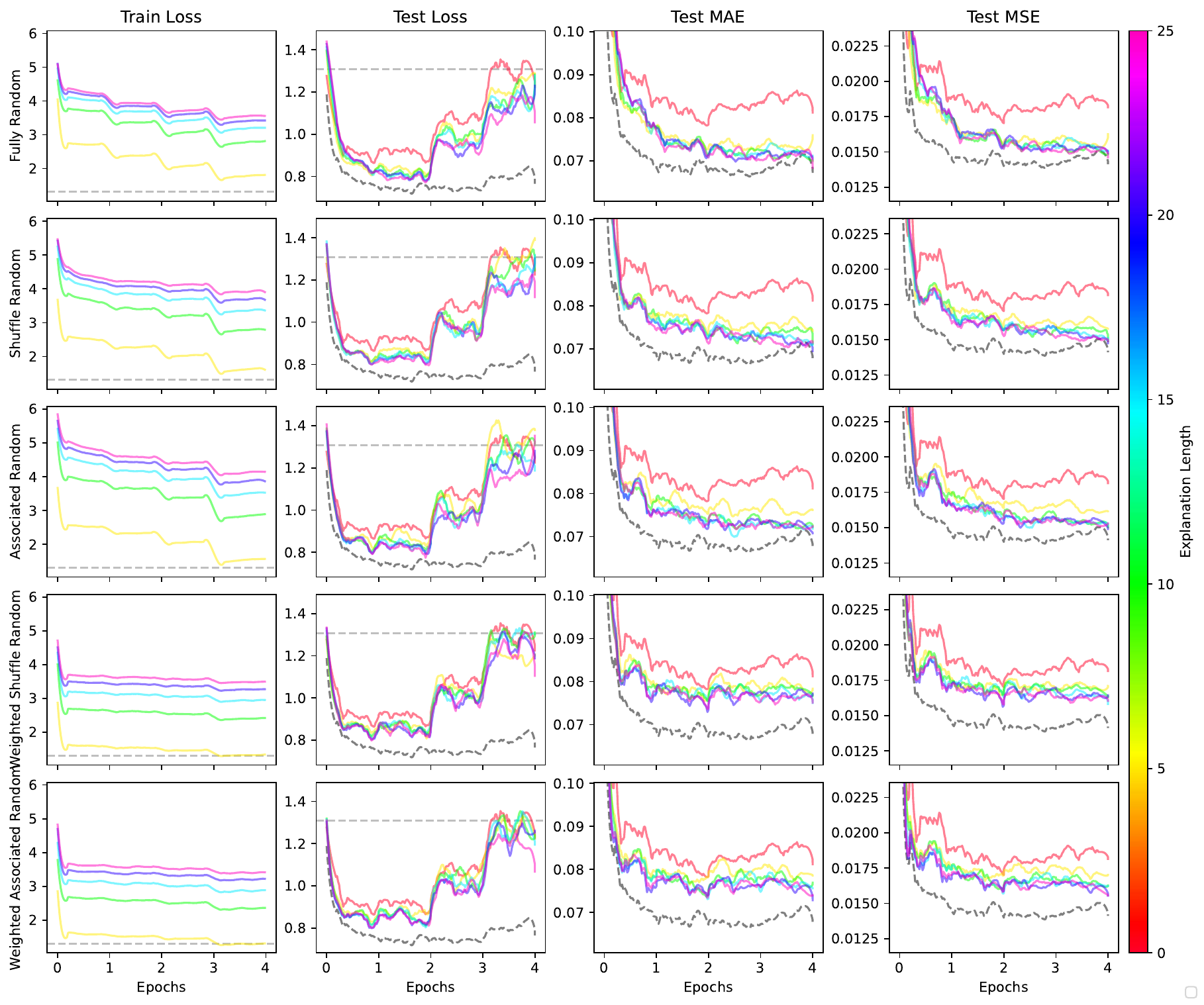}
    \caption{Fine-tuning performance on random datasets where the original explanations are replaced with different combinations of random tokens. The red line represents the baseline model fine-tuned without any explanations. The dashed black line depicts the model fine-tuned on the original, non-randomized explanations.}
\end{figure}

\FloatBarrier
\input{sections/91-appendix-lora-differences}
\FloatBarrier
\input{sections/92-appendix-rankings}
\FloatBarrier
\input{sections/93-appendix-entropy}

%% file: sections/91-appendix-lora-differences.tex
\subsection{LoRA Weight Differences}
\label{section:appendix-lora-weight-diffrences}
\FloatBarrier

\begin{figure}[!h]
    \centering
    \includegraphics[width=\textwidth]{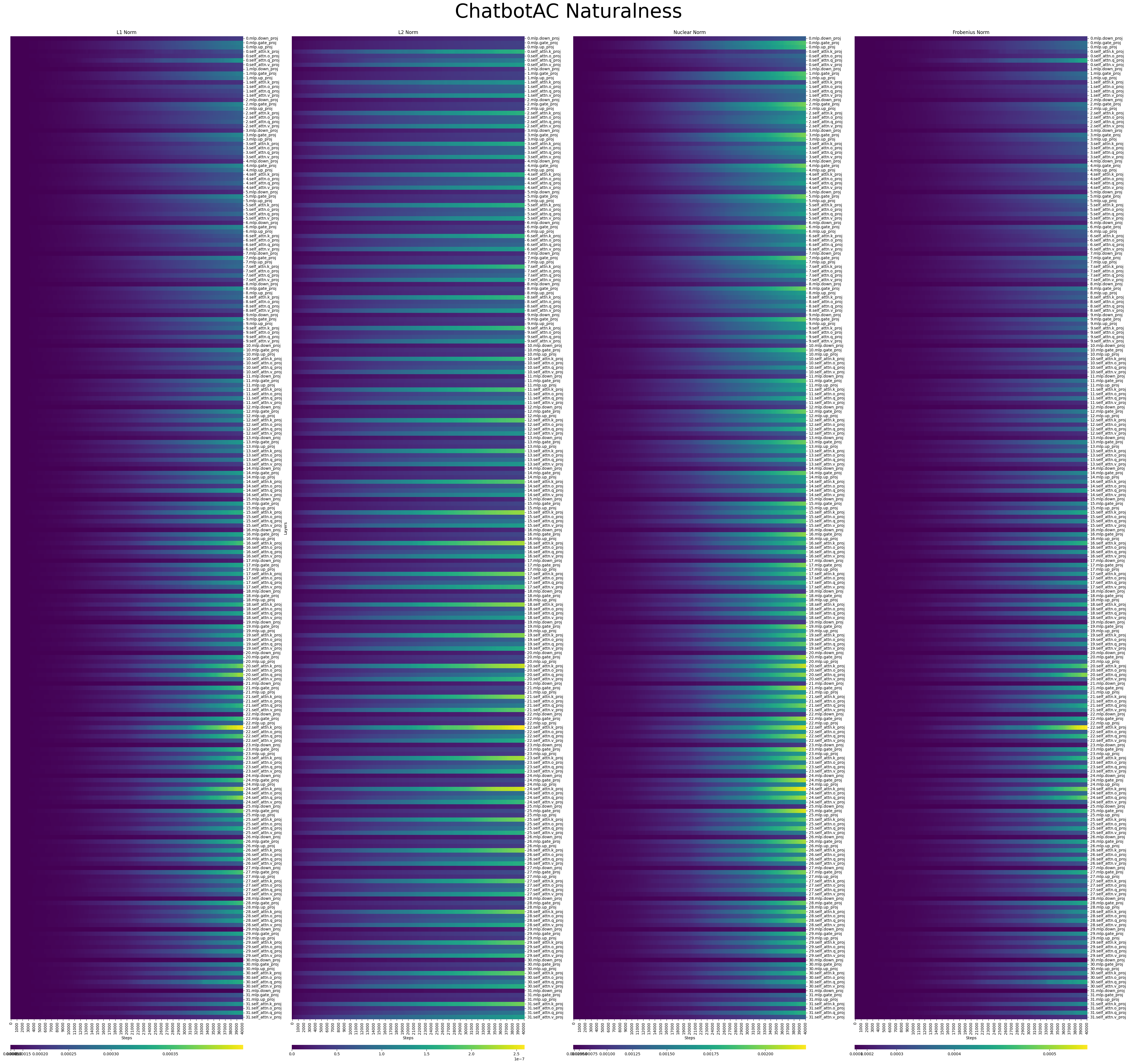}
    \caption{LoRA Weight Differences for ChatbotAC Dataset on Naturalness dimension}
\end{figure}
\begin{figure}[!h]
    \centering
    \includegraphics[width=\textwidth]{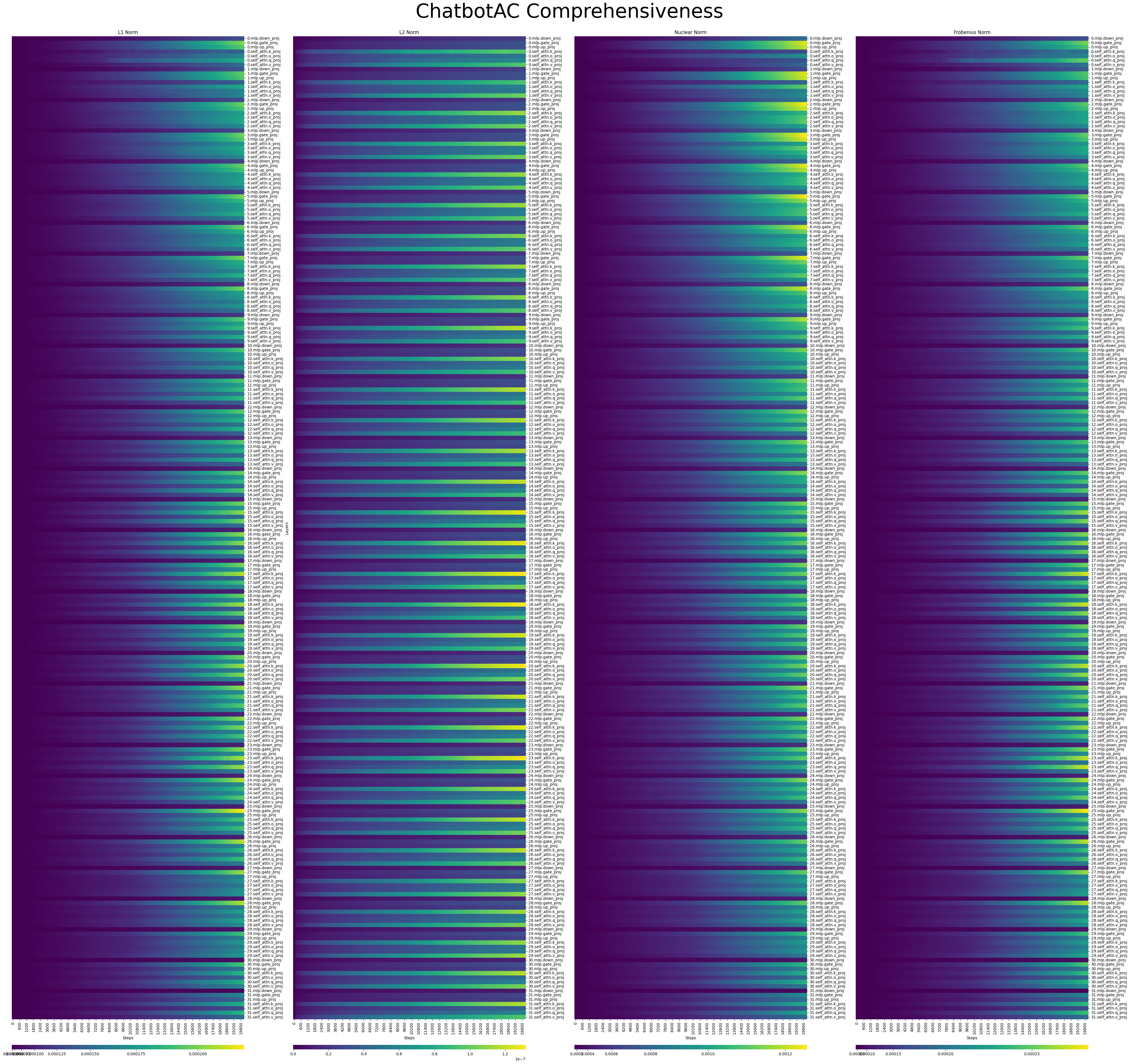}
    \caption{LoRA Weight Differences for ChatbotAC Dataset on Comprehensiveness dimension}
\end{figure}
\begin{figure}[!h]
    \centering
    \includegraphics[width=\textwidth]{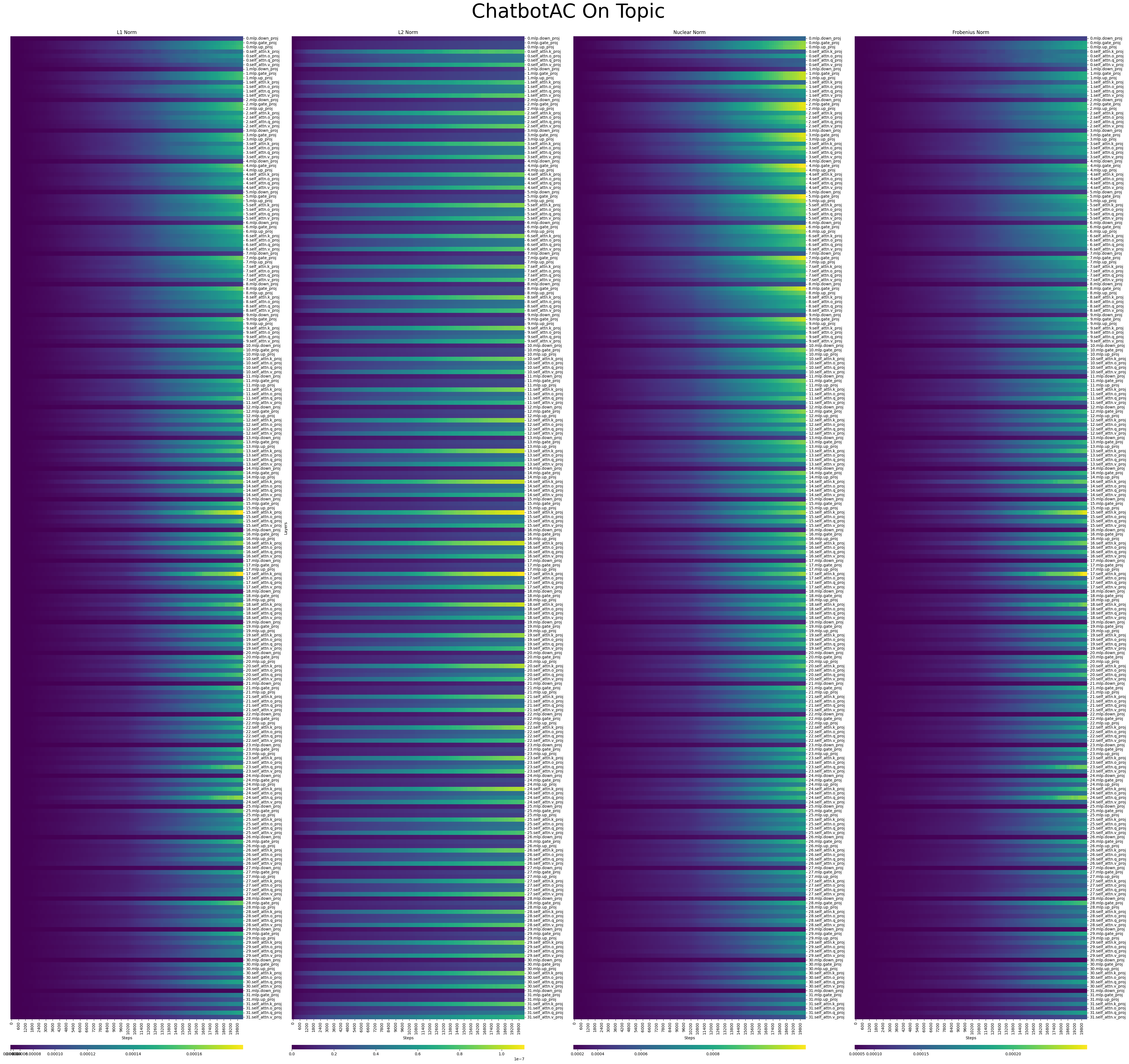}
    \caption{LoRA Weight Differences for ChatbotAC Dataset on On-Topic dimension}
\end{figure}

\begin{figure}[!h]
    \centering
    \includegraphics[width=\textwidth]{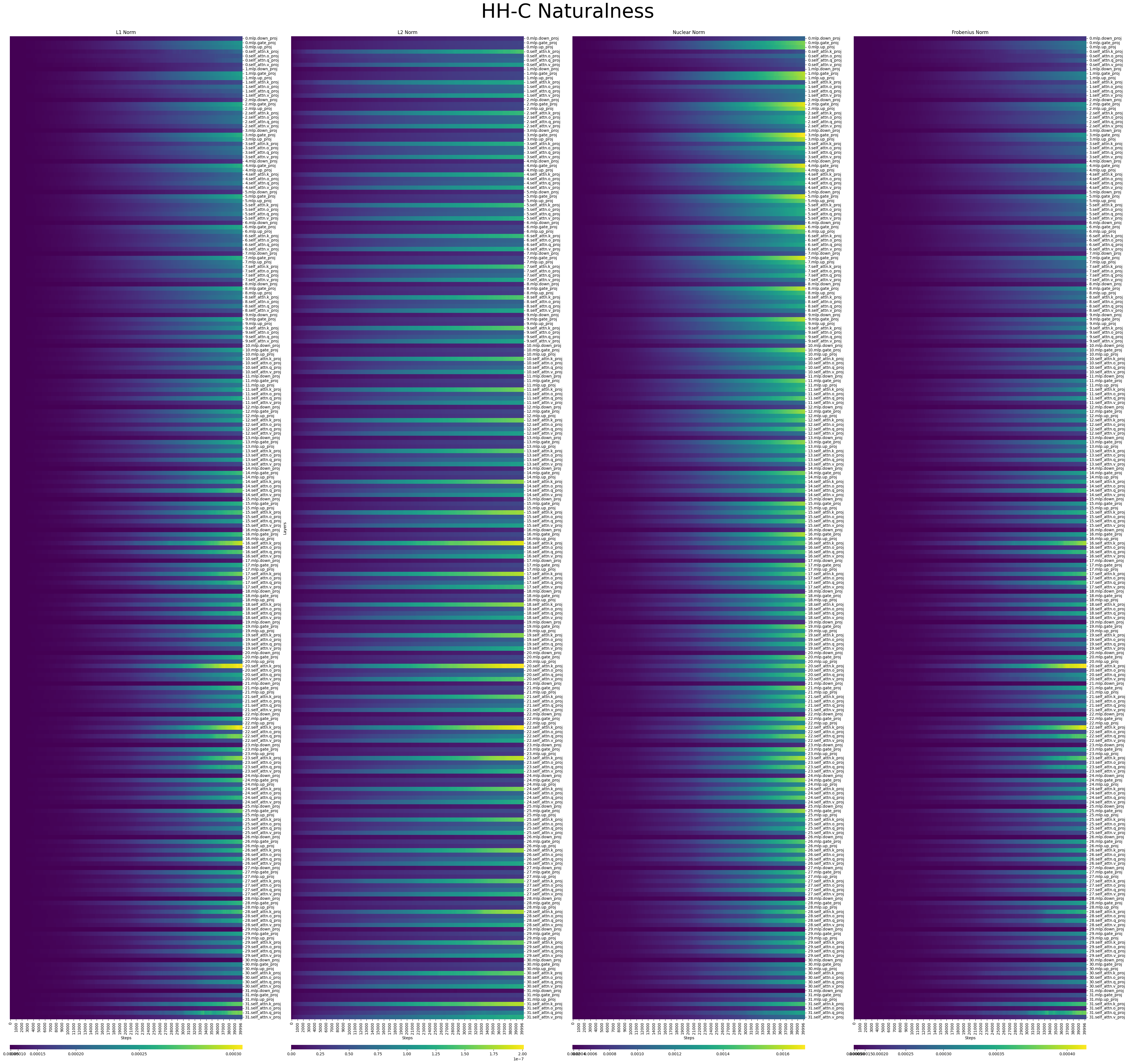}
    \caption{LoRA Weight Differences for HH-C Dataset on Naturalness dimension}
\end{figure}
\begin{figure}[!h]
    \centering
    \includegraphics[width=\textwidth]{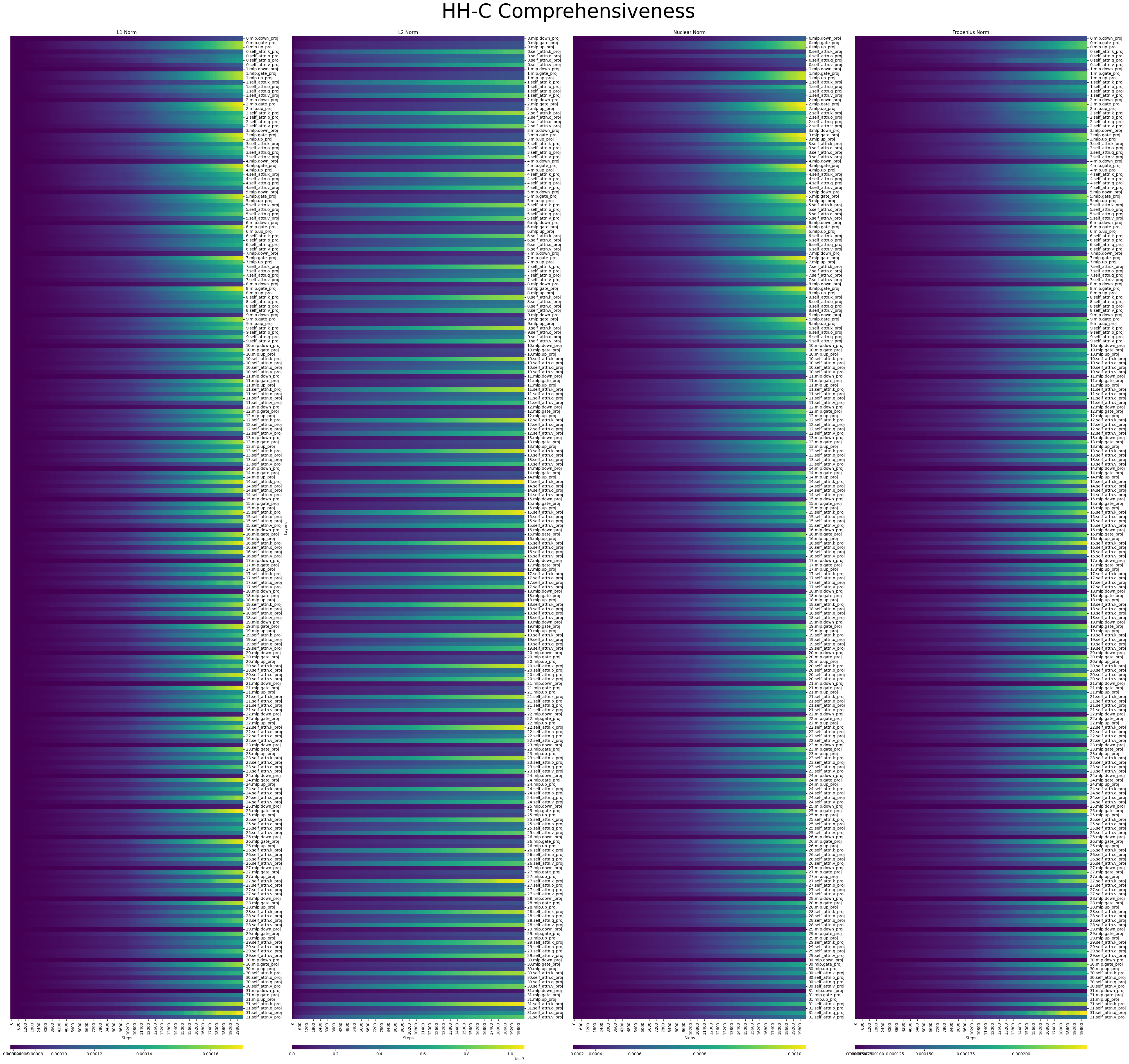}
    \caption{LoRA Weight Differences for HH-C Dataset on Comprehensiveness dimension}
\end{figure}
\begin{figure}[!h]
    \centering
    \includegraphics[width=\textwidth]{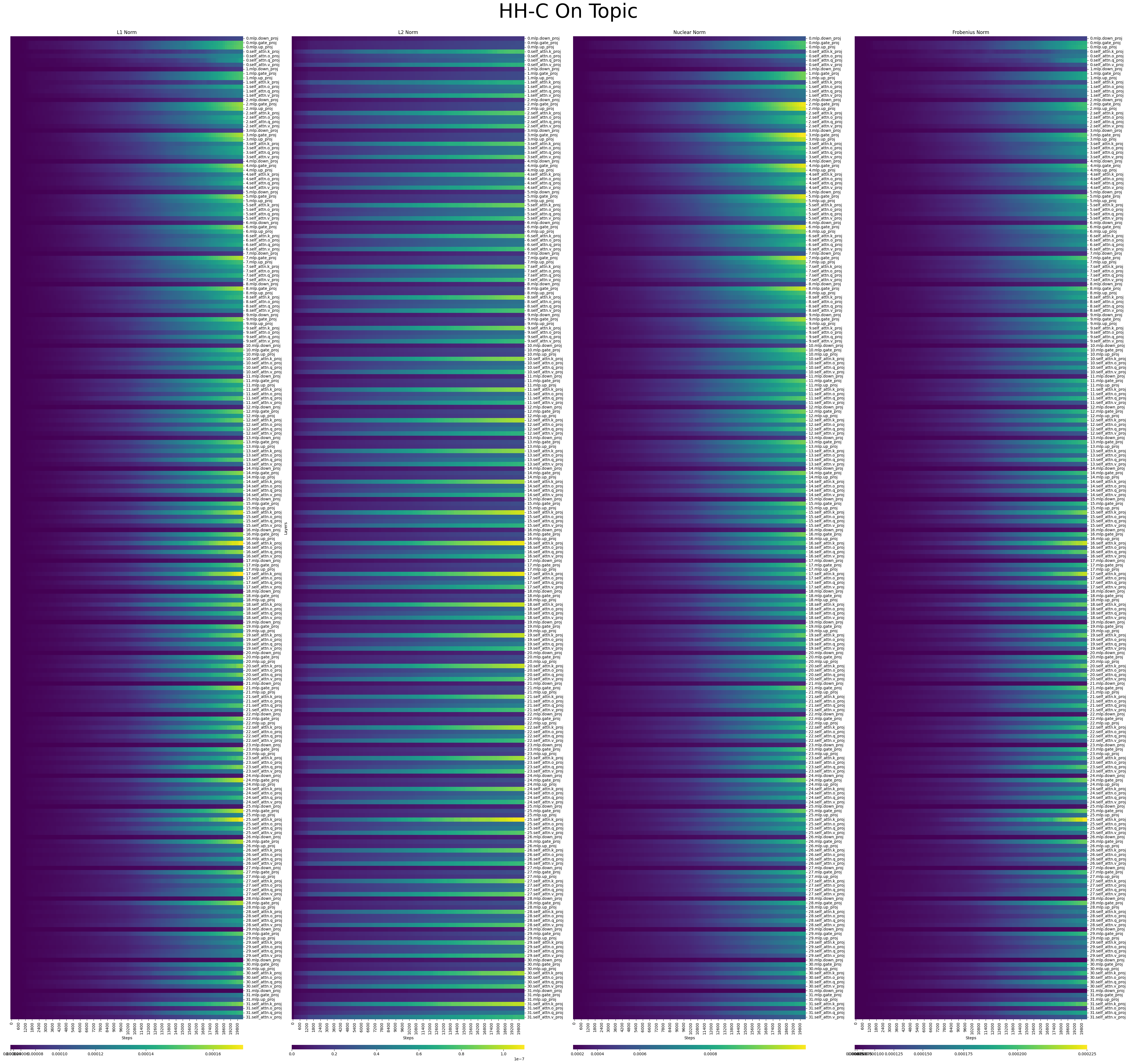}
    \caption{LoRA Weight Differences for HH-C Dataset on On-Topic dimension}
\end{figure}

\begin{figure}[!h]
    \centering
    \includegraphics[width=\textwidth]{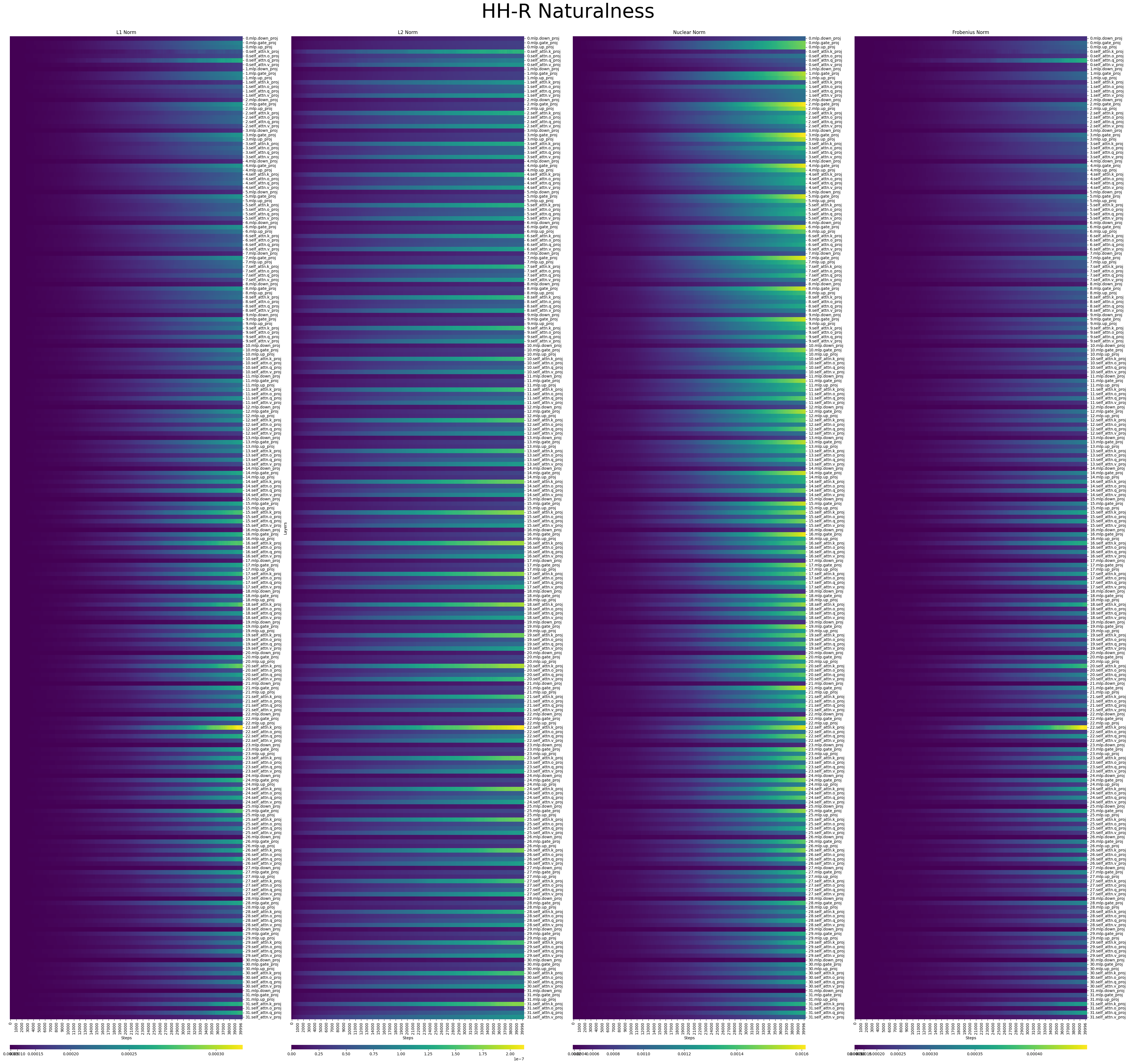}
    \caption{LoRA Weight Differences for HH-R Dataset on Naturalness dimension}
\end{figure}
\begin{figure}[!h]
    \centering
    \includegraphics[width=\textwidth]{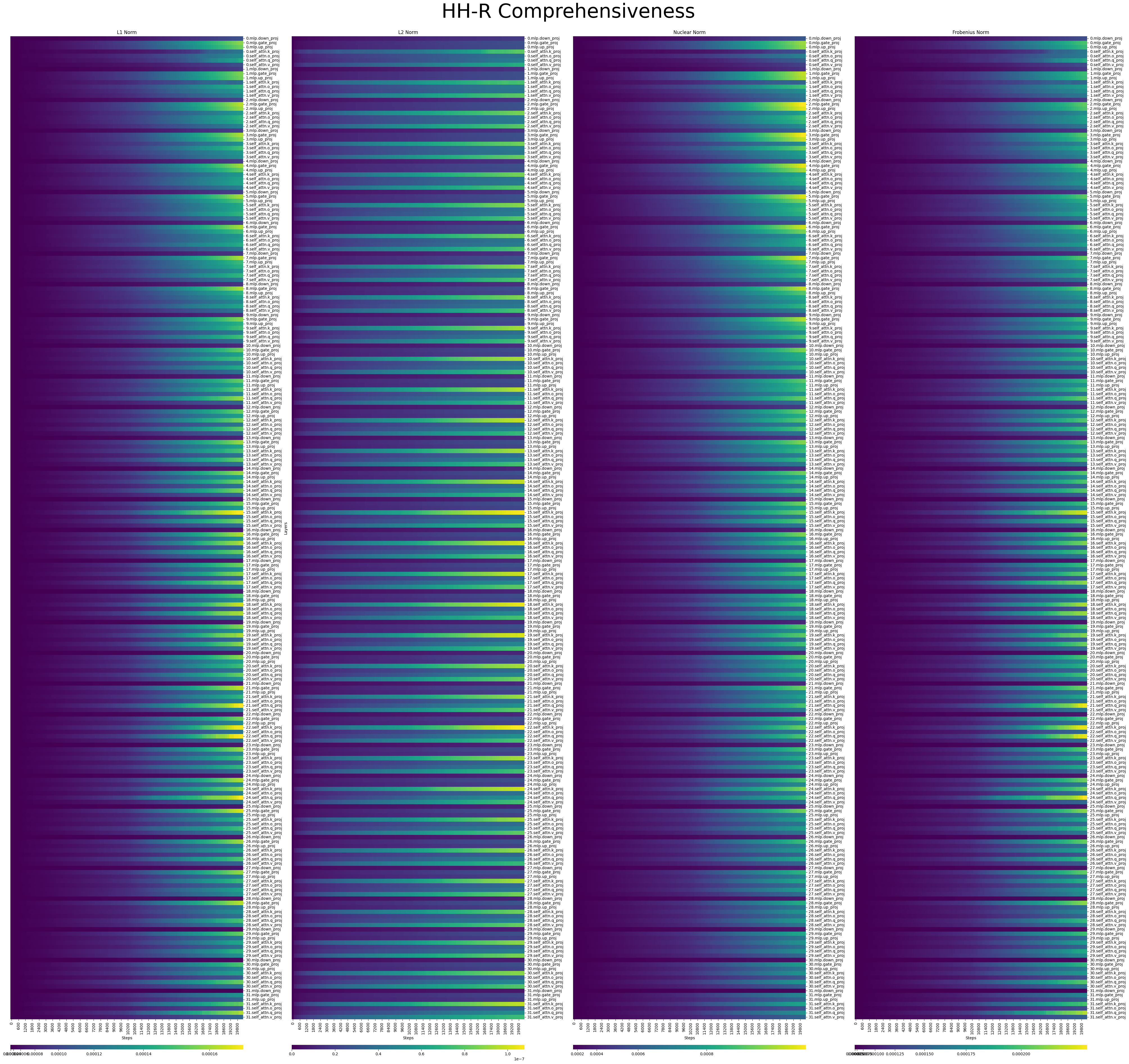}
    \caption{LoRA Weight Differences for HH-R Dataset on Comprehensiveness dimension}
\end{figure}
\begin{figure}[!h]
    \centering
    \includegraphics[width=\textwidth]{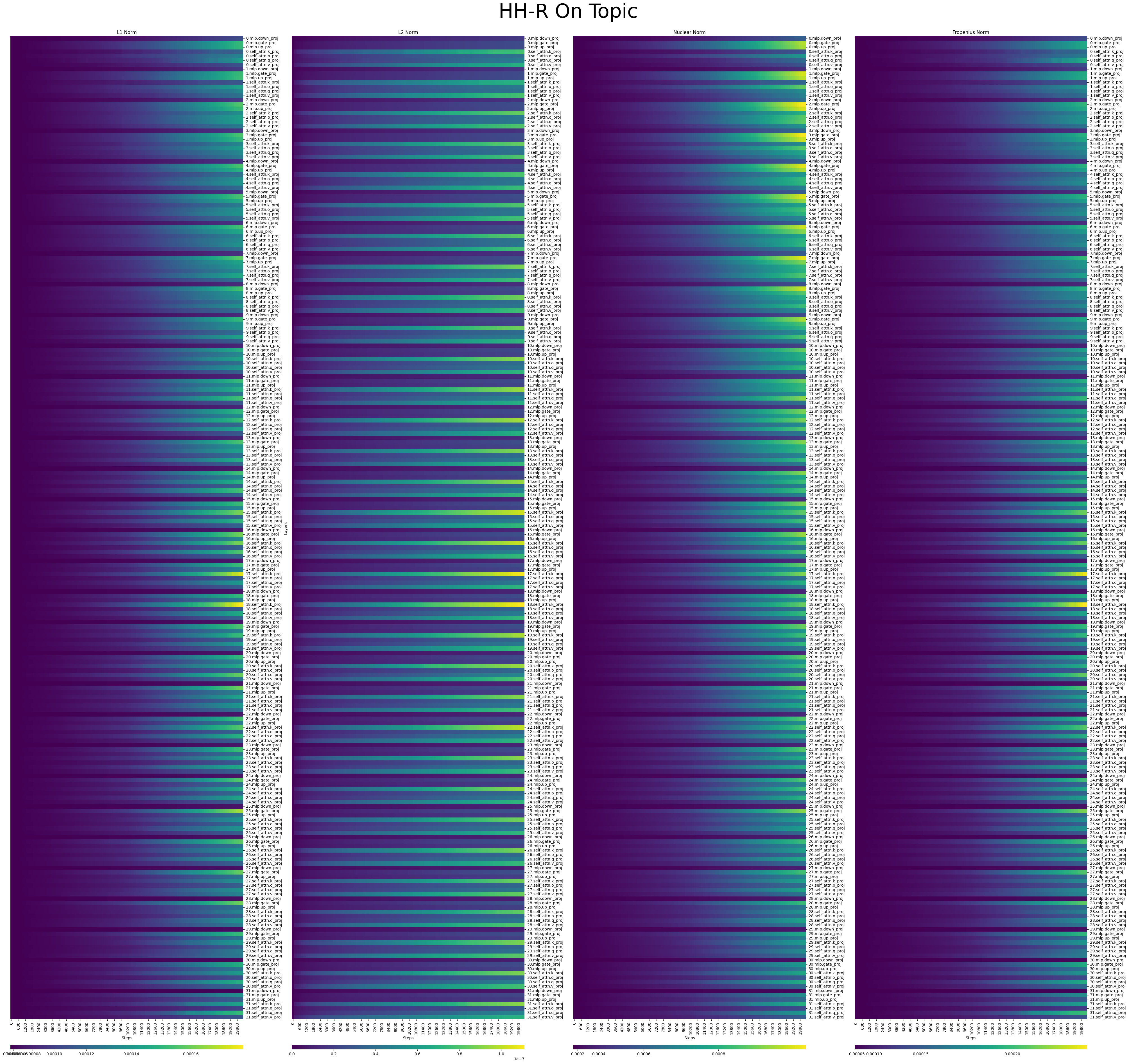}
    \caption{LoRA Weight Differences for HH-R Dataset on On-Topic dimension}
\end{figure}

\begin{figure}[!h]
    \centering
    \includegraphics[width=\textwidth]{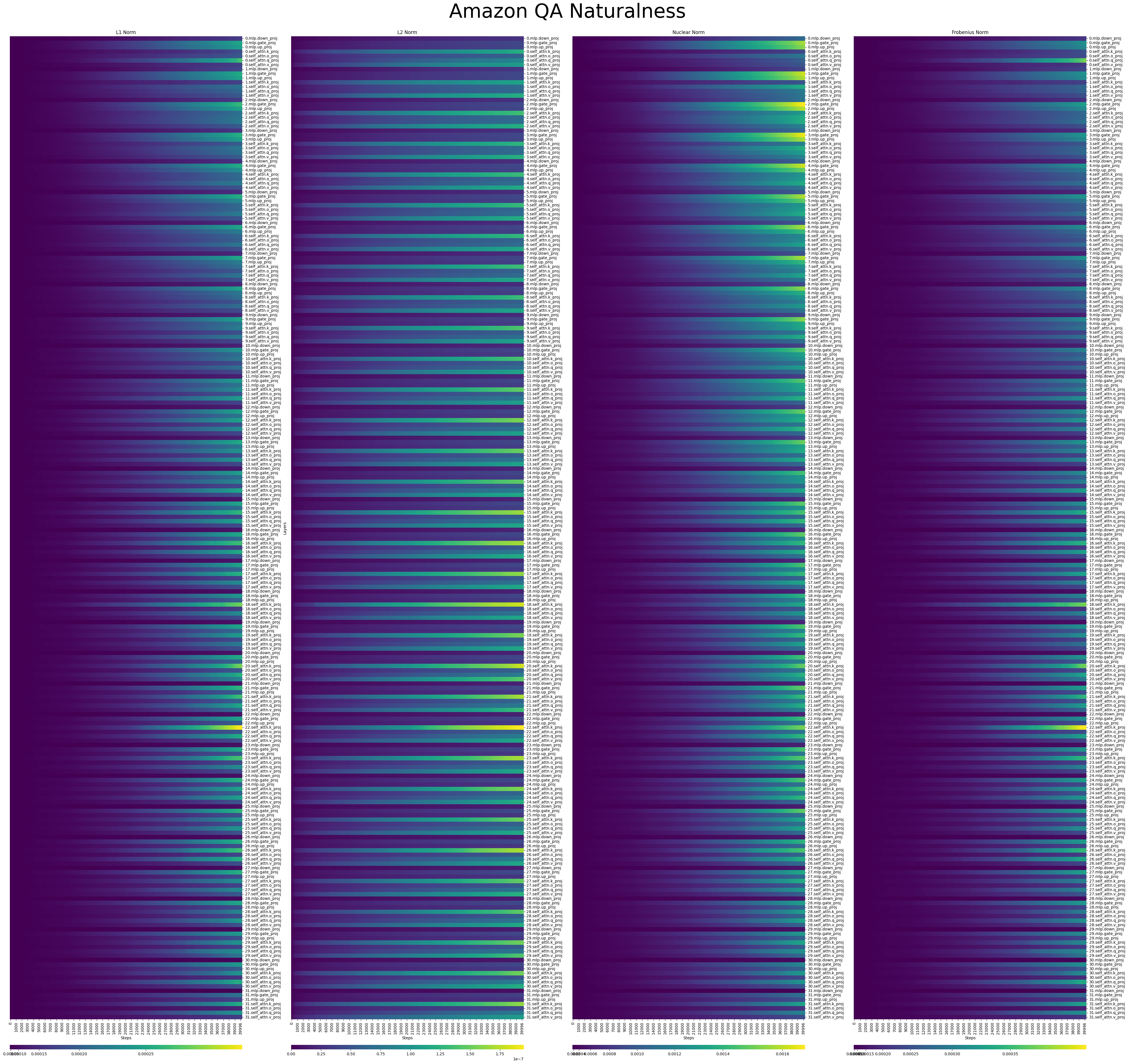}
    \caption{LoRA Weight Differences for AmazonQA Dataset on Naturalness dimension}
\end{figure}
\begin{figure}[!h]
    \centering
    \includegraphics[width=\textwidth]{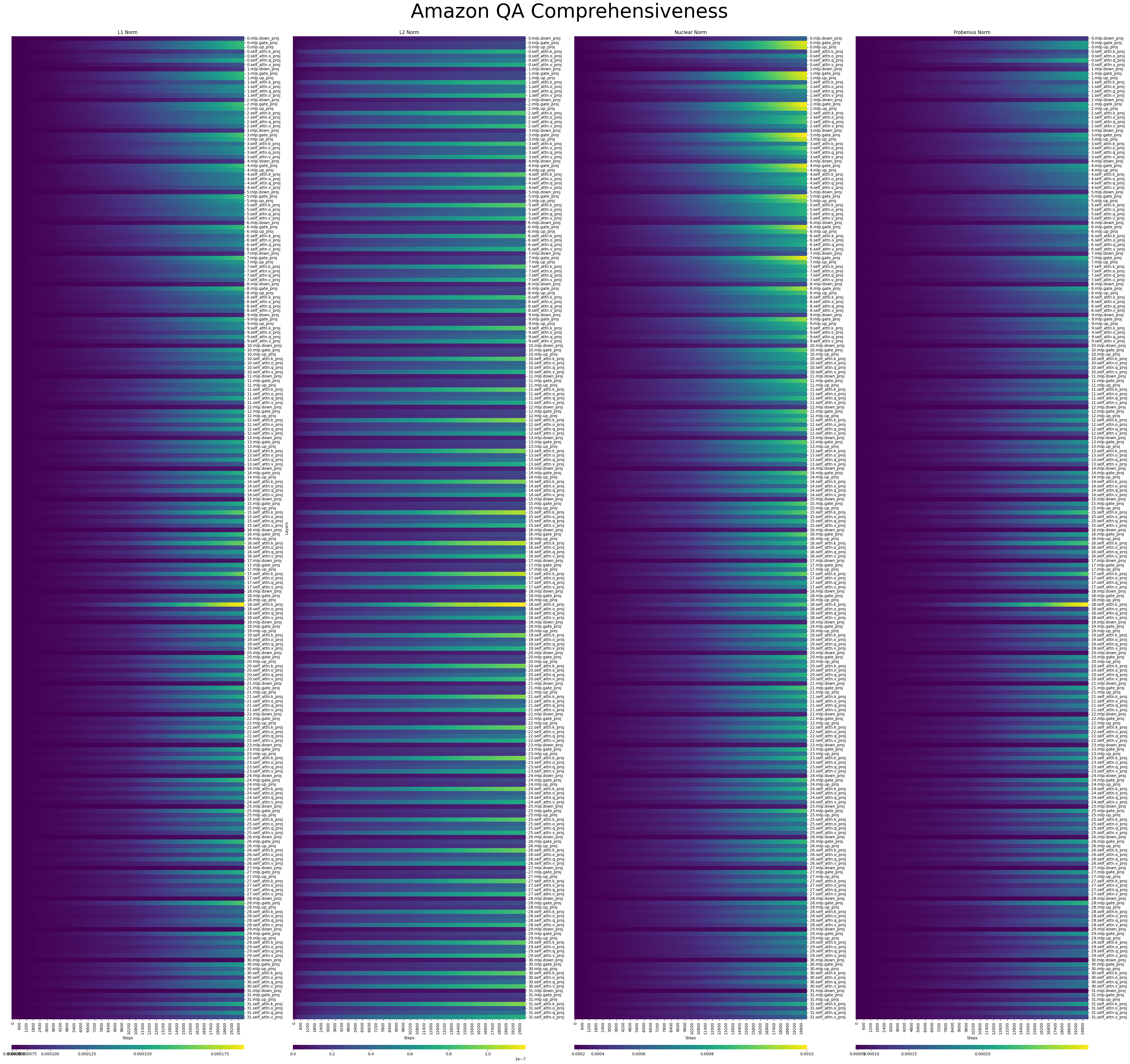}
    \caption{LoRA Weight Differences for AmazonQA Dataset on Comprehensiveness dimension}
\end{figure}
\begin{figure}[!h]
    \centering
    \includegraphics[width=\textwidth]{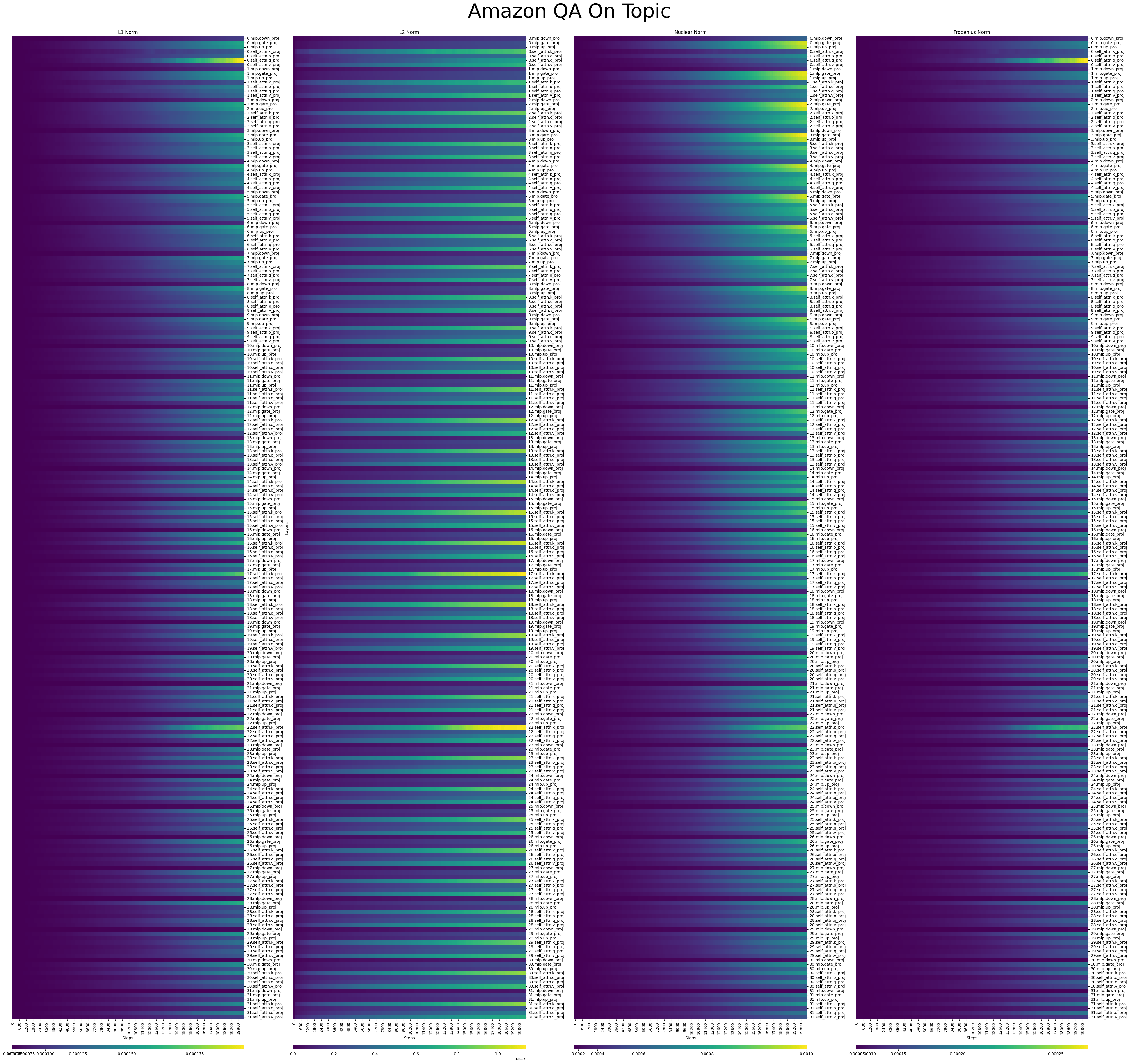}
    \caption{LoRA Weight Differences for AmazonQA Dataset on On-Topic dimension}
\end{figure}

\begin{figure}[!h]
    \centering
    \includegraphics[width=\textwidth]{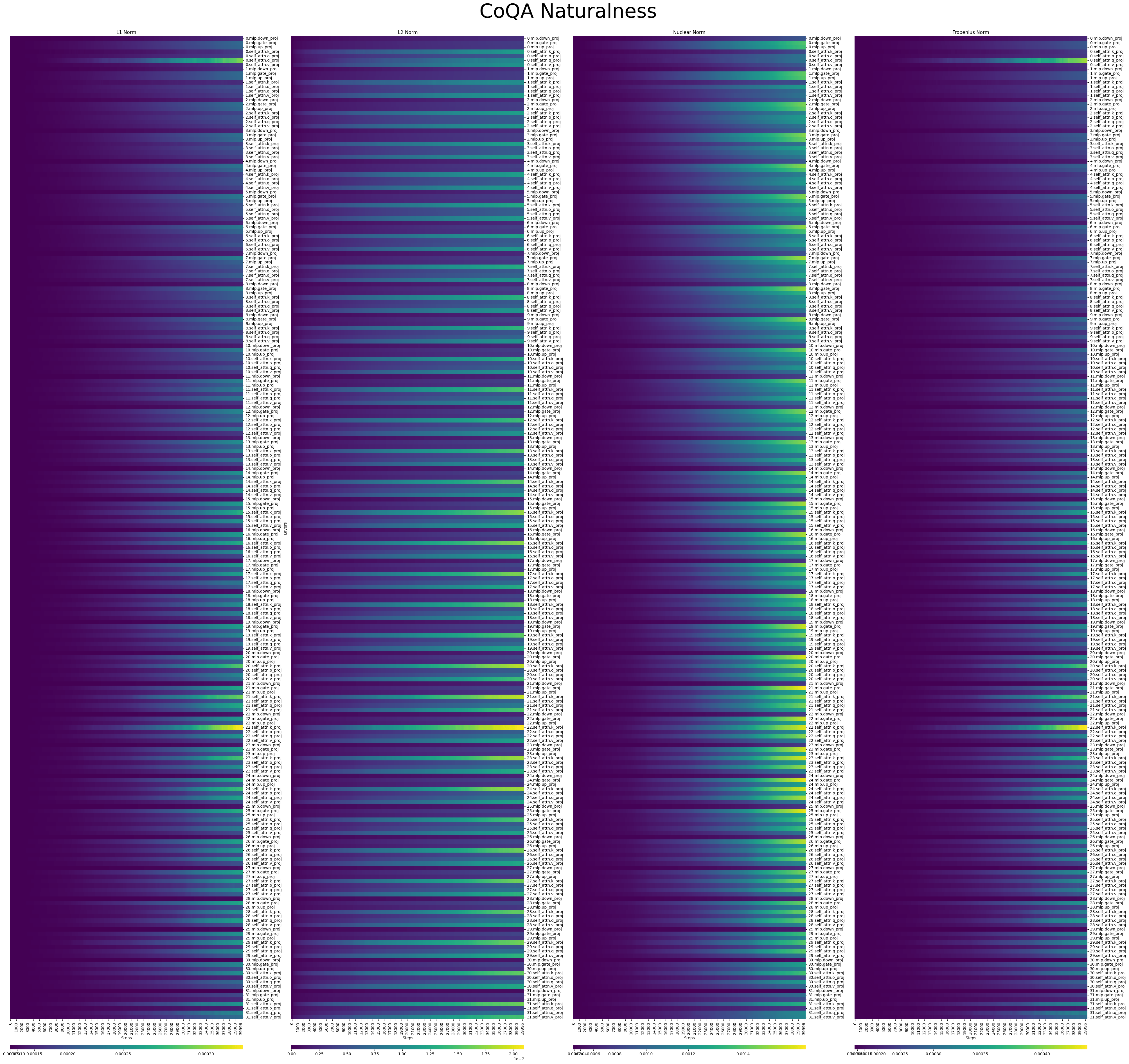}
    \caption{LoRA Weight Differences for CoQA Dataset on Naturalness dimension}
\end{figure}
\begin{figure}[!h]
    \centering
    \includegraphics[width=\textwidth]{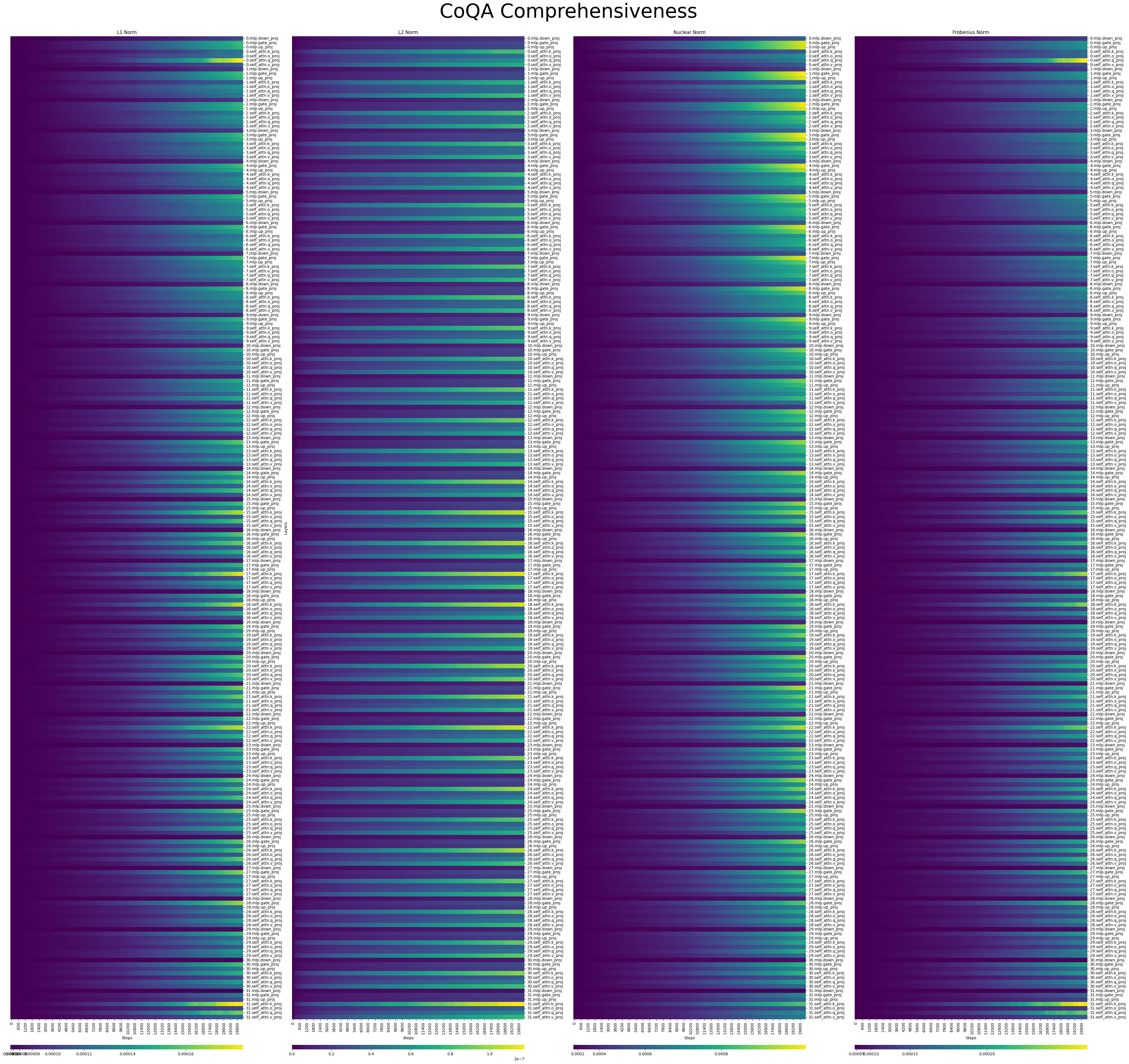}
    \caption{LoRA Weight Differences for CoQA Dataset on Comprehensiveness dimension}
\end{figure}
\begin{figure}[!h]
    \centering
    \includegraphics[width=\textwidth]{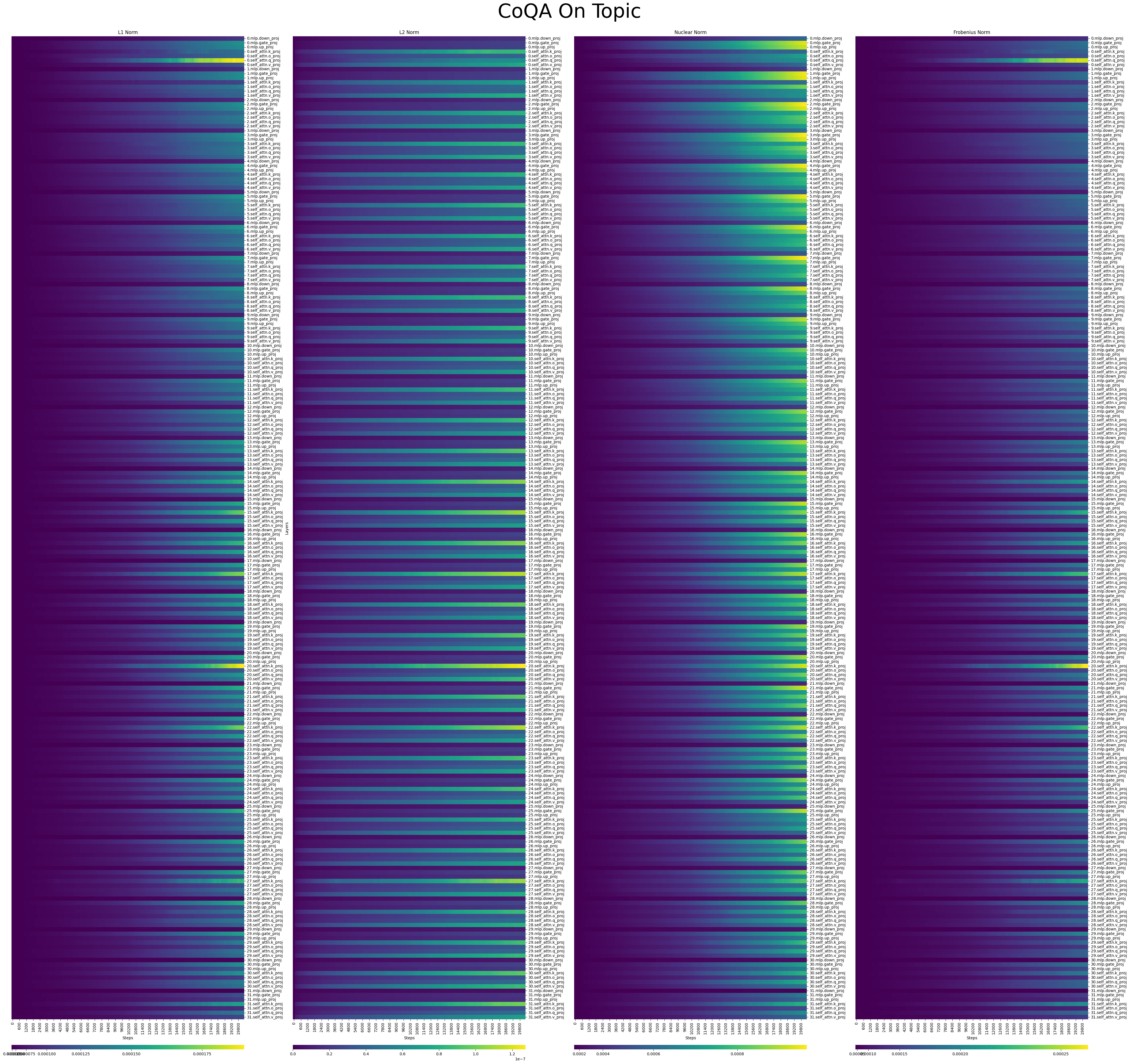}
    \caption{LoRA Weight Differences for CoQA Dataset on On-Topic dimension}
\end{figure}

\begin{figure}[!h]
    \centering
    \includegraphics[width=\textwidth]{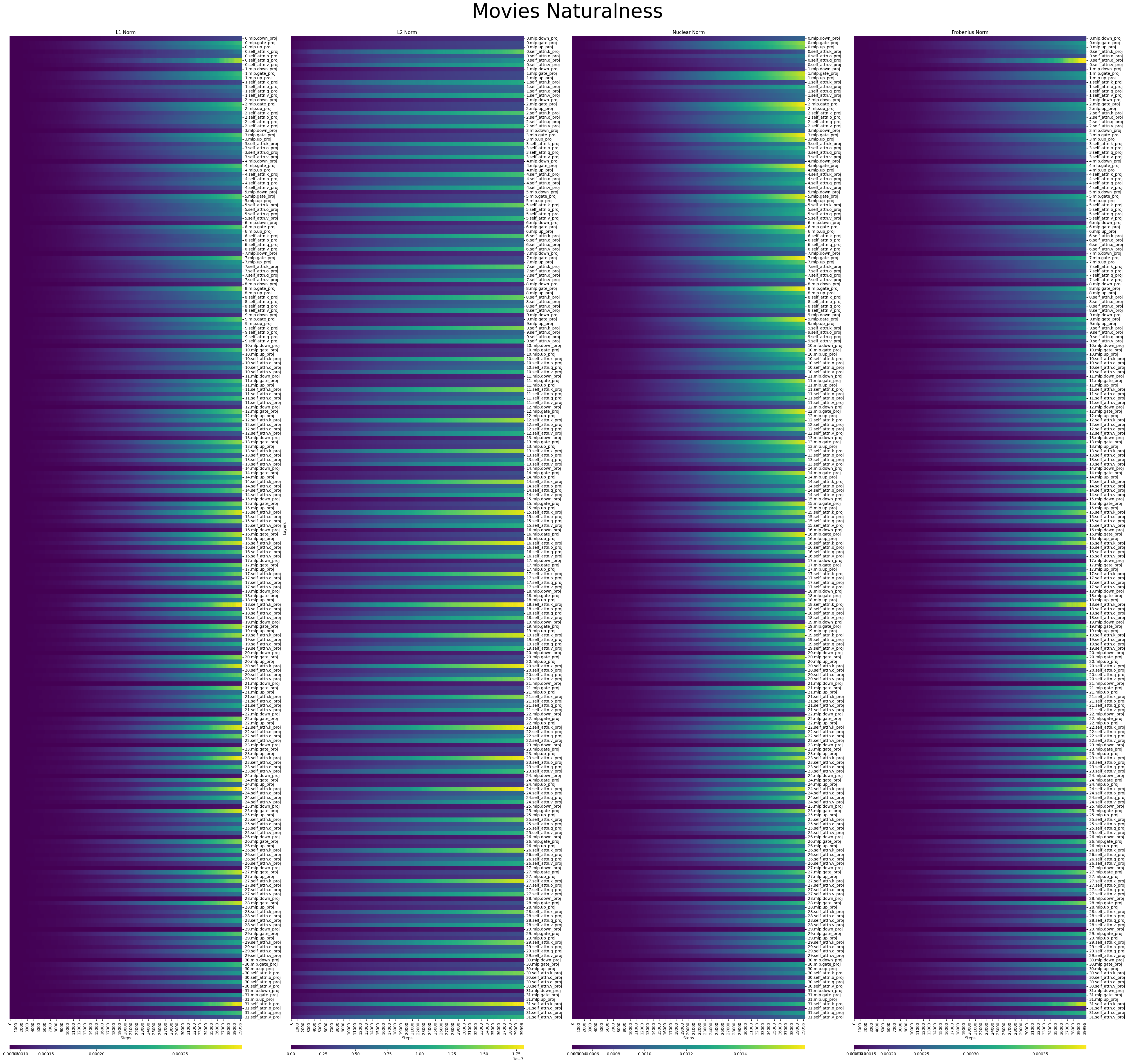}
    \caption{LoRA Weight Differences for Movies Dataset on Naturalness dimension}
\end{figure}
\begin{figure}[!h]
    \centering
    \includegraphics[width=\textwidth]{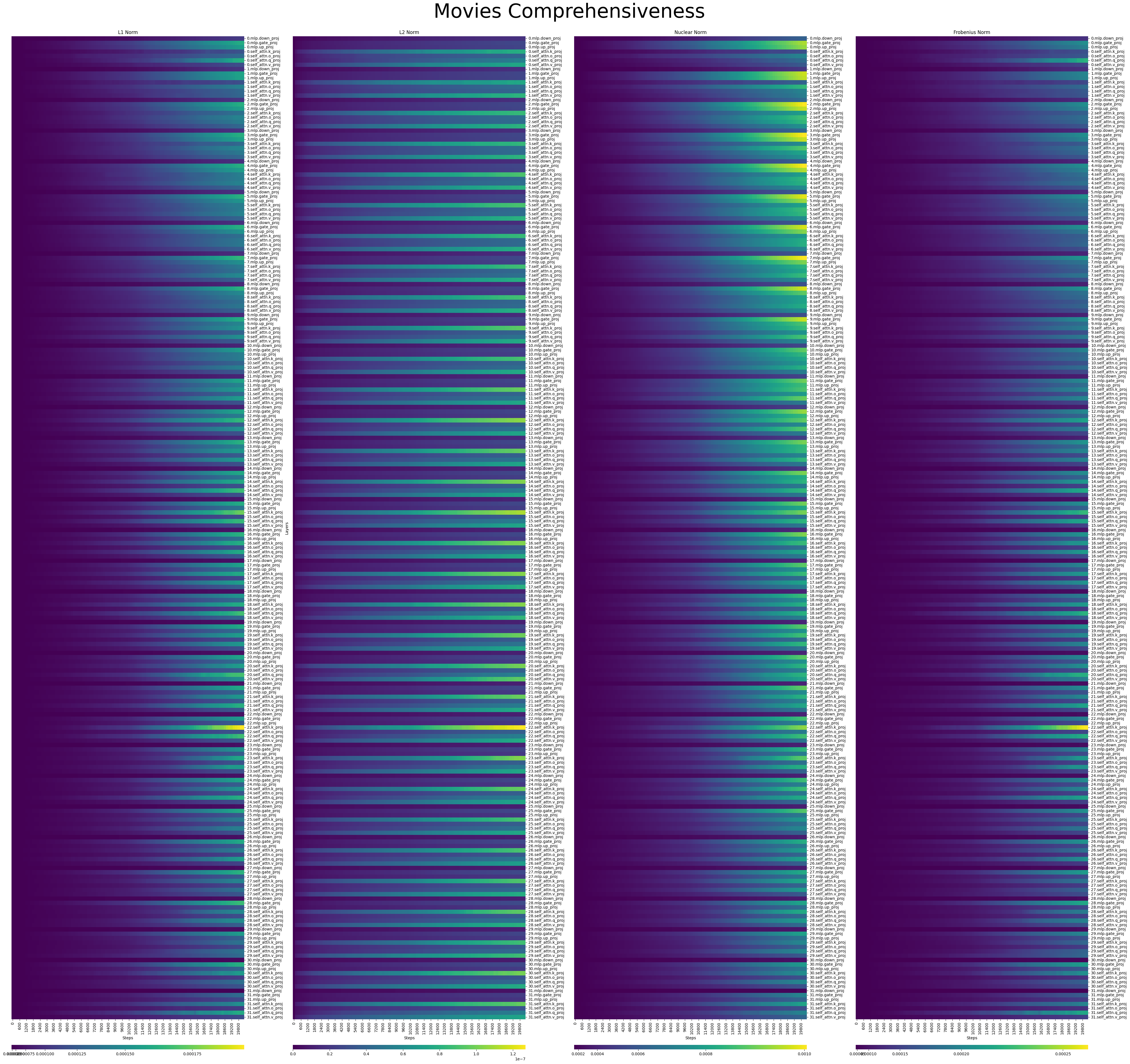}
    \caption{LoRA Weight Differences for Movies Dataset on Comprehensiveness dimension}
\end{figure}
\begin{figure}[!h]
    \centering
    \includegraphics[width=\textwidth]{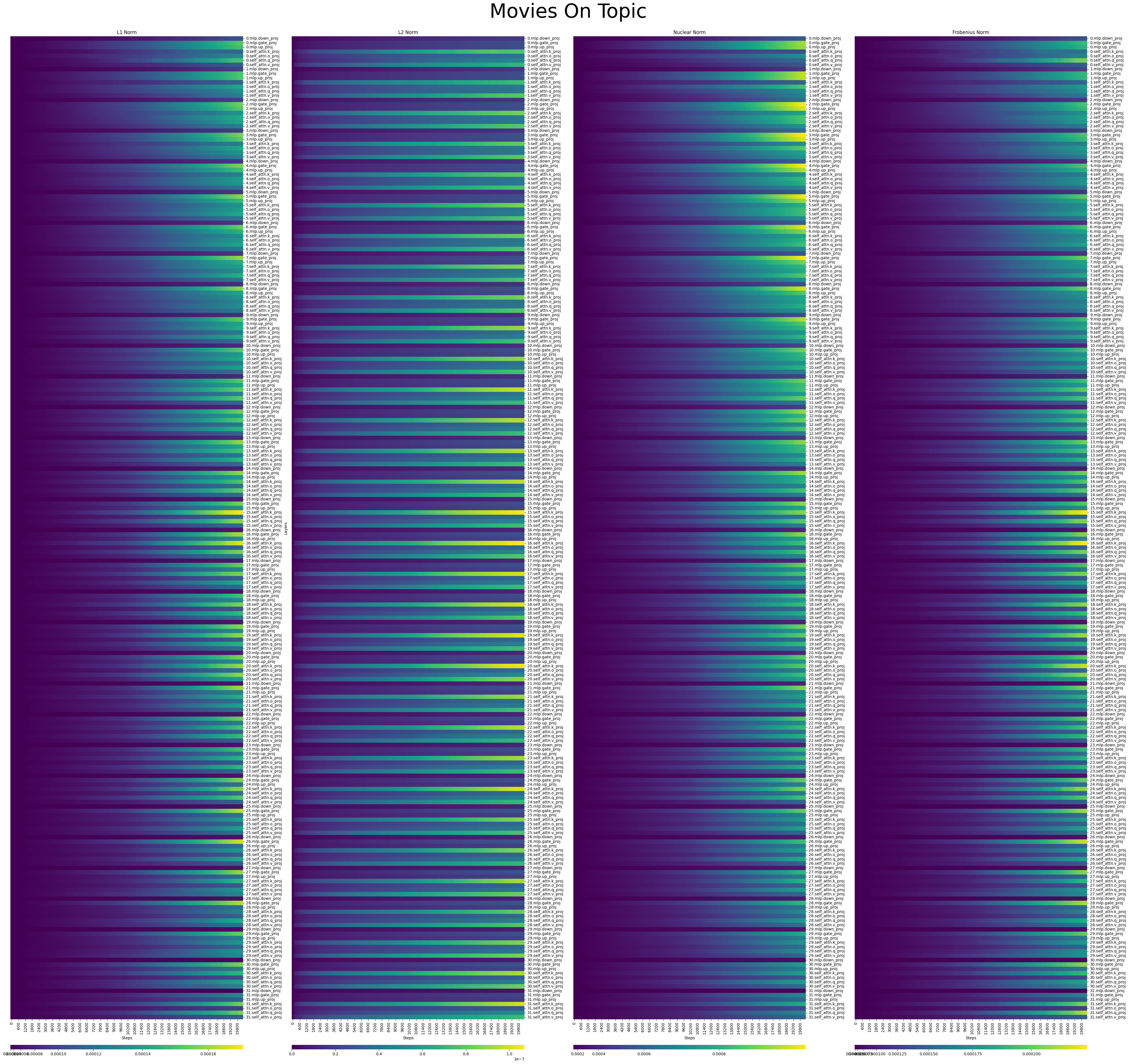}
    \caption{LoRA Weight Differences for Movies Dataset on On-Topic dimension}
\end{figure}

%% file: sections/92-appendix-rankings.tex
\subsection{Output Token Ranking}
\label{section:appendix-ouput-token-ranking}
\FloatBarrier

\begin{figure}[!h]
    \centering
    \includegraphics[width=\textwidth]{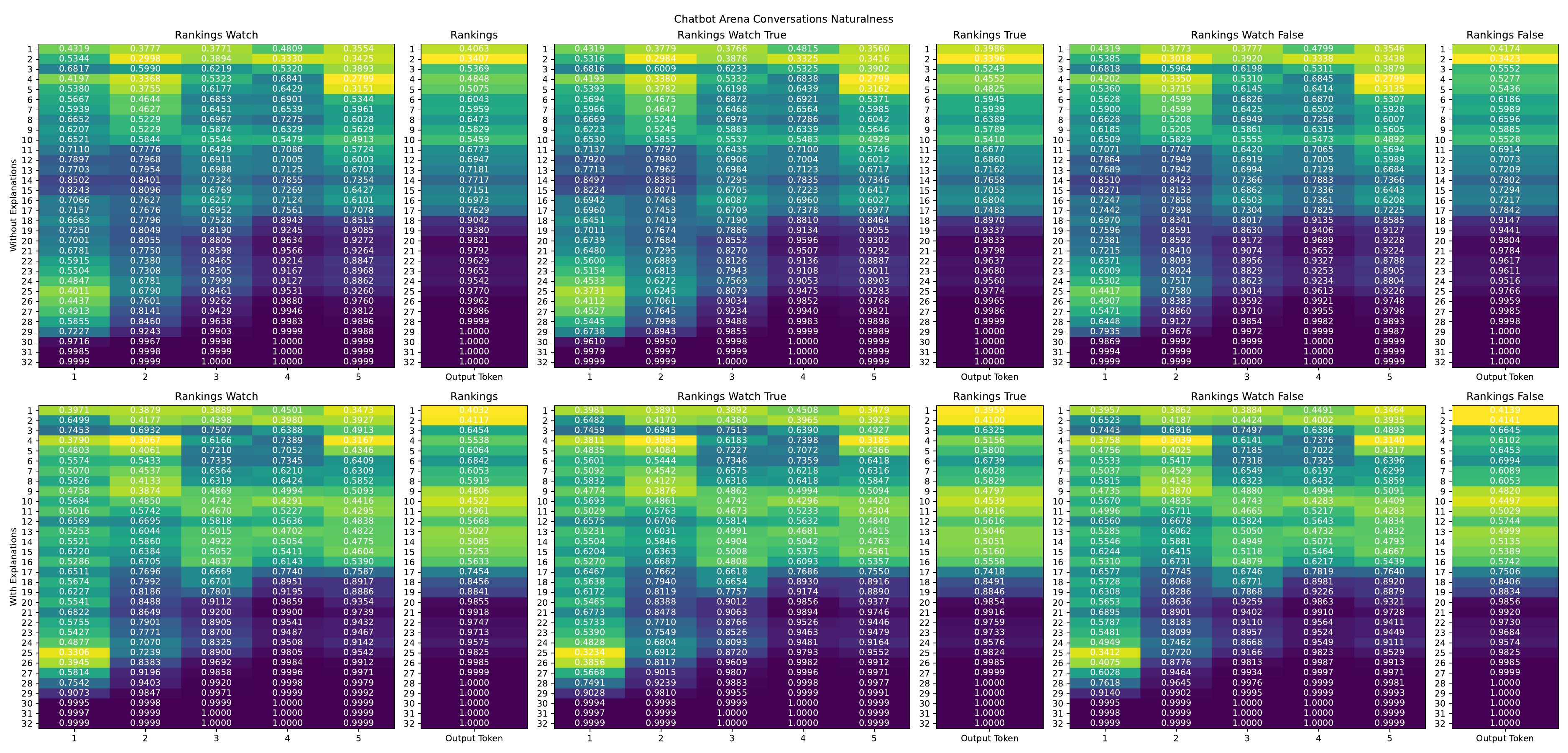}
    \caption{Output Token Ranking for ChatbotAC Dataset on Naturalness dimension using Ecco \citep{alammarEccoOpenSource2021}}
\end{figure}
\begin{figure}[!h]
    \centering
    \includegraphics[width=\textwidth]{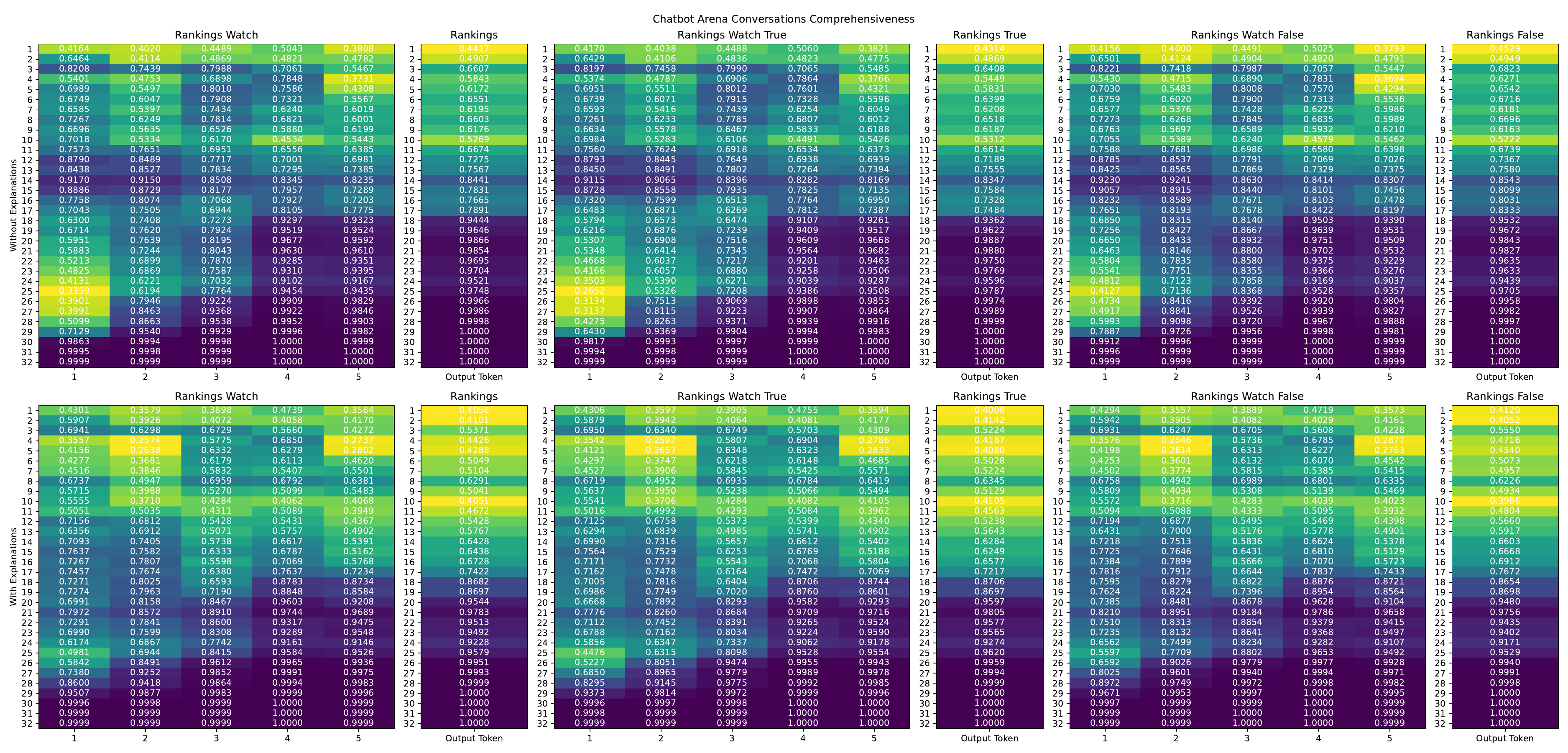}
    \caption{Output Token Ranking for ChatbotAC Dataset on Comprehensiveness dimension using Ecco \citep{alammarEccoOpenSource2021}}
\end{figure}
\begin{figure}[!h]
    \centering
    \includegraphics[width=\textwidth]{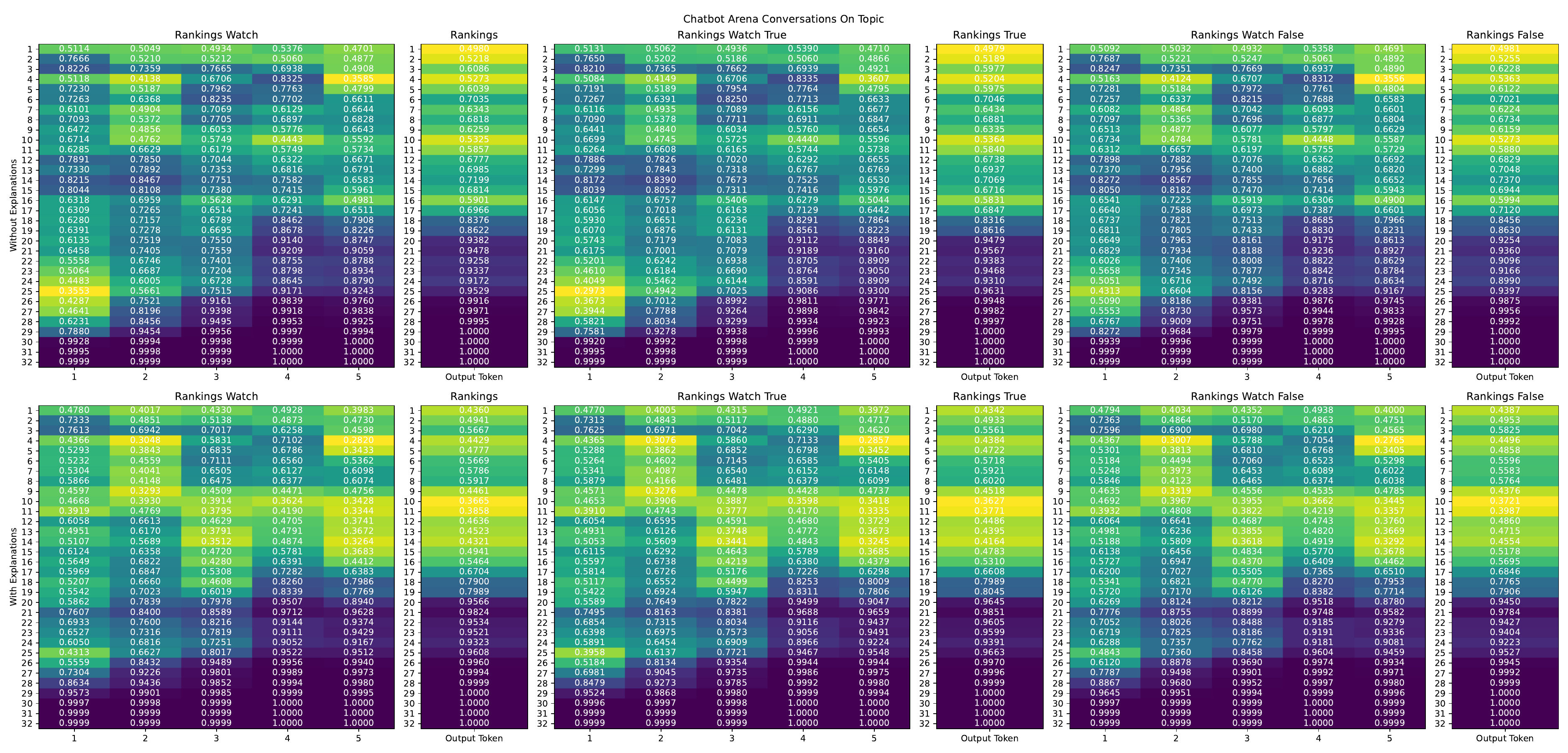}
    \caption{Output Token Ranking for ChatbotAC Dataset on On-Topic dimension using Ecco \citep{alammarEccoOpenSource2021}}
\end{figure}

\begin{figure}[!h]
    \centering
    \includegraphics[width=\textwidth]{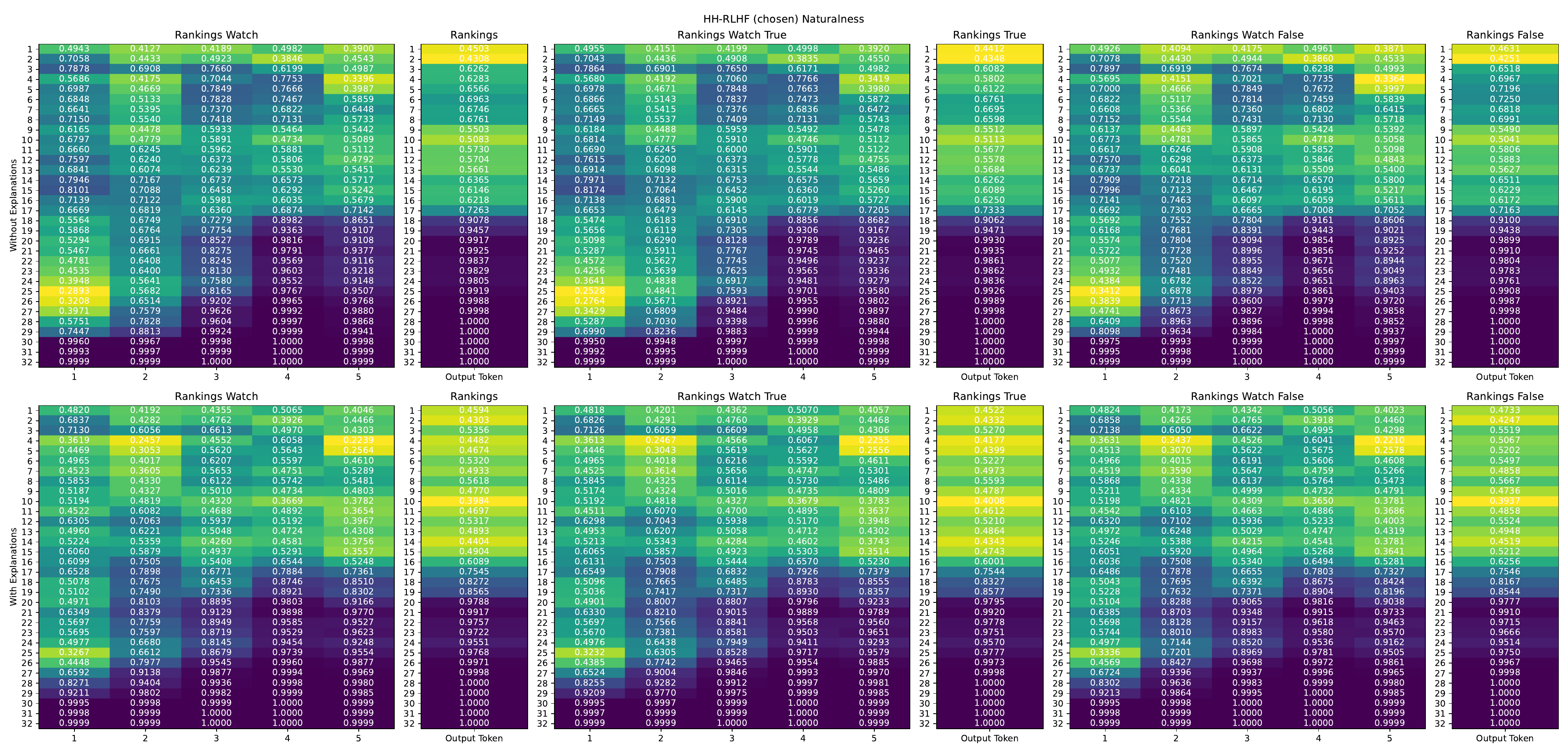}
    \caption{Output Token Ranking for HH-C Dataset on Naturalness dimension using Ecco \citep{alammarEccoOpenSource2021}}
\end{figure}
\begin{figure}[!h]
    \centering
    \includegraphics[width=\textwidth]{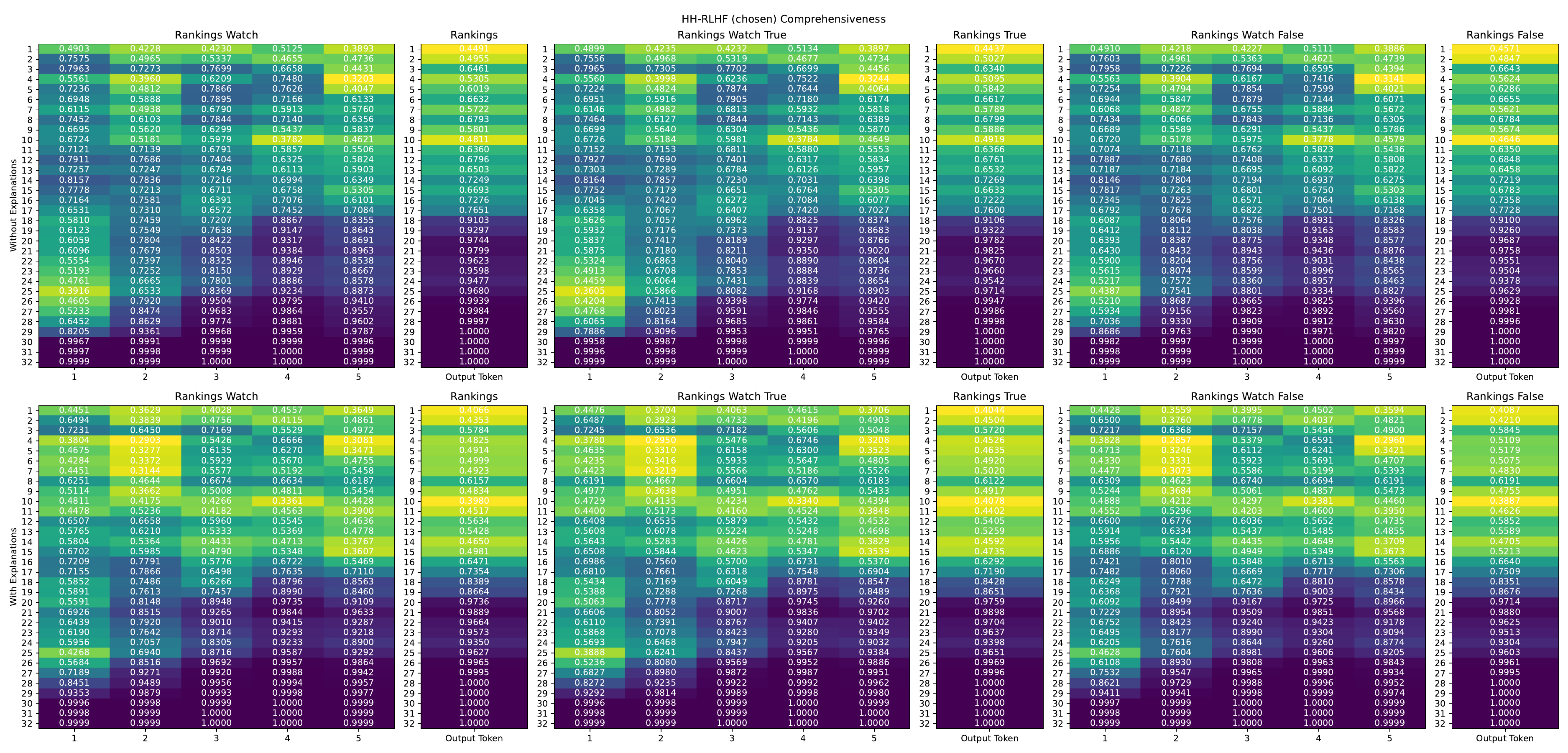}
    \caption{Output Token Ranking for HH-C Dataset on Comprehensiveness dimension using Ecco \citep{alammarEccoOpenSource2021}}
\end{figure}
\begin{figure}[!h]
    \centering
    \includegraphics[width=\textwidth]{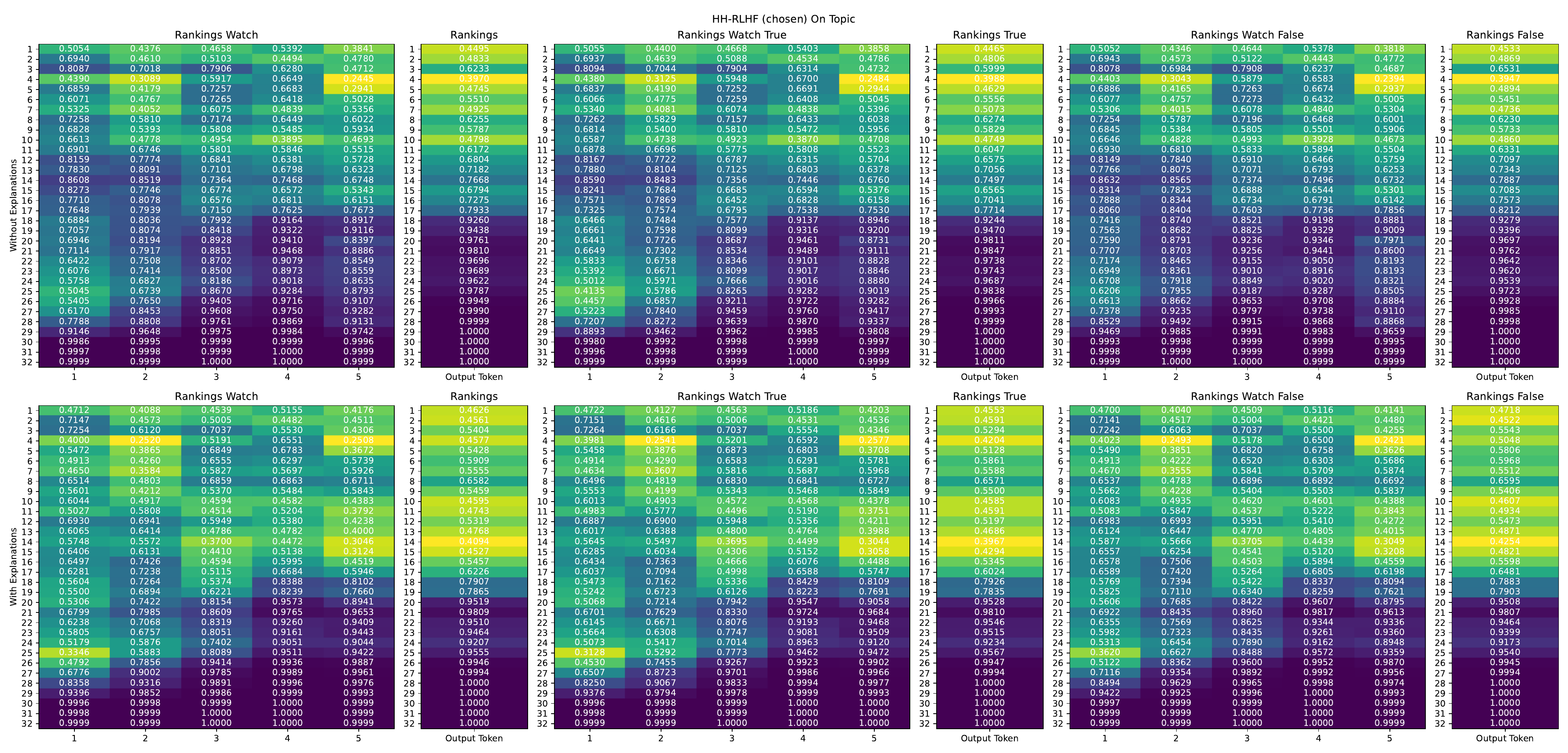}
    \caption{Output Token Ranking for HH-C Dataset on On-Topic dimension using Ecco \citep{alammarEccoOpenSource2021}}
\end{figure}

\begin{figure}[!h]
    \centering
    \includegraphics[width=\textwidth]{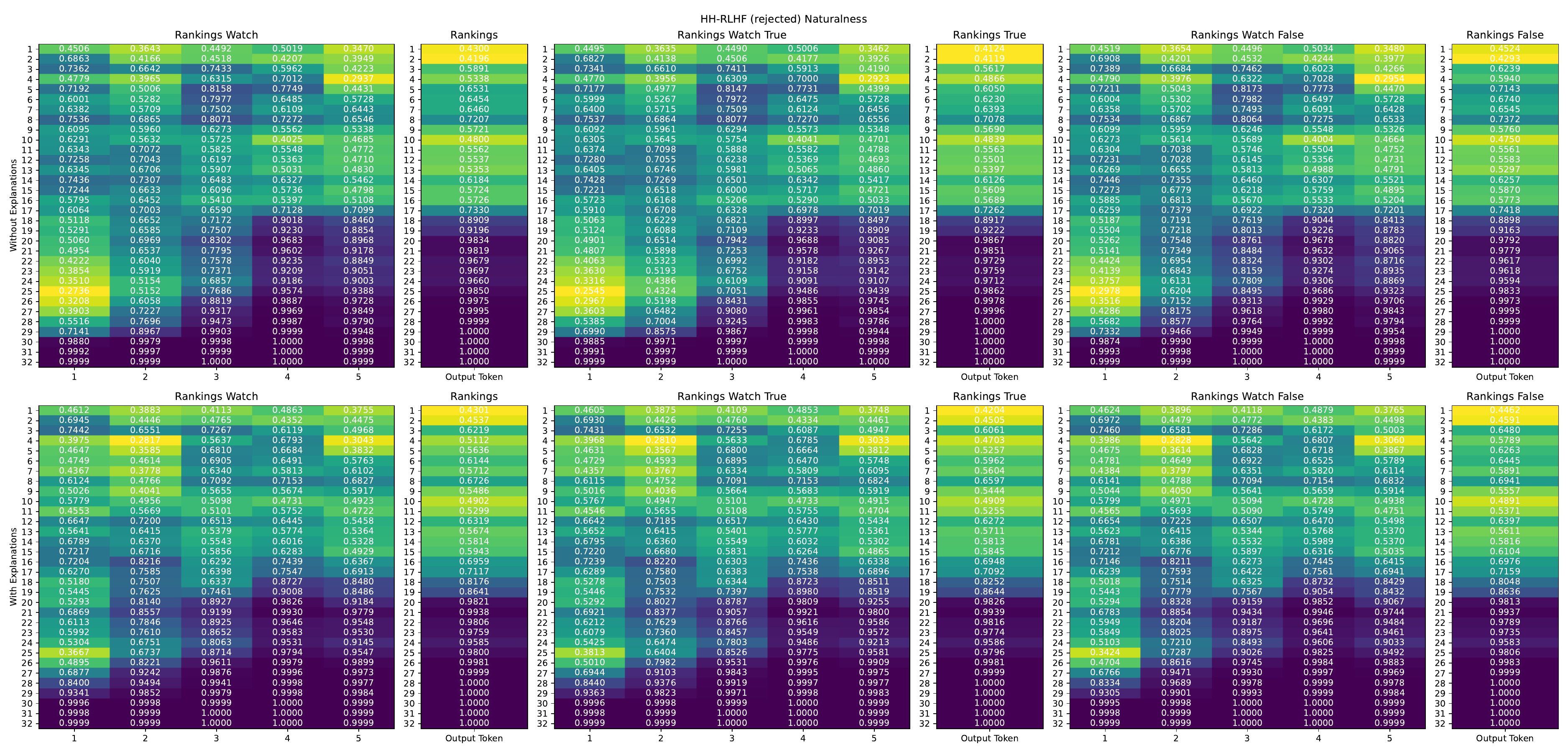}
    \caption{Output Token Ranking for HH-R Dataset on Naturalness dimension using Ecco \citep{alammarEccoOpenSource2021}}
\end{figure}
\begin{figure}[!h]
    \centering
    \includegraphics[width=\textwidth]{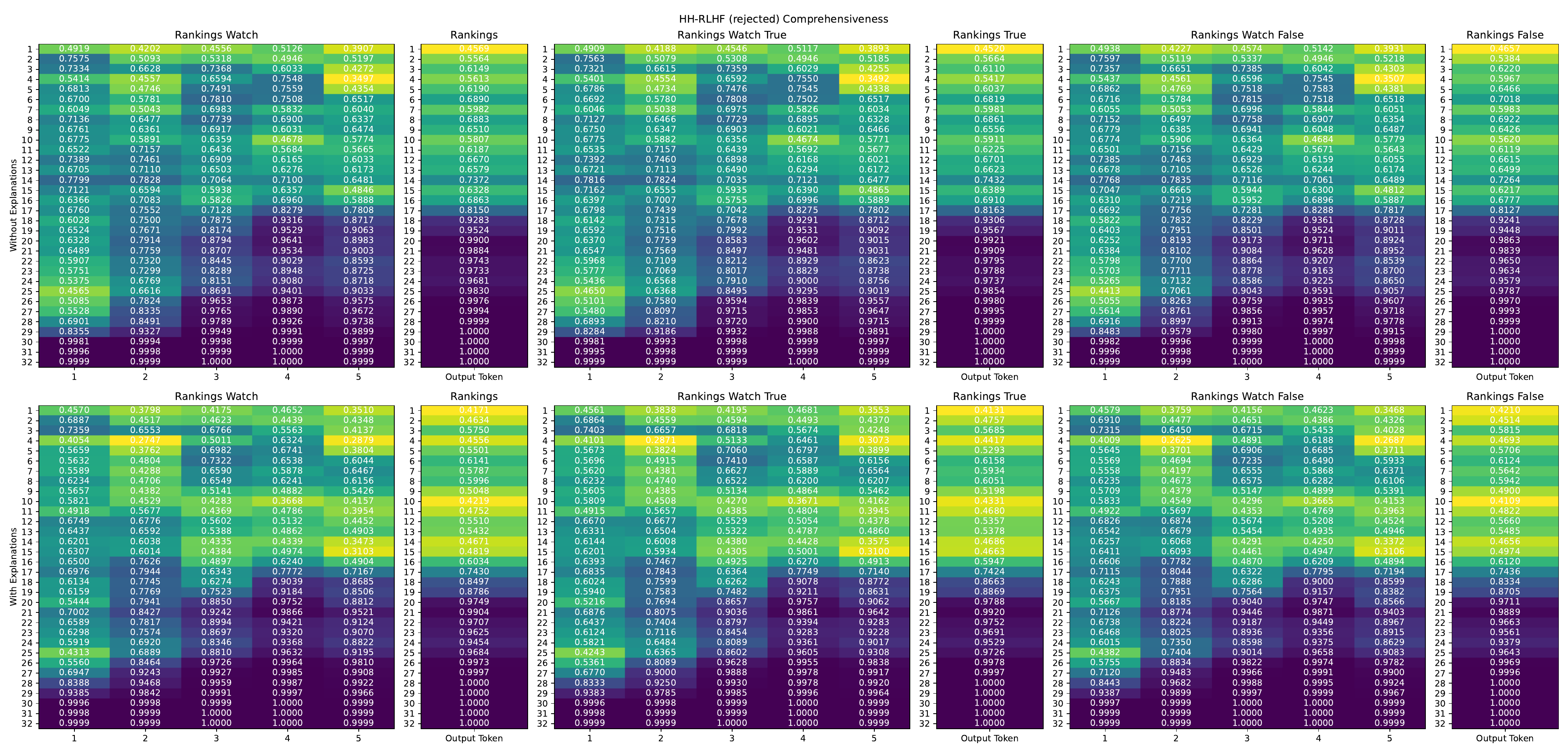}
    \caption{Output Token Ranking for HH-R Dataset on Comprehensiveness dimension using Ecco \citep{alammarEccoOpenSource2021}}
\end{figure}
\begin{figure}[!h]
    \centering
    \includegraphics[width=\textwidth]{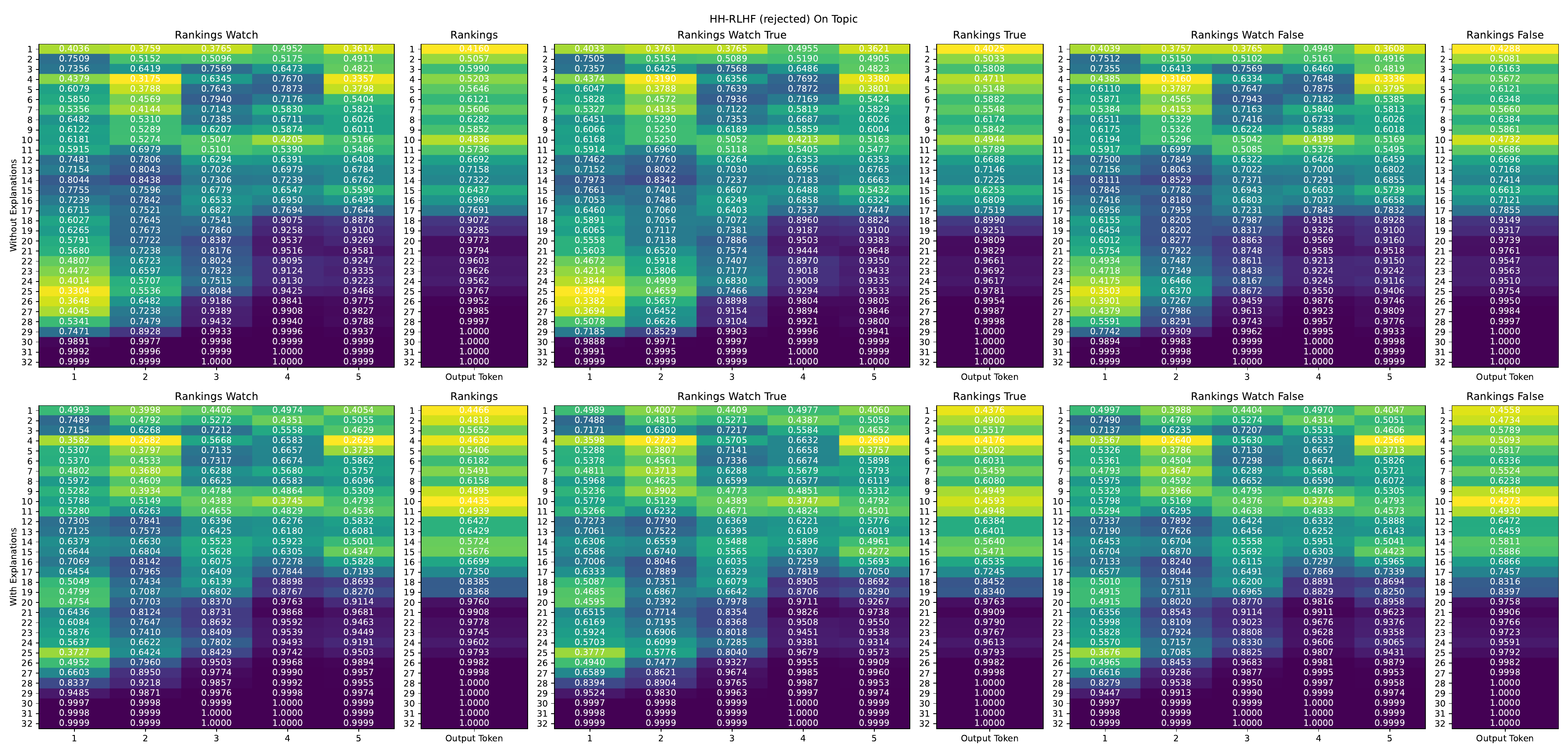}
    \caption{Output Token Ranking for HH-R Dataset on On-Topic dimension using Ecco \citep{alammarEccoOpenSource2021}}
\end{figure}

\begin{figure}[!h]
    \centering
    \includegraphics[width=\textwidth]{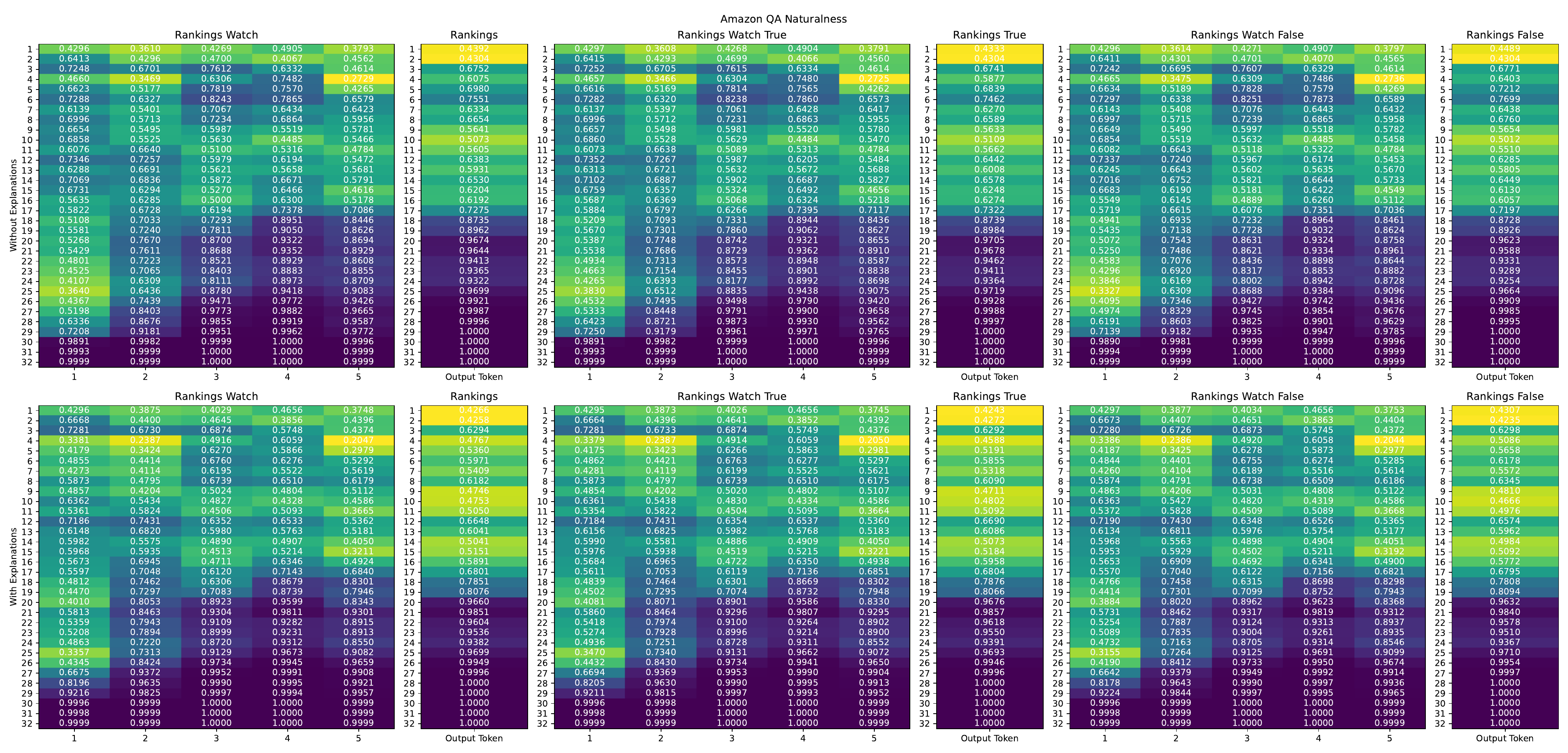}
    \caption{Output Token Ranking for AmazonQA Dataset on Naturalness dimension using Ecco \citep{alammarEccoOpenSource2021}}
\end{figure}
\begin{figure}[!h]
    \centering
    \includegraphics[width=\textwidth]{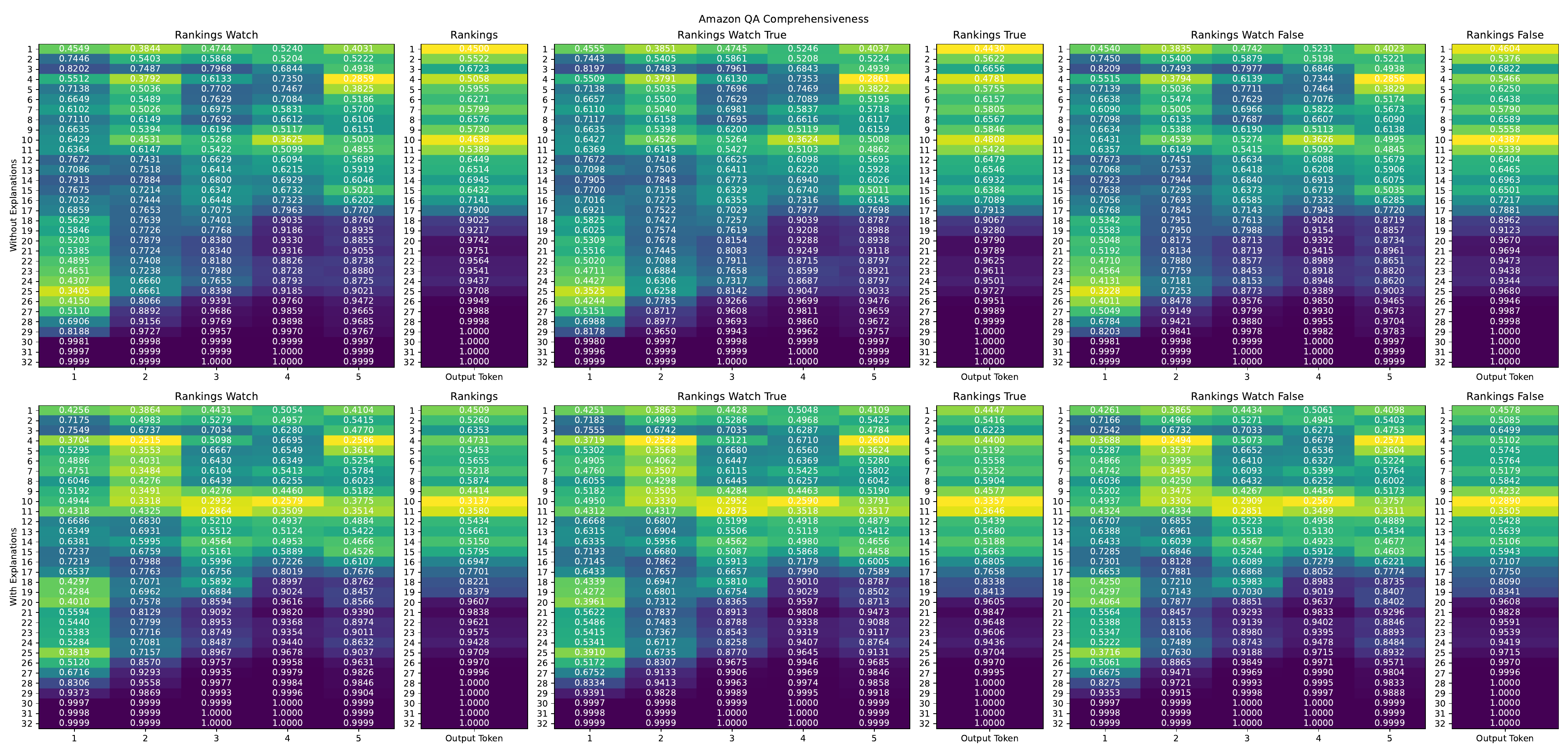}
    \caption{Output Token Ranking for AmazonQA Dataset on Comprehensiveness dimension using Ecco \citep{alammarEccoOpenSource2021}}
\end{figure}
\begin{figure}[!h]
    \centering
    \includegraphics[width=\textwidth]{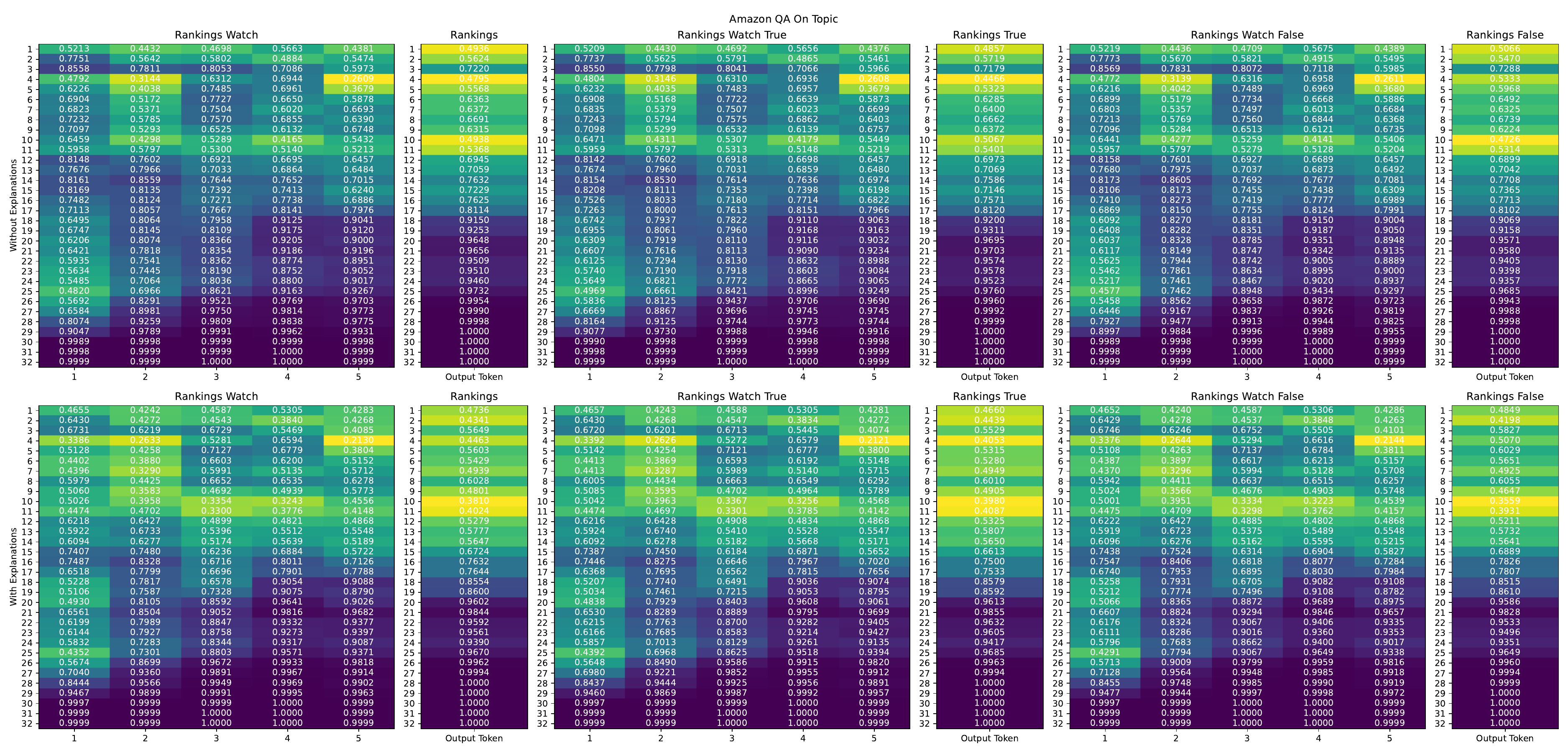}
    \caption{Output Token Ranking for AmazonQA Dataset on On-Topic dimension using Ecco \citep{alammarEccoOpenSource2021}}
\end{figure}

\begin{figure}[!h]
    \centering
    \includegraphics[width=\textwidth]{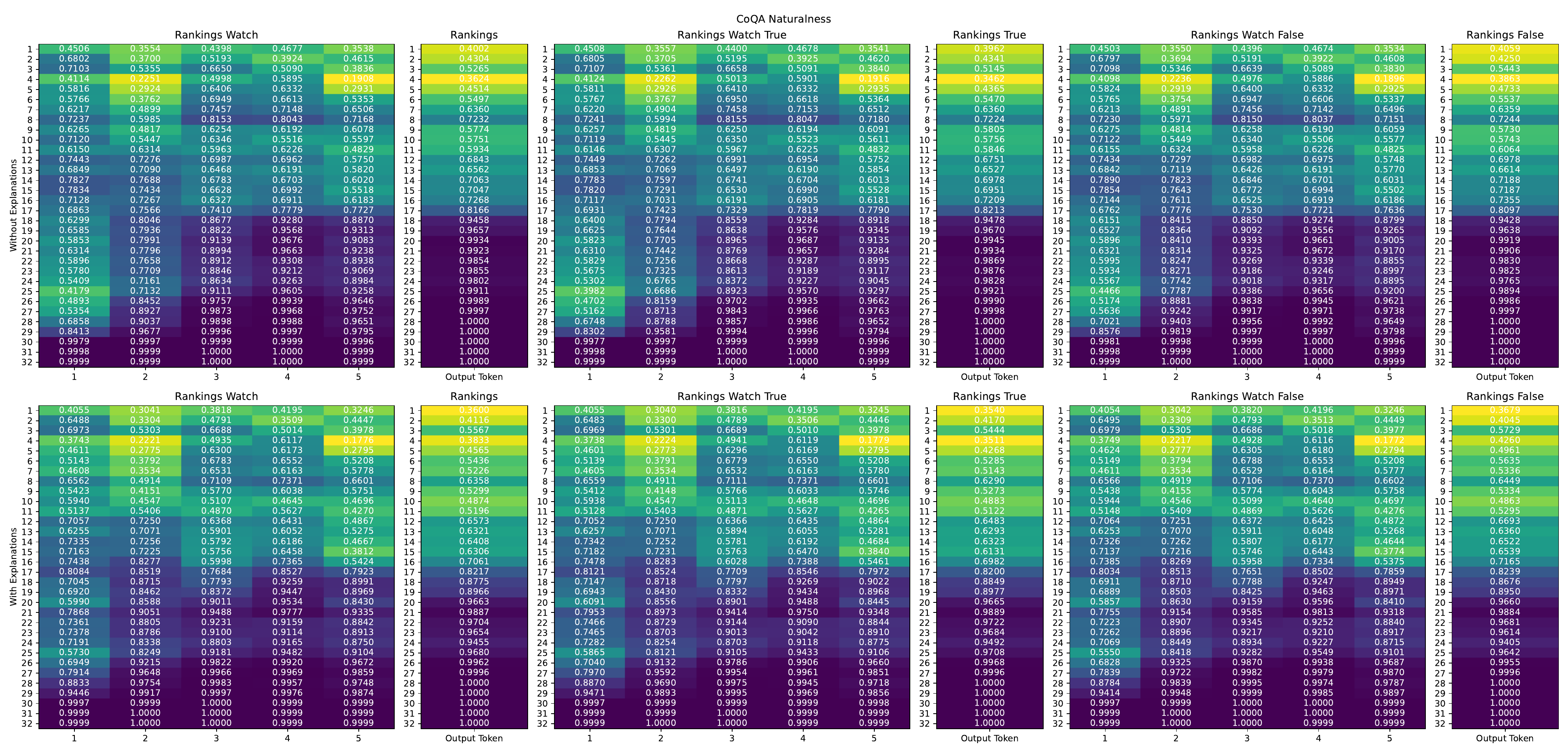}
    \caption{Output Token Ranking for CoQA Dataset on Naturalness dimension using Ecco \citep{alammarEccoOpenSource2021}}
\end{figure}
\begin{figure}[!h]
    \centering
    \includegraphics[width=\textwidth]{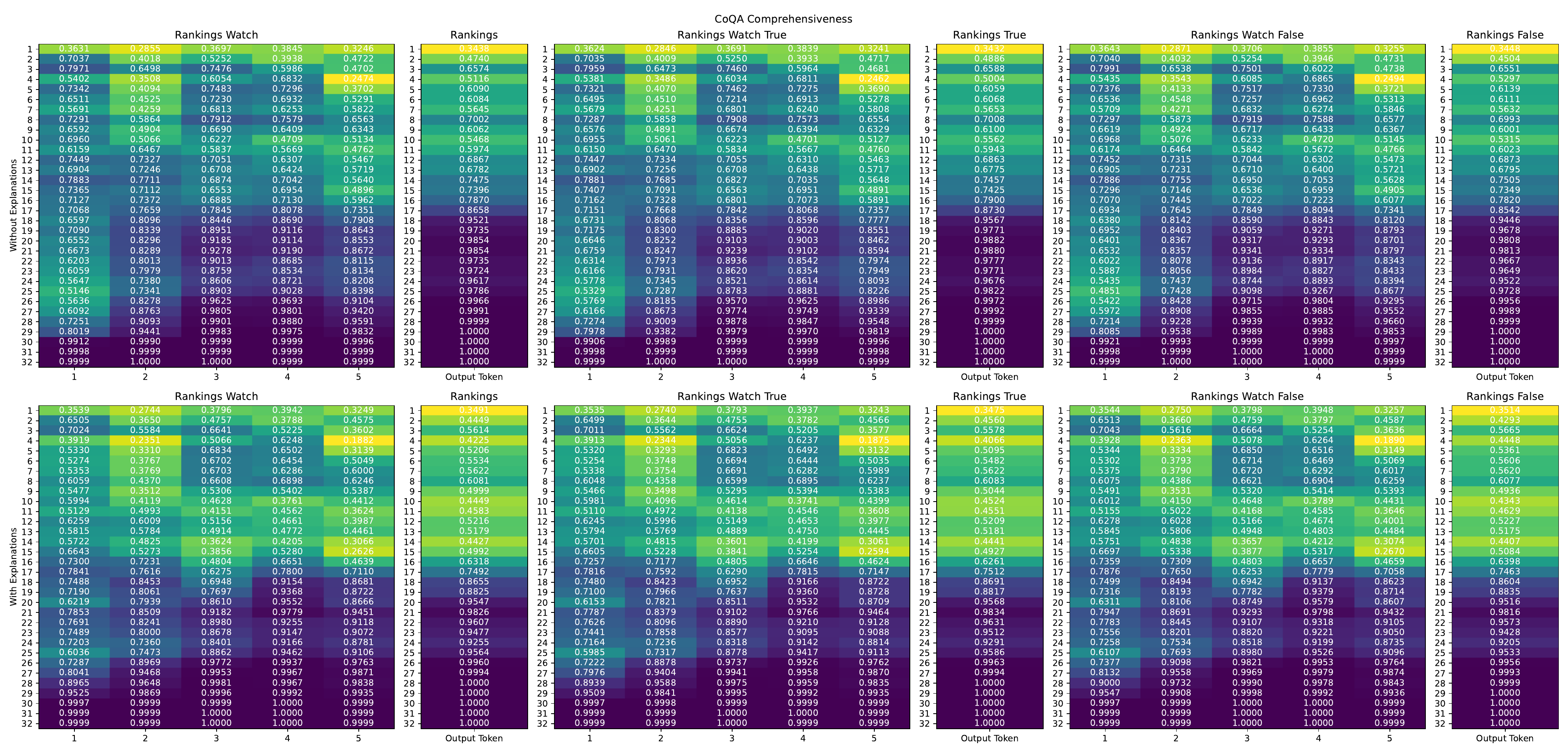}
    \caption{Output Token Ranking for CoQA Dataset on Comprehensiveness dimension using Ecco \citep{alammarEccoOpenSource2021}}
\end{figure}
\begin{figure}[!h]
    \centering
    \includegraphics[width=\textwidth]{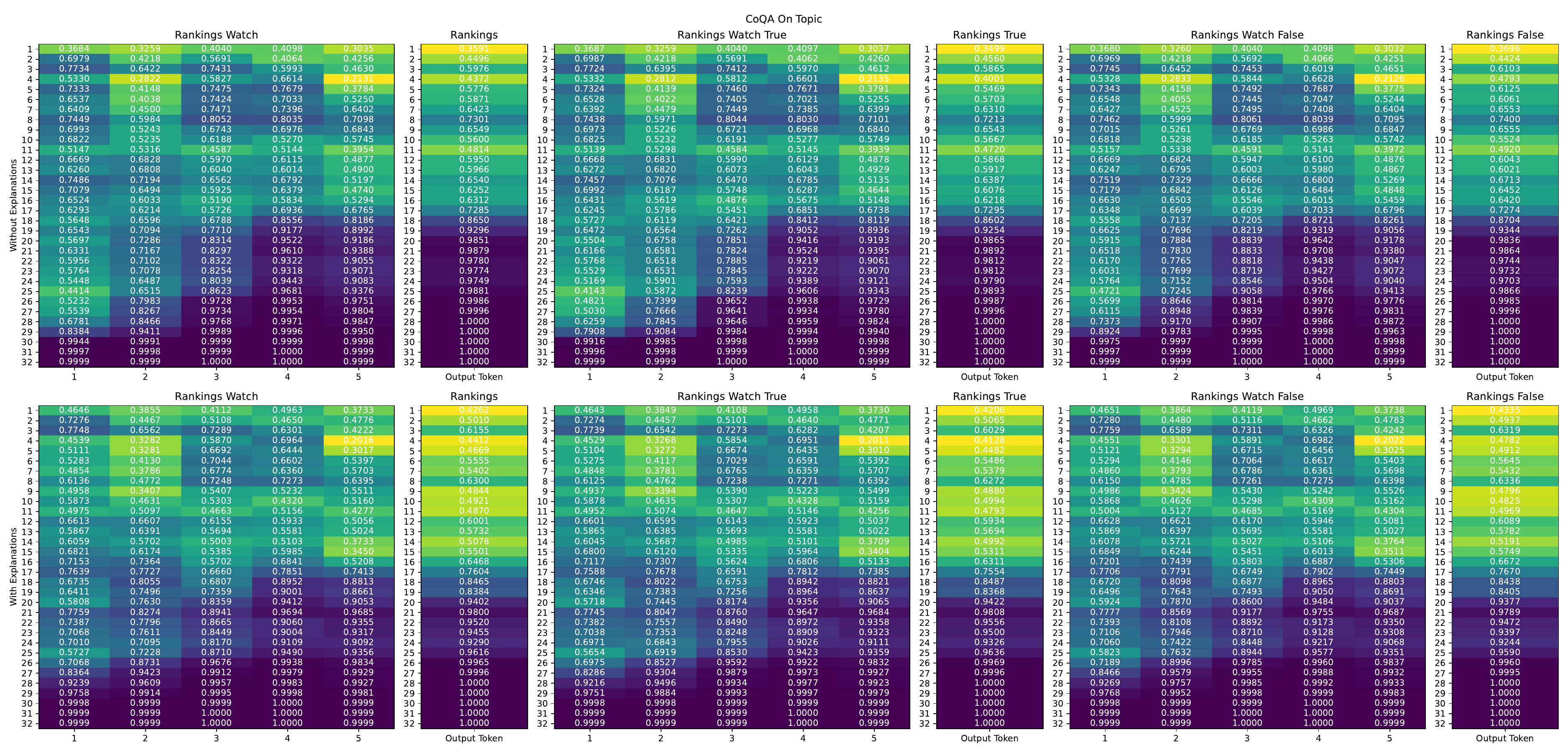}
    \caption{Output Token Ranking for CoQA Dataset on On-Topic dimension using Ecco \citep{alammarEccoOpenSource2021}}
\end{figure}

\begin{figure}[!h]
    \centering
    \includegraphics[width=\textwidth]{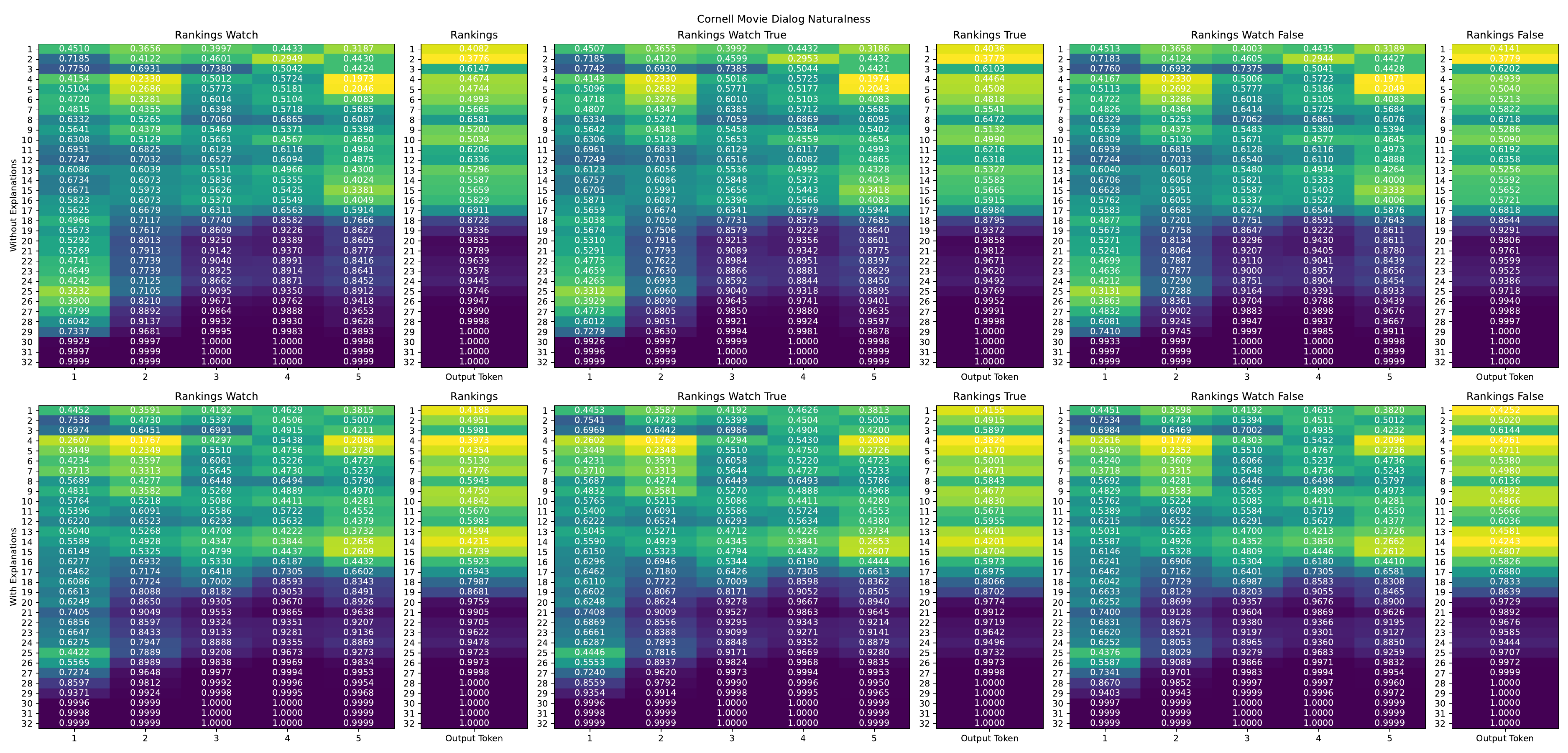}
    \caption{Output Token Ranking for Movies Dataset on Naturalness dimension using Ecco \citep{alammarEccoOpenSource2021}}
\end{figure}
\begin{figure}[!h]
    \centering
    \includegraphics[width=\textwidth]{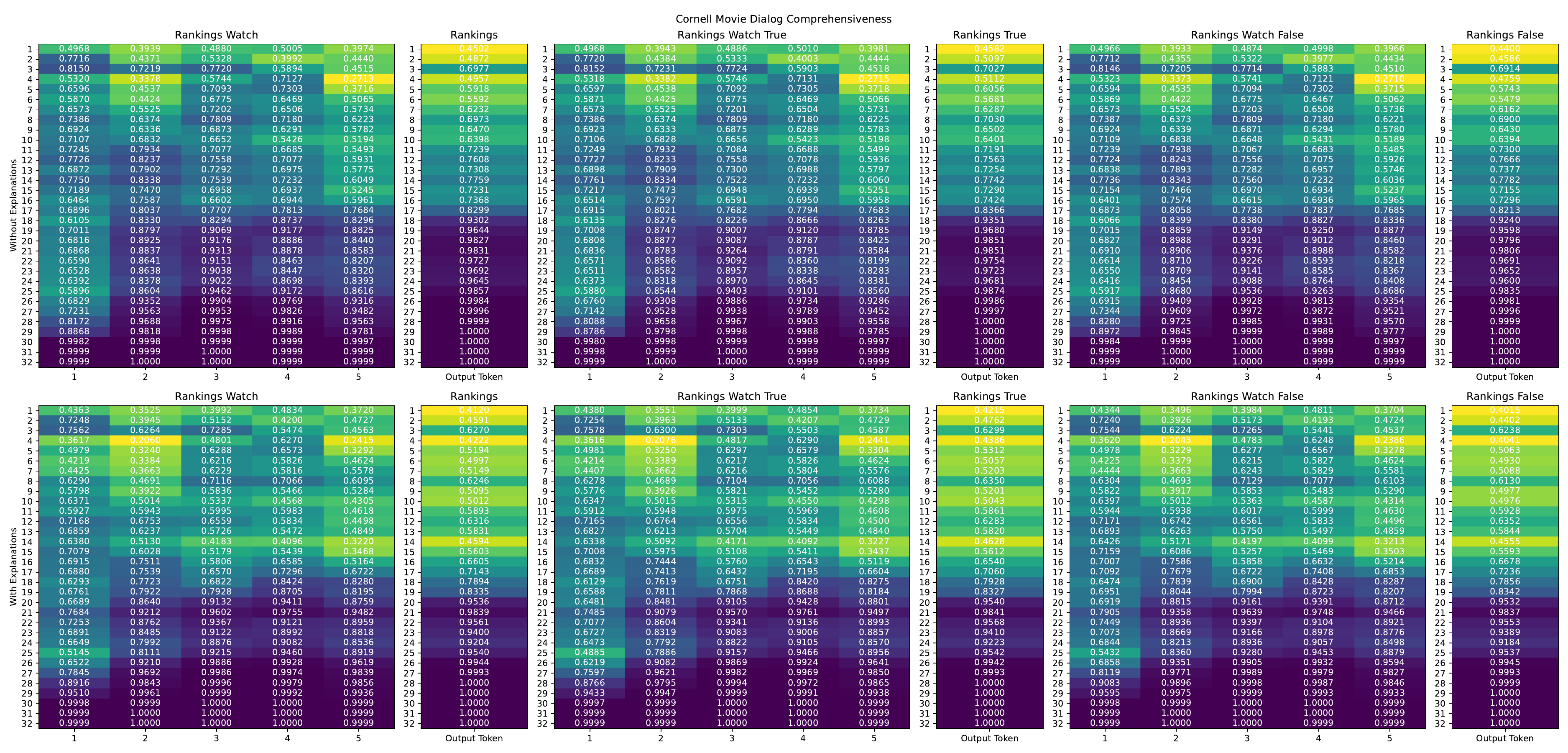}
    \caption{Output Token Ranking for Movies Dataset on Comprehensiveness dimension using Ecco \citep{alammarEccoOpenSource2021}}
\end{figure}
\begin{figure}[!h]
    \centering
    \includegraphics[width=\textwidth]{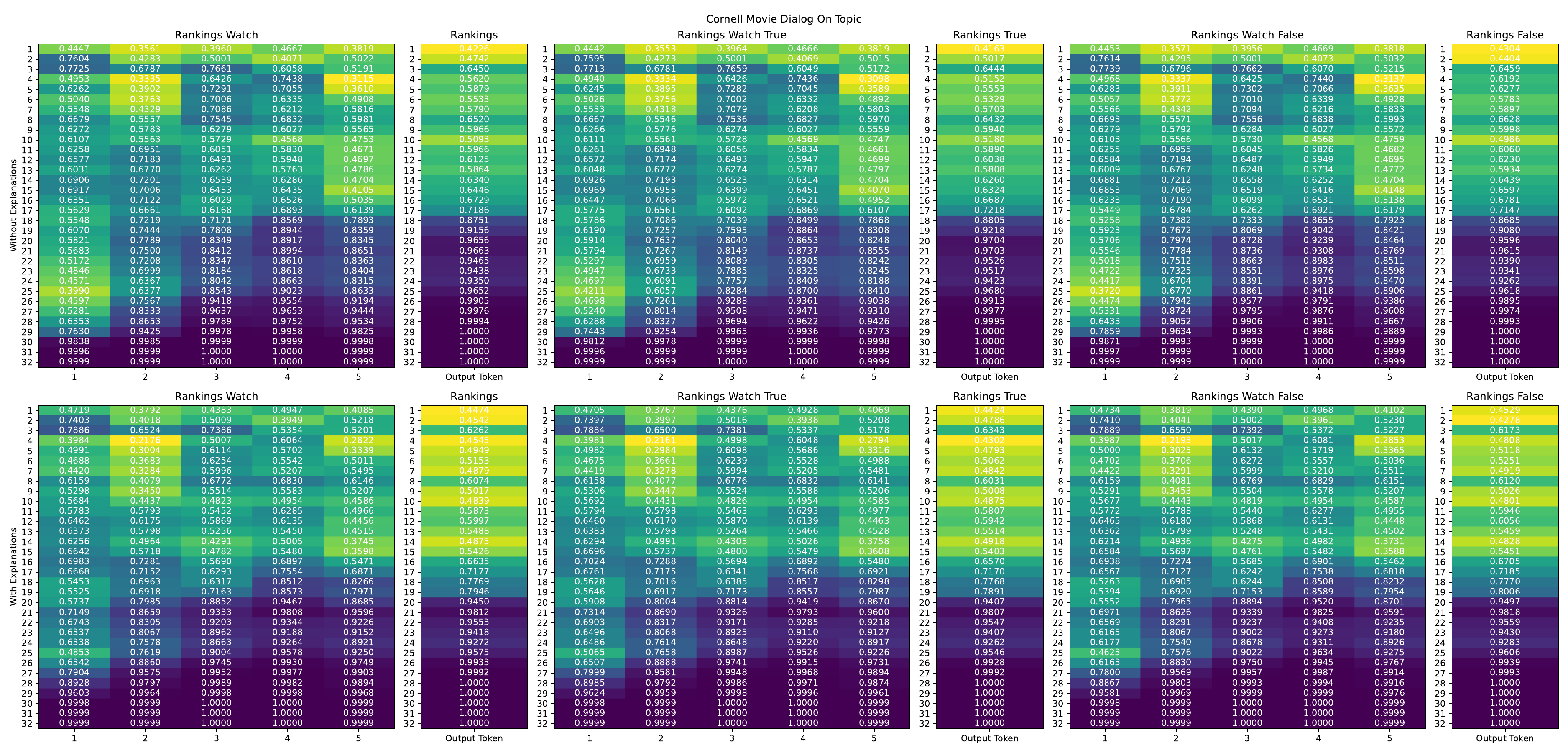}
    \caption{Output Token Ranking for Movies Dataset on On-Topic dimension using Ecco \citep{alammarEccoOpenSource2021}}
\end{figure}

%% file: sections/93-appendix-entropy.tex
\newpage
\FloatBarrier
\subsection{Entropy of Output Token}
\label{section:appendix-entropy-ouput-token}
\FloatBarrier

\begin{figure}[!h]
    \centering
    \includegraphics[width=\textwidth]{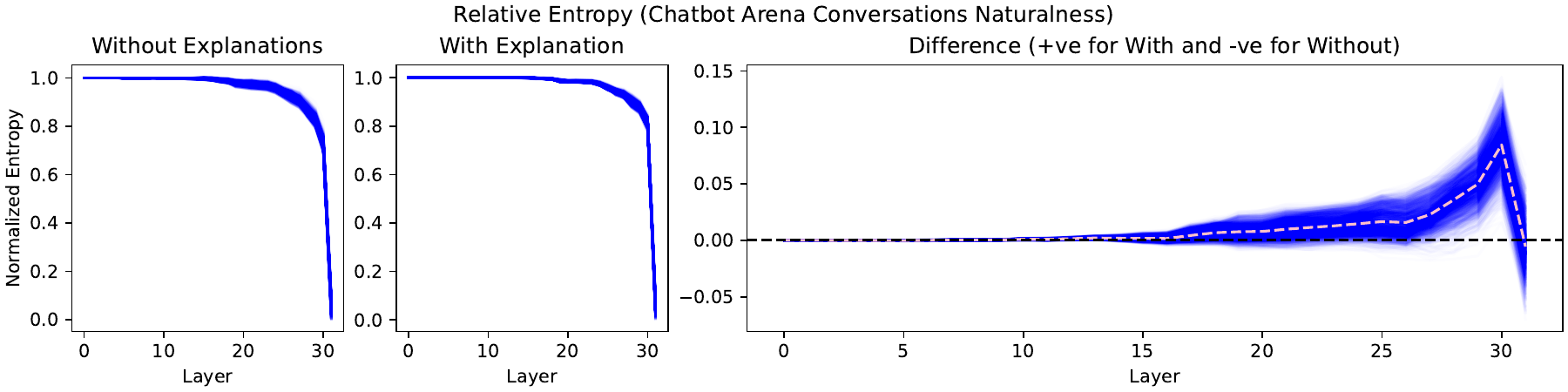}
    \caption{Entropy of Output Token for ChatbotAC Dataset on Naturalness dimension using Ecco \citep{alammarEccoOpenSource2021}}
\end{figure}
\begin{figure}[!h]
    \centering
    \includegraphics[width=\textwidth]{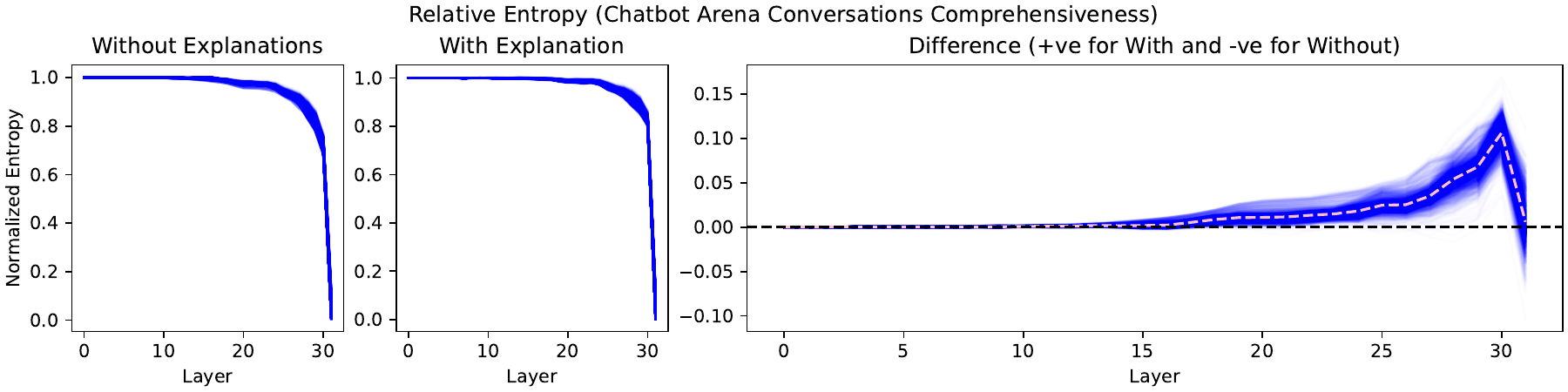}
    \caption{Entropy of Output Token for ChatbotAC Dataset on Comprehensiveness dimension using Ecco \citep{alammarEccoOpenSource2021}}
\end{figure}
\begin{figure}[!h]
    \centering
    \includegraphics[width=\textwidth]{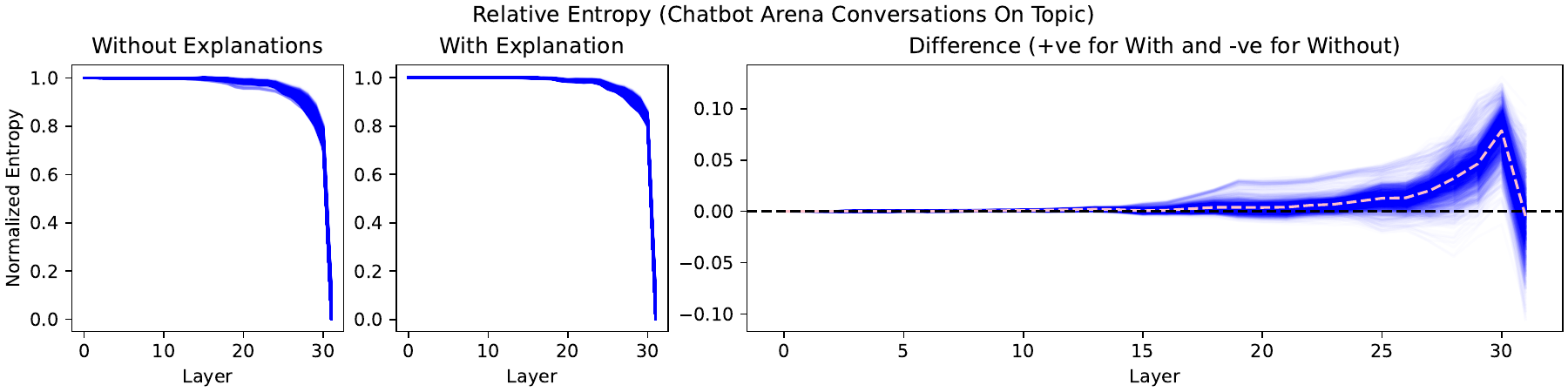}
    \caption{Entropy of Output Token for ChatbotAC Dataset on On-Topic dimension using Ecco \citep{alammarEccoOpenSource2021}}
\end{figure}

\begin{figure}[!h]
    \centering
    \includegraphics[width=\textwidth]{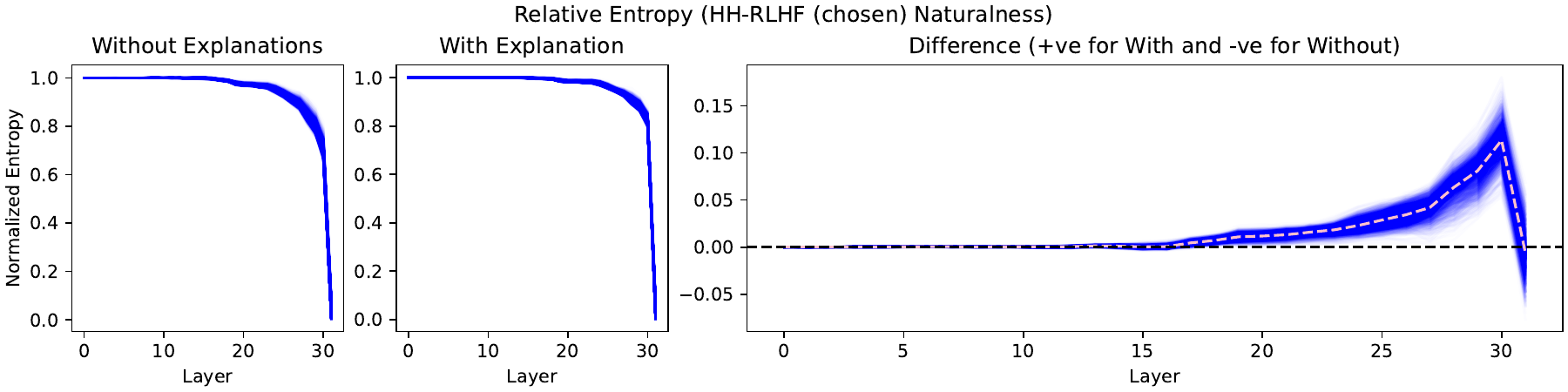}
    \caption{Entropy of Output Token for HH-C Dataset on Naturalness dimension using Ecco \citep{alammarEccoOpenSource2021}}
\end{figure}
\begin{figure}[!h]
    \centering
    \includegraphics[width=\textwidth]{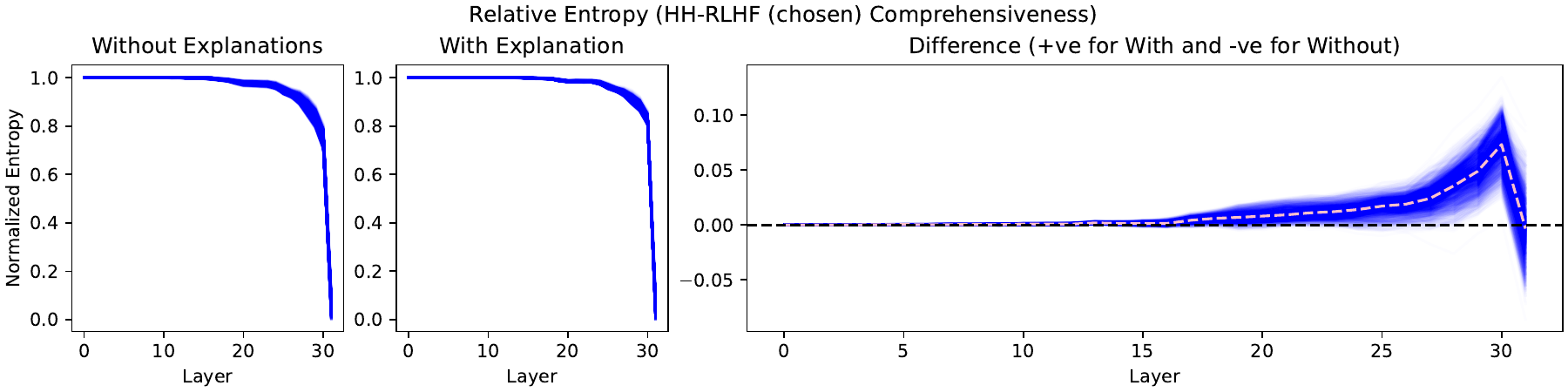}
    \caption{Entropy of Output Token for HH-C Dataset on Comprehensiveness dimension using Ecco \citep{alammarEccoOpenSource2021}}
\end{figure}
\begin{figure}[!h]
    \centering
    \includegraphics[width=\textwidth]{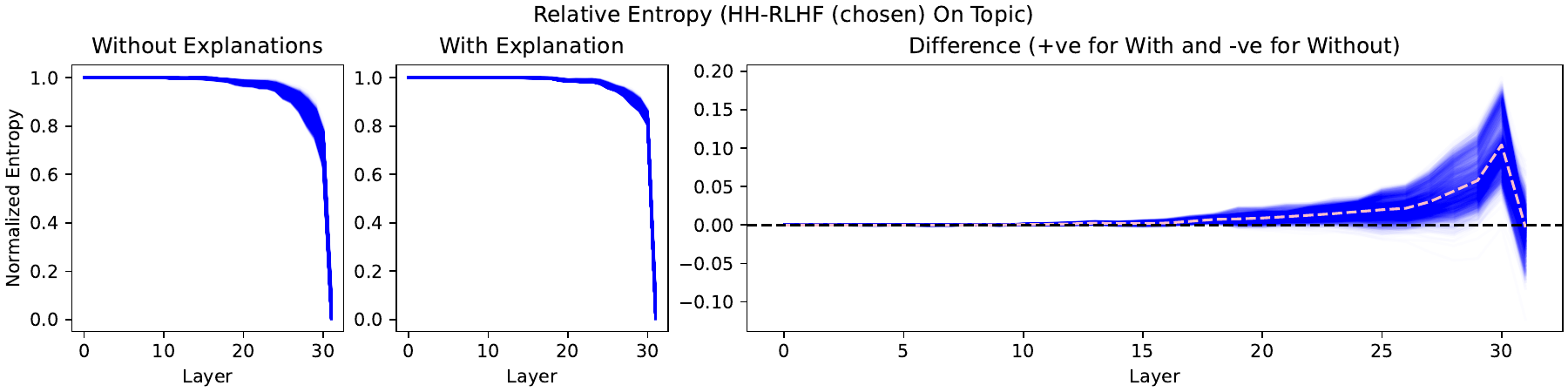}
    \caption{Entropy of Output Token for HH-C Dataset on On-Topic dimension using Ecco \citep{alammarEccoOpenSource2021}}
\end{figure}

\begin{figure}[!h]
    \centering
    \includegraphics[width=\textwidth]{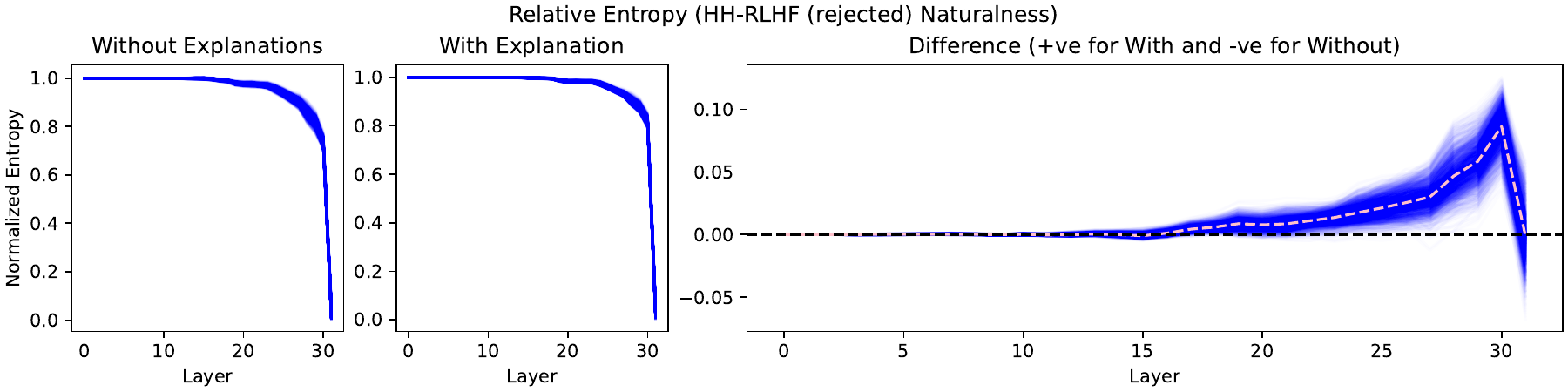}
    \caption{Entropy of Output Token for HH-R Dataset on Naturalness dimension using Ecco \citep{alammarEccoOpenSource2021}}
\end{figure}
\begin{figure}[!h]
    \centering
    \includegraphics[width=\textwidth]{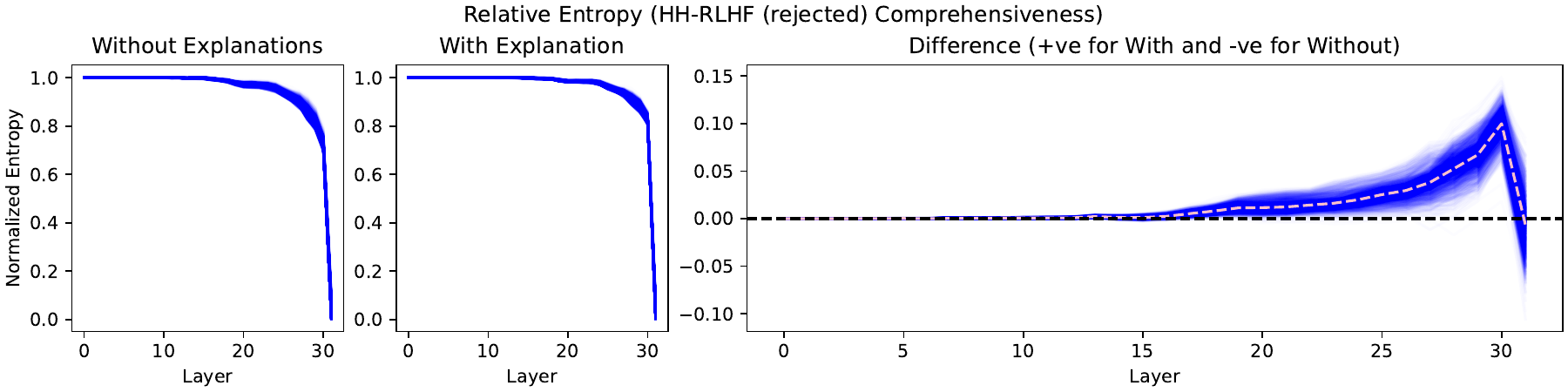}
    \caption{Entropy of Output Token for HH-R Dataset on Comprehensiveness dimension using Ecco \citep{alammarEccoOpenSource2021}}
\end{figure}
\begin{figure}[!h]
    \centering
    \includegraphics[width=\textwidth]{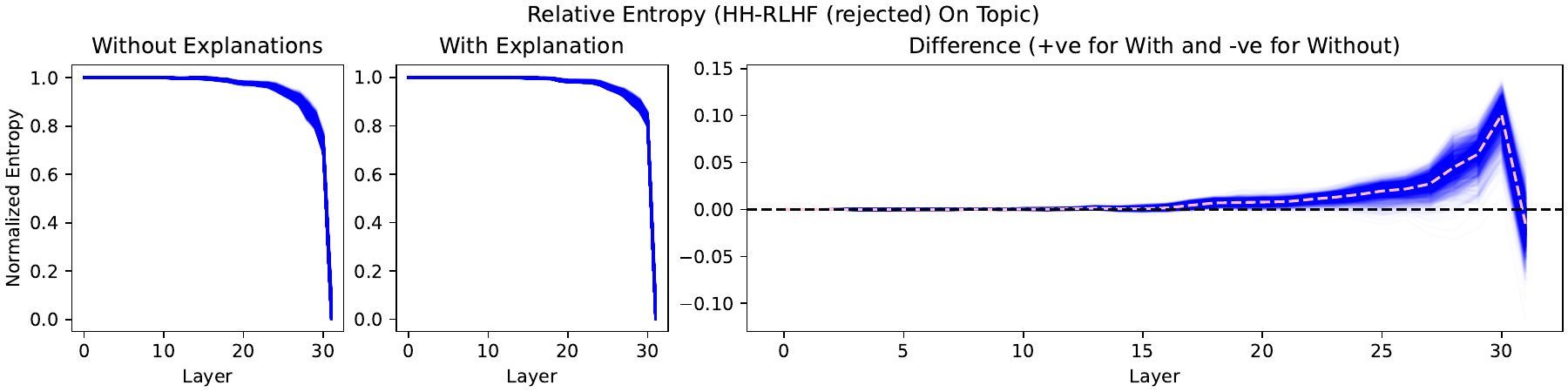}
    \caption{Entropy of Output Token for HH-R Dataset on On-Topic dimension using Ecco \citep{alammarEccoOpenSource2021}}
\end{figure}

\begin{figure}[!h]
    \centering
    \includegraphics[width=\textwidth]{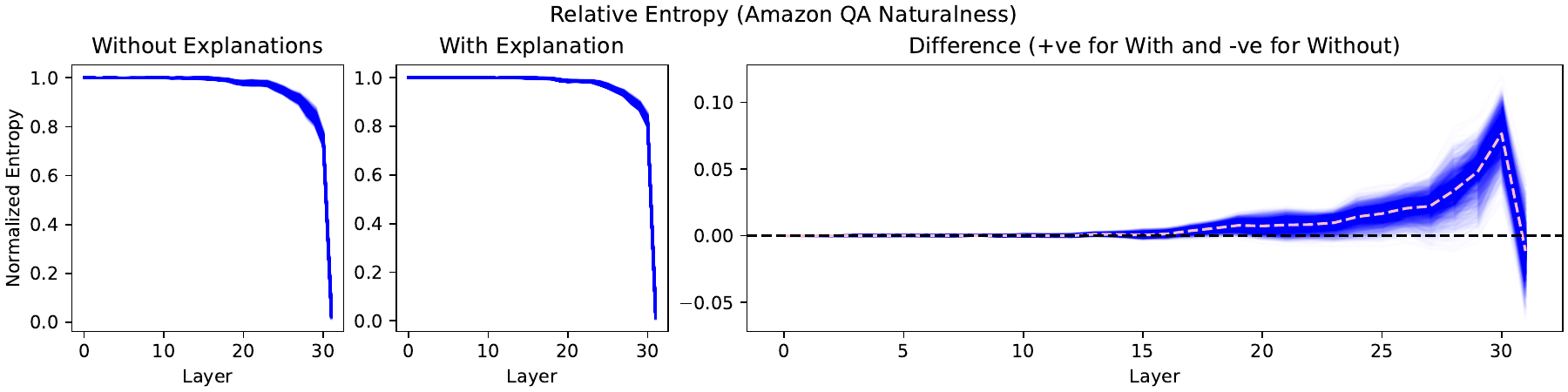}
    \caption{Entropy of Output Token for AmazonQA Dataset on Naturalness dimension using Ecco \citep{alammarEccoOpenSource2021}}
\end{figure}
\begin{figure}[!h]
    \centering
    \includegraphics[width=\textwidth]{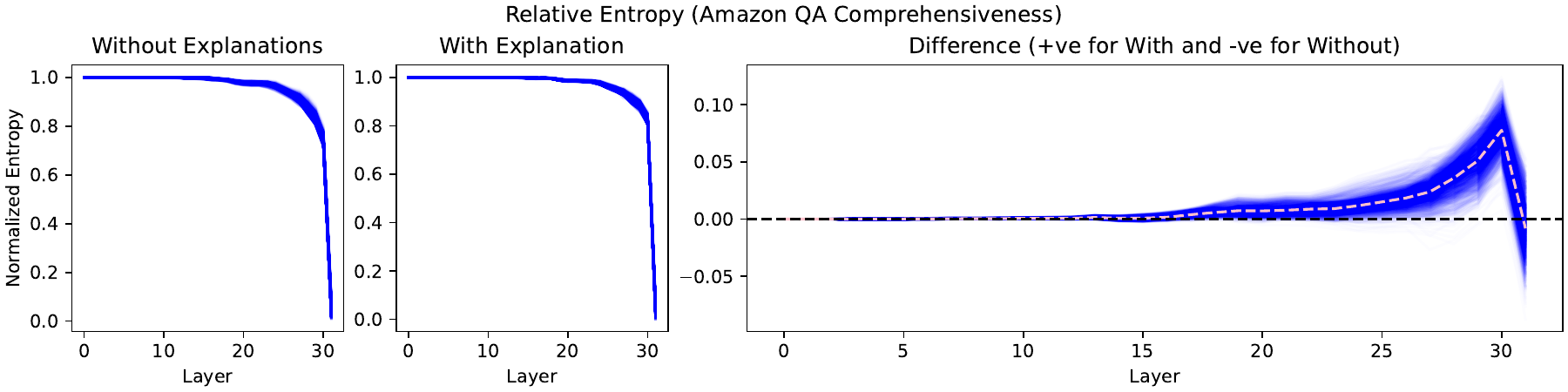}
    \caption{Entropy of Output Token for AmazonQA Dataset on Comprehensiveness dimension using Ecco \citep{alammarEccoOpenSource2021}}
\end{figure}
\begin{figure}[!h]
    \centering
    \includegraphics[width=\textwidth]{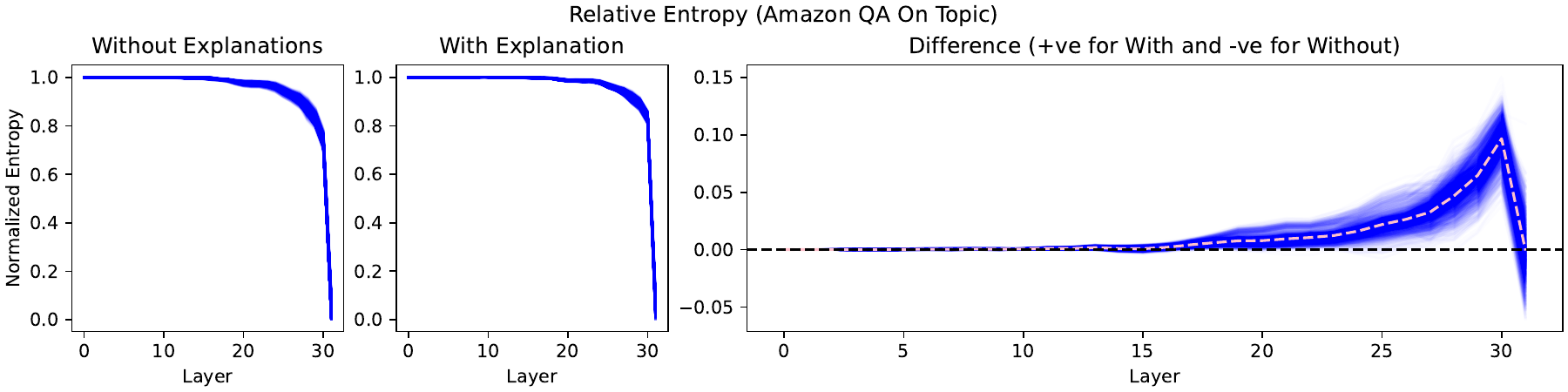}
    \caption{Entropy of Output Token for AmazonQA Dataset on On-Topic dimension using Ecco \citep{alammarEccoOpenSource2021}}
\end{figure}

\begin{figure}[!h]
    \centering
    \includegraphics[width=\textwidth]{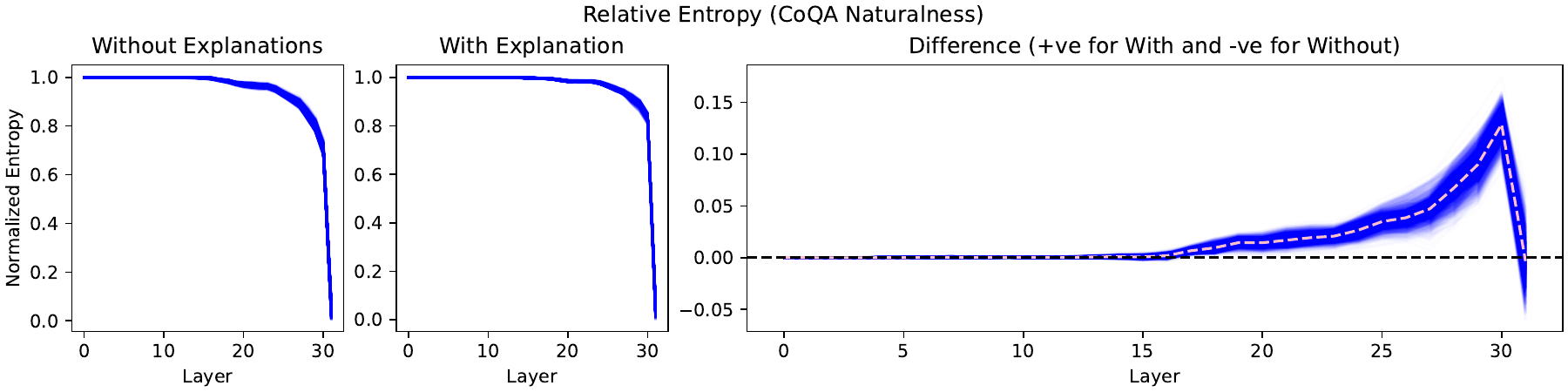}
    \caption{Entropy of Output Token for CoQA Dataset on Naturalness dimension using Ecco \citep{alammarEccoOpenSource2021}}
\end{figure}
\begin{figure}[!h]
    \centering
    \includegraphics[width=\textwidth]{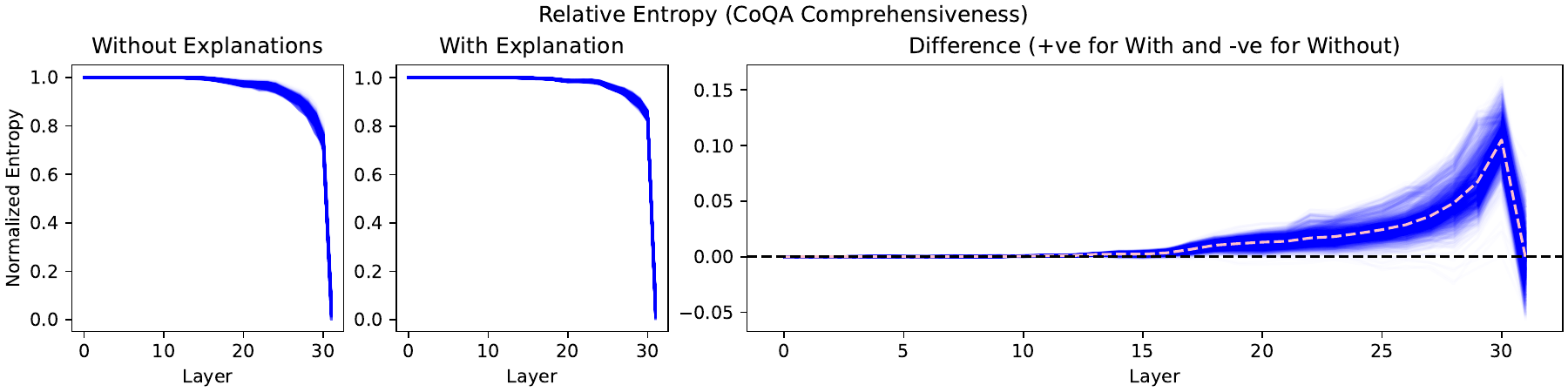}
    \caption{Entropy of Output Token for CoQA Dataset on Comprehensiveness dimension using Ecco \citep{alammarEccoOpenSource2021}}
\end{figure}
\begin{figure}[!h]
    \centering
    \includegraphics[width=\textwidth]{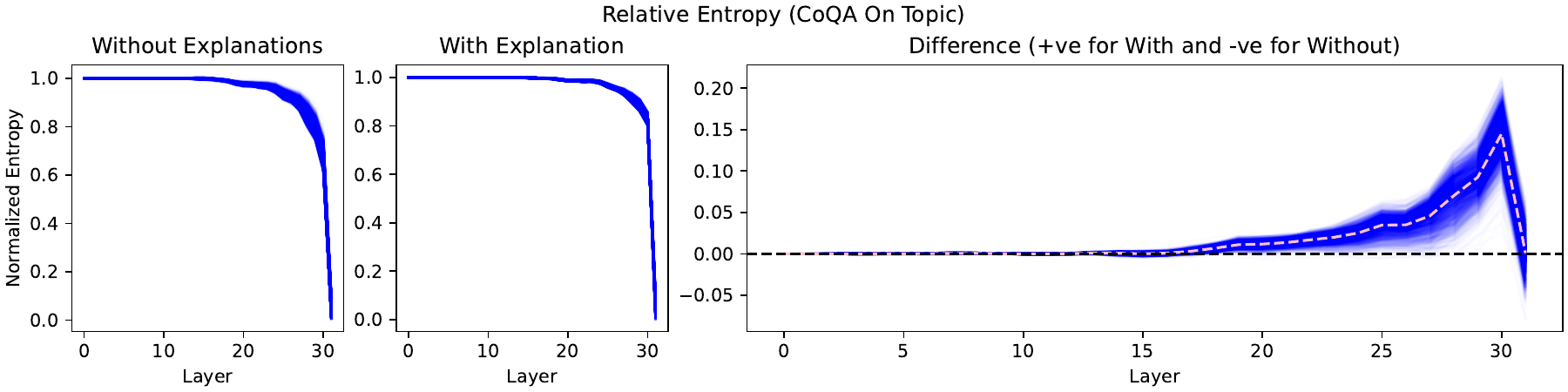}
    \caption{Entropy of Output Token for CoQA Dataset on On-Topic dimension using Ecco \citep{alammarEccoOpenSource2021}}
\end{figure}

\begin{figure}[!h]
    \centering
    \includegraphics[width=\textwidth]{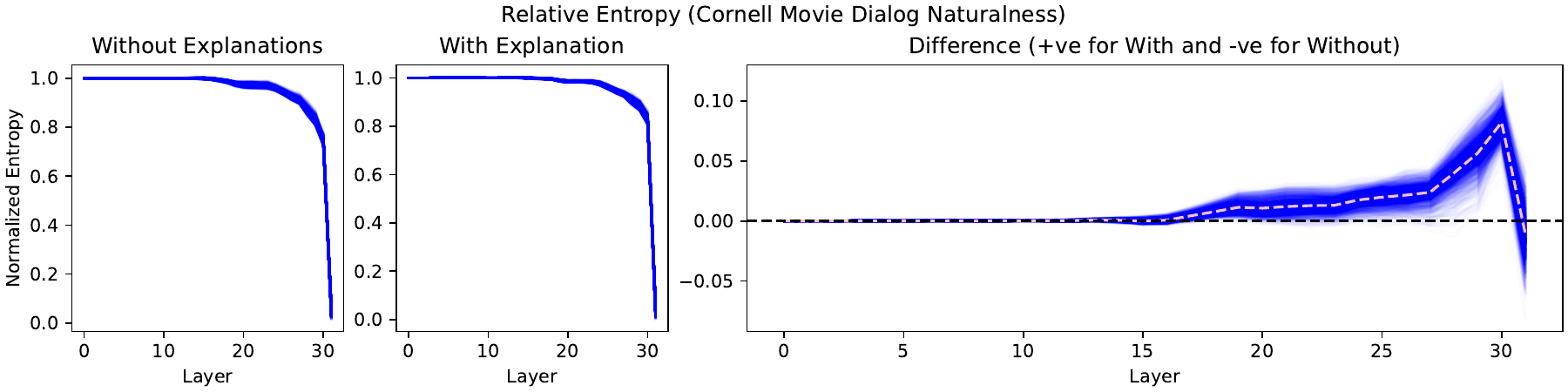}
    \caption{Entropy of Output Token for Movies Dataset on Naturalness dimension using Ecco \citep{alammarEccoOpenSource2021}}
\end{figure}
\begin{figure}[!h]
    \centering
    \includegraphics[width=\textwidth]{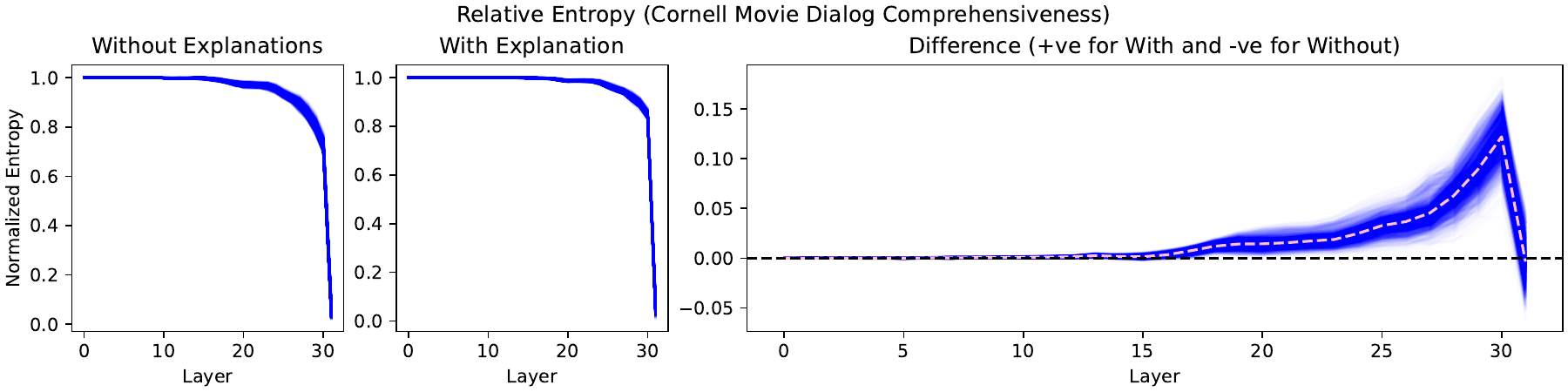}
    \caption{Entropy of Output Token for Movies Dataset on Comprehensiveness dimension using Ecco \citep{alammarEccoOpenSource2021}}
\end{figure}
\begin{figure}[!h]
    \centering
    \includegraphics[width=\textwidth]{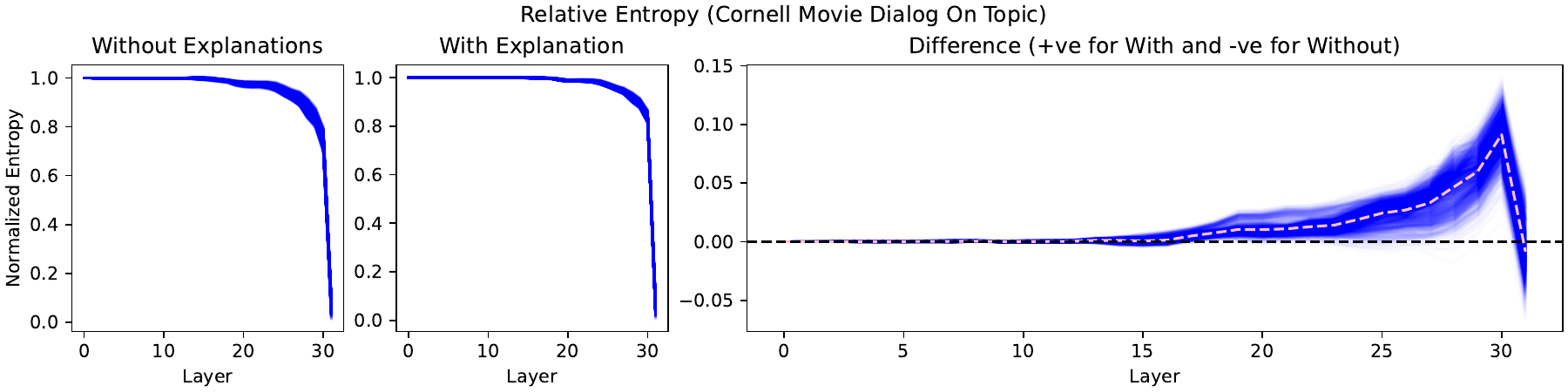}
    \caption{Entropy of Output Token for Movies Dataset on On-Topic dimension using Ecco \citep{alammarEccoOpenSource2021}}
\end{figure}